\documentclass{article} 
\usepackage{iclr2022_conference,times}

\usepackage{amsmath,amsfonts,bm}

\def\Figref#1{Figure~\ref{#1}}

\def\eqref#1{equation~\ref{#1}}

\def\1{\bm{1}}

\def\vr{{\bm{r}}}

\def\vx{{\bm{x}}}

\def\mF{{\bm{F}}}

\def\mK{{\bm{K}}}

\DeclareMathAlphabet{\mathsfit}{\encodingdefault}{\sfdefault}{m}{sl}
\SetMathAlphabet{\mathsfit}{bold}{\encodingdefault}{\sfdefault}{bx}{n}

\def\sR{{\mathbb{R}}}
\def\sS{{\mathbb{S}}}

\usepackage{hyperref}
\usepackage{url}
\usepackage{booktabs}
\usepackage{graphicx}
\usepackage{subcaption}
\usepackage{enumitem}
\usepackage{multirow}

\newcommand{\eg}{\emph{e.g.}}
\newcommand{\ie}{\emph{i.e.}}

\newcommand{\std}{\texttt{STD}~}

\title{Network Pruning Spaces}

\author{Xuanyu He$^1$, Yu-I Yang$^2$, Ran Song$^1$, Jiachen Pu$^2$, Conggang Hu$^2$, \\
\textbf{Feijun Jiang$^2$, Wei Zhang$^1$, Huanghao Ding$^2$} \\
$^1$Shandong University, $^2$Alibaba Group \\
\texttt{xyhe14@gmail.com}
}

\iclrfinalcopy 
\begin{document}

\maketitle

\begin{abstract}
Network pruning techniques, including weight pruning and filter pruning, reveal that most state-of-the-art neural networks can be accelerated without a significant performance drop. This work focuses on filter pruning which enables accelerated inference with any off-the-shelf deep learning library and hardware. We propose the concept of \emph{network pruning spaces} that parametrize populations of subnetwork architectures. Based on this concept, we explore the structure aspect of subnetworks that result in minimal loss of accuracy in different pruning regimes and arrive at a series of observations by comparing subnetwork distributions. We conjecture through empirical studies that there exists an optimal FLOPs-to-parameter-bucket ratio related to the design of original network in a pruning regime. Statistically, the structure of a winning subnetwork guarantees an approximately optimal ratio in this regime. Upon our conjectures, we further refine the initial pruning space to reduce the cost of searching a good subnetwork architecture. Our experimental results on ImageNet show that the subnetwork we found is superior to those from the state-of-the-art pruning methods under comparable FLOPs.
\end{abstract}


\section{Introduction}
\label{sec:intro}

Large neural networks are usually preferred as they exhibit better generalization capability than small ones. In many model families such as ResNets~\citep{he2016deep} and MobileNets~\citep{howard2017mobilenets}, large networks consistently achieve higher accuracy and are superior to small ones. Although large networks have revolutionized many fields such as computer vision and natural language processing, they suffer from significant inference costs in practice, especially when used with embedded sensors or mobile devices. For many practical applications, computational efficiency and small network sizes are crucial factors in addition to performance.

Recent works on network pruning reveal that most state-of-the-art models are over-parameterized~\citep{li2017pruning,frankle2018lottery}.
We can remove weights~\citep{frankle2018lottery}, filters~\citep{li2017pruning,luo2017thinet} and other structures~\citep{meng2020pruning} from large networks without inducing significant performance drop by network pruning techniques.
Such strategies reduce the resource demands of neural network inference, including storage requirements, energy consumption and latency, and thus increase runtime efficiency.
This work focuses on filter pruning, which prunes the original network to a slimmer subnetwork. Unlike weight pruning approaches that lead to irregular sparsity patterns and require specialized libraries or hardware for computational speedups, filter pruning enables accelerated inference with any off-the-shelf deep learning library and hardware.

In the literature, a well-established paradigm is to train a large network to completions, prune the network according to some heuristics and retrain the pruned subnetwork to recover the accuracy loss~\citep{renda2020comparing}. Despite the effectiveness of this \emph{prune-then-retrain} paradigm, existing filter pruning methods have one major limitation: their outcome is merely a single pruning recipe tuned to a specific setting\footnote{
In this work, we refer to the combination of pruning ratios for the layers in a network as \emph{pruning recipe}.
} and may fail to generalize to new settings. For example, we find that some filter pruning methods cannot produce subnetworks that outperform a regular subnetwork with an uniform pruning ratio in some extreme pruning regimes where we reduce more than $90\%$ FLOPs of a network. Moreover, the prune-then-retrain paradigm does not deepen our understanding about the reason why these recipes produced by pruning approaches work well in specific regimes.

Instead of producing the best single pruning recipe like existing works~\citep{li2017pruning,he2018amc,li2020eagleeye}, we explore the general principles that can help us understand and refine pruning algorithms.
Inspired by~\citet{radosavovic2019network,radosavovic2020designing}, we present \emph{network pruning spaces} in this work.
Given a network, its network pruning space is a large population of subnetwork architectures produced by pruning recipes.
This is different from architecture search spaces~\citep{zoph2017neural,tan2019efficientnet} and network design spaces~\citep{radosavovic2019network,radosavovic2020designing}, which aim to produce model families with good generalization from scratch.
Specifically, we start with a well-designed network architecture and sample subnetworks from its pruning space, which gives rise to a subnetwork distribution. Based on this distribution, we present a tool to analyze the pruning space. With a constraint on pruning recipes, we divide the initial pruning space into several subspaces to explore the structure aspect of winning subnetworks and reduce the cost of searching such a subnetwork under different settings.
The implementation is still filter pruning in common, but elevated to the population level and guided via distribution estimates following~\citet{radosavovic2019network}.

In our empirical studies, the core observations on network pruning spaces are simple but interesting:
(\emph{a}) There exists an optimal FLOPs-to-parameter-bucket ratio in a pruning regime;
(\emph{b}) Subnetworks with the optimal FLOPs-to-parameter-bucket ratios will be winning ones;
(\emph{c}) The limitation of performance in a pruning regime is predictable by a function of the optimal FLOPs-to-parameter-bucket ratio.
With these observations, we return to examine some subnetworks produced by existing pruning methods~\citep{luo2017thinet,li2017pruning}.
We find that these subnetworks with good performance match our observations (\emph{a} and \emph{b}). Moreover, our Observation \emph{c} is consistent with \citet{rosenfeld2021predictability} in weight pruning, suggesting that we are able to use an empirical function to make trade-offs between efficiency and accuracy when pruning filters in a network.

The contributions of this work are threefold:
\begin{itemize}[leftmargin=*]
\item We present the new concept of network pruning spaces that parametrize populations of subnetwork architectures in the field of filter pruning. With this concept, we empirically study the structure aspect of winning subnetworks in different pruning regimes and explore general pruning principles.

\item Based on our experimental results, we make a series of conjectures that help us understand filter pruning principles. With these conjectures, we refine the initial pruning space with a constraint on FLOPs-to-parameter-bucket ratio to reduce the cost of searching a winning subnetwork.

\item We analytically show that the limitation of performance in a pruning regime is predictable, suggesting that we can build an empirical tool for reasoning trade-offs between efficiency and accuracy in filter pruning.
\end{itemize}

\section{Revisit filter pruning}

In this section, we revisit existing techniques in filter pruning and introduce an efficient setting for further exploration.
Generally, a pruning method involves instantiating each of the steps in the prune-then-retrain paradigm from a range of choices. The combination of such choices has a major impact on the final performance~\citep{renda2020comparing}. The combinations of pruning implementations vary from each other in accuracy and efficiency. Since we aim to explore general pruning principles that can interpret the behaviors of filter pruning, a qualified and efficient setting is needed.
In the following, we assess different pruning implementations and finally introduce a standard setting for conducting empirical studies. Our main experiments use ResNet-50~\citep{he2016deep} on CIFAR-10~\citep{krizhevsky2009learning}.
The input images are resized to $224 \times 224$ such that we only modify the last fully-connected layer in ResNet-50.

\def\pruningfigwidth{0.325}
\begin{figure}[t]
\centering
\begin{subfigure}[b]{\pruningfigwidth\textwidth}
\includegraphics[width=\linewidth]{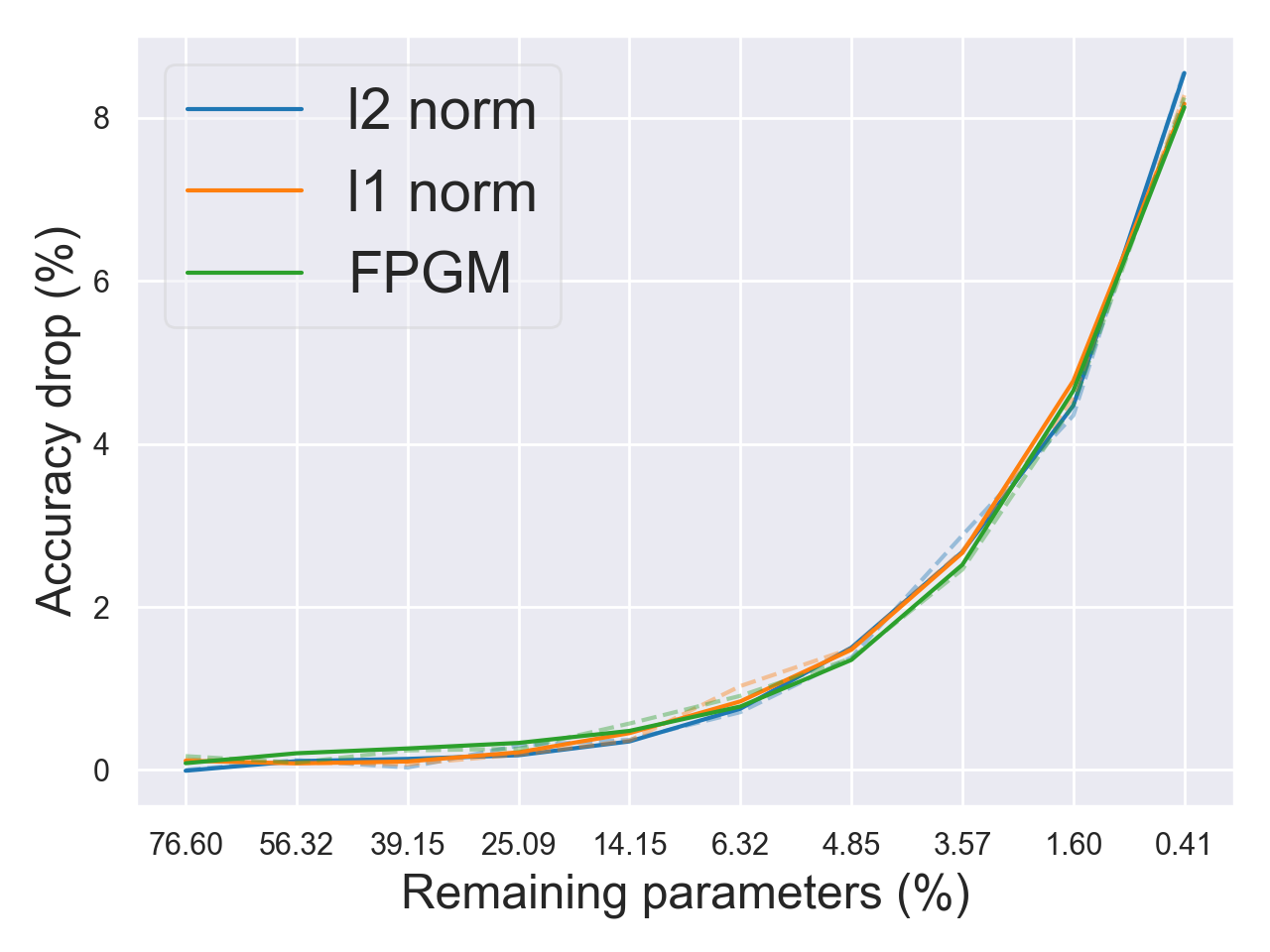}
\end{subfigure}
\begin{subfigure}[b]{\pruningfigwidth\linewidth}
\includegraphics[width=\linewidth]{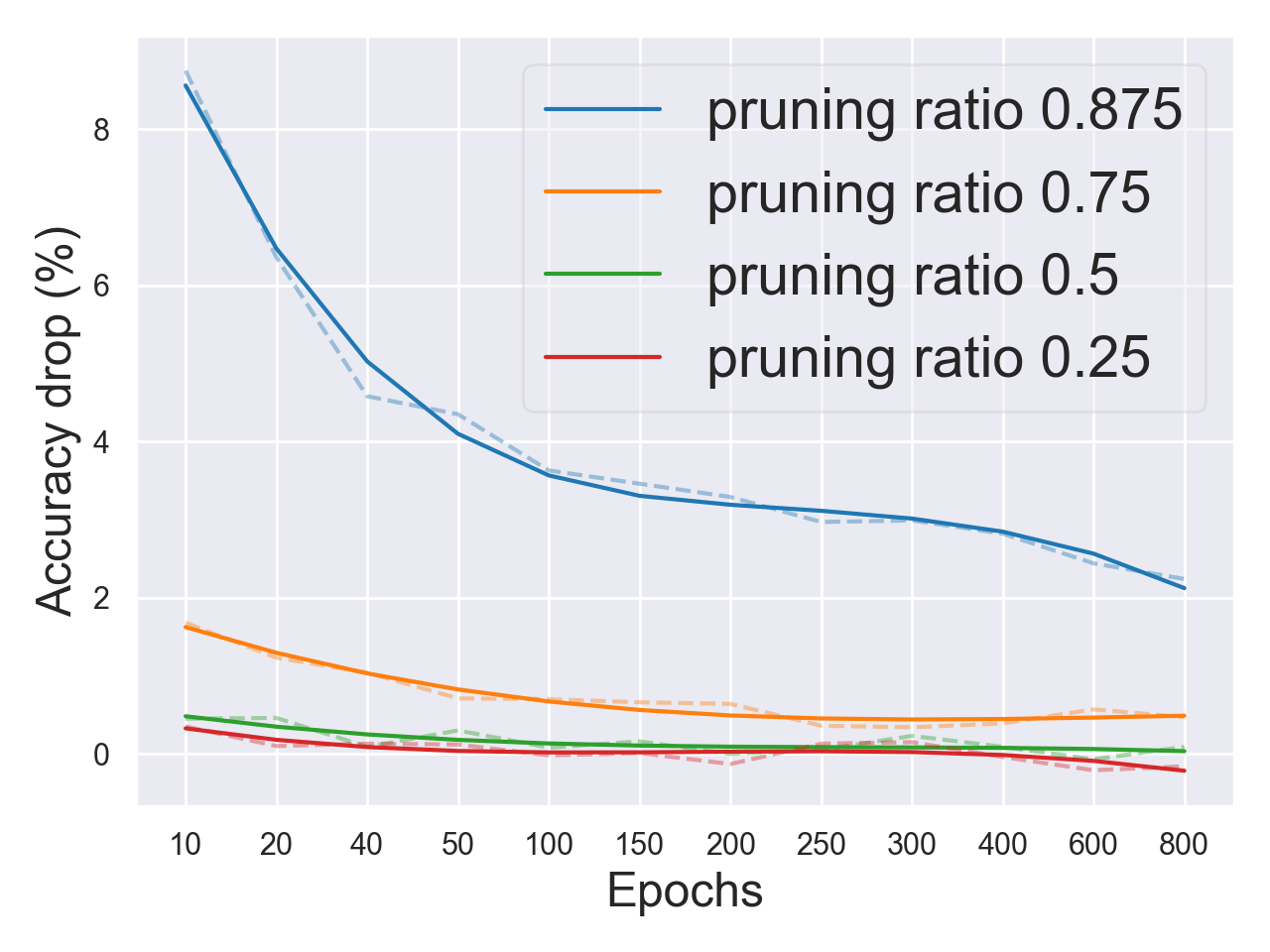}
\end{subfigure}
\begin{subfigure}[b]{\pruningfigwidth\linewidth}
\includegraphics[width=\linewidth]{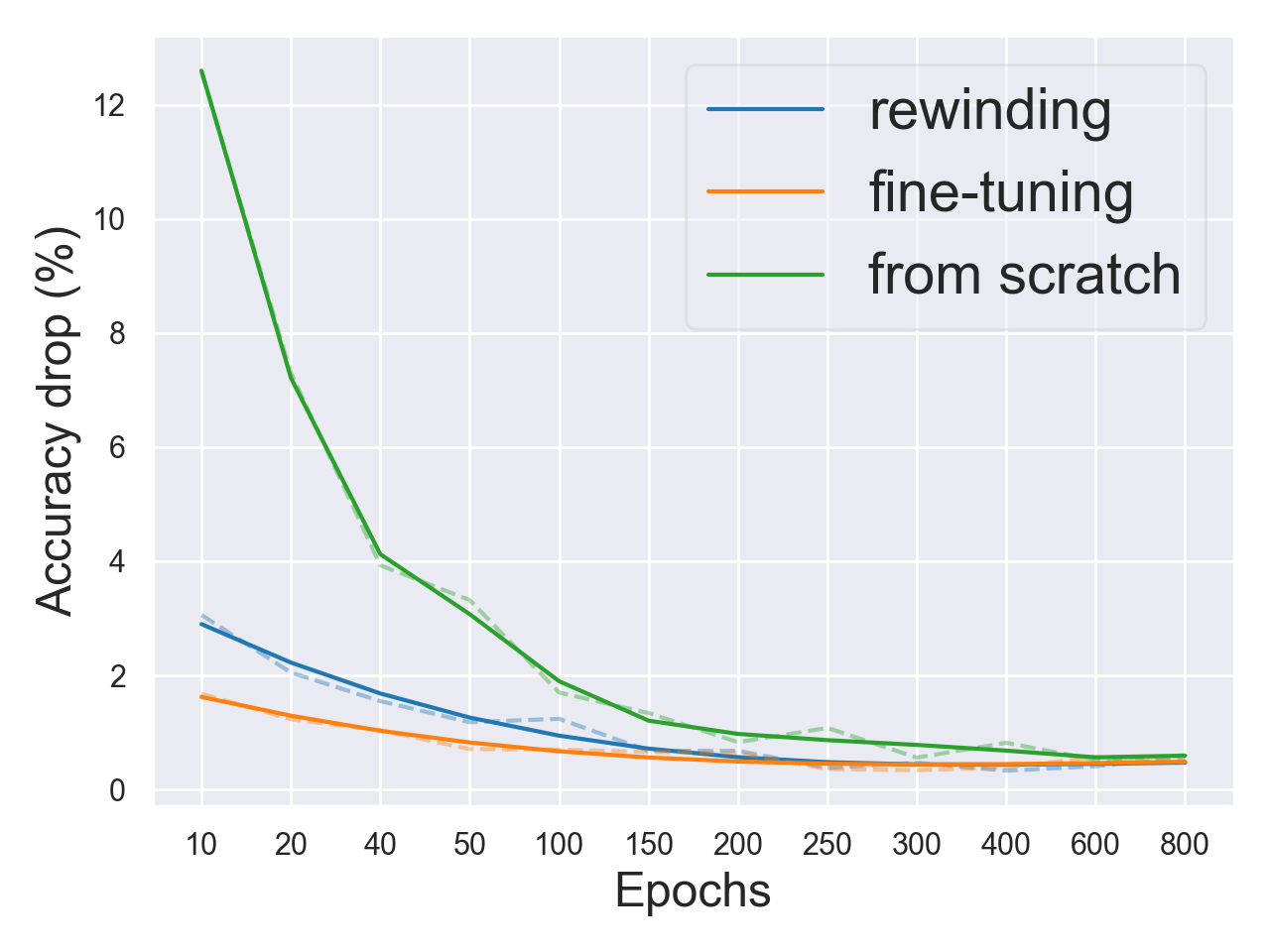}
\end{subfigure}
\caption{Pruning filters of ResNet-50 on CIFAR-10.
(\emph{Left}) Pruning filters of ResNet-50 based on different pruning heuristics, including $\ell_1$ norm, $\ell_2$ norm and FPGM~\citep{he2019filter}.
(\emph{Center}) Experiments on fine-tuning epochs with different pruning ratios.
(\emph{Right}) Comparison of fine-tuning, rewinding and retraining from scratch with an uniform pruning ratio of $0.75$.}
\label{fig:pruning}
\end{figure}

\subsection{Pruning filters}

We briefly describe the procedure of pruning filters on the convolutional layers.
Extending it to other layers is straightforward.
Let $c_i$ denote the number of input channels for the $i$th convolutional layer in a network and $h_i / w_i$ be the height/width of the input feature maps.
The convolutional layer transforms the input feature maps $\vx_i \in \sR^{c_i \times h_i \times w_i}$ into the output feature maps $\vx_{i + 1} \in \sR^{c_{i + 1} \times h_{i + 1} \times w_{i + 1}}$, which are used as input feature maps for the next convolutional layer, by applying $c_{i + 1}$ filters on the $c_i$ input channels.

Each filter $\mF_{i, j} \in \sR^{c_i \times k \times k}$ is composed by $c_i$ 2D kernels $\mK \in \sR^{k \times k}$ and generates one feature map $\vx_{i + 1, j} \in \sR^{1 \times h_{i + 1} \times w_{i + 1}}$ on top of $\vx_i \in \sR^{c_i \times h_i \times w_i}$.
All the filters constitute the convolutional kernel $\mF_i \in \sR^{c_i \times c_{i + 1} \times k \times k}$.
When $m$ filters are pruned with a ratio $r_i = m / c_{i + 1}$, $m$ corresponding feature maps are removed.
It reduces $m / c_{i + 1}$ of the computational cost for both layers $i$ and $i + 1$.

\textbf{Filter importance.}
There exist a wide variety of pruning heuristics for evaluating the importance of filters.
\Figref{fig:pruning} (left) shows accuracy drop curves of some pruning heuristics when we prune more filters.
The difference between these pruning heuristics is small and they have the same trend when pruning filters.
We adopt $\ell_2$ norm pruning heuristic in this work because of its simplicity.

\textbf{One-shot pruning.}
We prune the network to a target size at once.
In contrast, the prune-retrain circle can be repeated until a target network size is reached, which is known as iterative pruning.
\citet{renda2020comparing} and our experiments (not shown) suggest that iterative pruning is slightly superior to one-shot pruning in high-pruning-ratio regimes; however, it takes much more time on retraining (about $5 \times$).
Such an accuracy gain can be ignored considering that one-shot pruning saves a lot of time, especially when we need to study a large population of pruned subnetworks in this work.

\subsection{Retraining subnetworks}
\label{sec:baseline.retraining}

When the filters of the large network are removed after pruning, accuracy significantly decreases.
It is standard to retrain the pruned subnetwork with the remaining weights to recover the original accuracy.
We discuss and evaluate conventional retraining techniques in the literature.
Our experiments prune the network with different uniform pruning ratios and retrain pruned subnetworks using existing techniques.

\textbf{Fine-tuning.}
The first common retraining technique is fine-tuning, which retrains pruned subnetworks for a specified number of epochs $t$ using a small learning rate $lr$~\citep{molchanov2019importance,renda2020comparing}.
In this work, we use a cosine annealing learning rate schedule~\citep{loshchilov2017sgdr} with an initial $lr = 0.01$ and empirically study the influence of $t$ in the field of filter pruning.
As shown in~\Figref{fig:pruning} (center), in low-pruning-ratio regimes, it is sufficient to fine-tune pruned subnetworks for a few epochs (\eg, 50 epochs under an uniform pruning ratio of 0.25).
In high-pruning-ratio regimes, where more than $90\%$ parameters are removed after pruning, subnetworks require more epochs for fine-tuning to recover the original accuracy.

\textbf{Learning rate rewinding.}
\citet{renda2020comparing} introduce a retraining technique, namely learning rate rewinding, which retrains the pruned networks using the original schedule from last $t$ epochs of training.
During retraining, the learning rate in rewinding is always larger than that used in fine-tuning.
In this work, we adopt a slightly different rewinding implementation, which starts from the initial learning rate $lr$ in training and retrains pruned subnetworks for $t$ epochs with warming up for 5 epochs.
Our implementation is similar to learning rate restarting~\citep{le2021network} with a cosine annealing schedule in practice.\footnote{
We consider learning rate rewinding and learning rate restarting as the same type of retraining technique, as both of them rely on a large learning rate.}
\Figref{fig:pruning} (right) shows the experimental results in a high-pruning-ratio regime.
Comparison between rewinding and fine-tuning shows that their performance is almost the same after retraining 200 epochs.
However, in a low-epoch retraining regime, fine-tuning is more efficient than rewinding for filter pruning.

\textbf{From scratch.}
We also conduct experiments that retrain pruned subnetworks from scratch, where the parameters of a subnetwork are re-initialized after pruning.
The original learning rate schedule is adopted during retraining.
As shown in~\Figref{fig:pruning} (right, green line), retraining from scratch requires more epochs to reduce accuracy drop compared to fine-tuning and rewinding.
When retrained for enough epochs (\eg, 600 or 800 epochs), retraining from scratch is able to produce comparable performance, which is consistent with~\citet{liu2019rethinking}.\footnote{
In our experiments, the others outperform retraining from scratch with a less accuracy drop of 0.3.}
Moreover, our experimental results suggest that it is unfair to compare retraining techniques under different epochs, since all of them benefit from more epochs.

\textbf{Discussion.}
In this subsection, we empirically study existing retraining techniques, including \emph{fine-tuning}, \emph{rewinding} and \emph{retraining from scratch}.
Our experimental results show that these retraining techniques can achieve similar performance if we retrain pruned subnetworks with enough epochs.
However, fine-tuning for a few epochs is a more efficient choice among them as our focus is not on recovering the original accuracy as much as possible in the following.
Finally, we introduce a standard setting for further exploration.
Given a pruning recipe, the standard setting prunes filters by their $\ell_2$ norms in a one-shot manner, and fine-tunes pruned subnetworks with $lr = 0.01$ for $t$ epochs.

\section{Network pruning spaces}
\label{sec:method}

In this section, we introduce \emph{network pruning spaces}, which are inspired by the concept of network design spaces~\citep{radosavovic2019network}.
Given a trained network $f_\theta$, a network pruning space $\sS$ comprises a large population of subnetworks $\{f_{\theta_{\textnormal{sub}}}\}$ pruned from $f_\theta$.
The initial pruning space $\sS$ is loosely defined by a constraint on FLOPs.
We refer to the constraint on FLOPs as $c_{\textnormal{flops}}$, which is a relative value calculated by $c_{\textnormal{flops}} = \textnormal{FLOPs}(f_{\theta_{\textnormal{sub}}}) / \textnormal{FLOPs}(f_\theta)$ in this work.
In practice, we use a range $c_{\textnormal{flops}} \pm \delta$ ($\delta = 0.002$) for efficiency, as it is difficult to search recipes with a fixed constraint.

Based on the specified pruning heuristics (\eg, our standard setting), a subnetwork $f_{\theta_{\textnormal{sub}}}$ in $\sS$ is generated by a pruning recipe, which consists of pruning ratios $\vr = \{r_1, r_2, \dots, r_N\}$ for $N$ layers.
We sample a number of pruning recipes from $\sS$, giving rise to a subnetwork distribution, and analyze the pruning space.

\textbf{Tools for pruning spaces.}
We adopt accuracy drop empirical distribution functions (EDFs)~\citep{radosavovic2019network,radosavovic2020designing} to analyze the pruning space.
Given a set of $n$ subnetworks with accuracy drops $\{e_i\}$, the accuracy drop EDF is given by:
\begin{equation}
F(e) = \frac{1}{n} \sum^n_{i = 1} \mathbf{1} [e_i < e],
\end{equation}
where $\mathbf{1}$ is the indicator function.
$F(e)$ represents the fraction of subnetworks with accuracy drop less than $e$ after retraining.

\subsection{The \texttt{STD} pruning spaces}

\def\spacesfigwidth{0.245}
\begin{figure}[!t]
\centering
\begin{subfigure}[b]{\spacesfigwidth\linewidth}
\includegraphics[width=\linewidth]{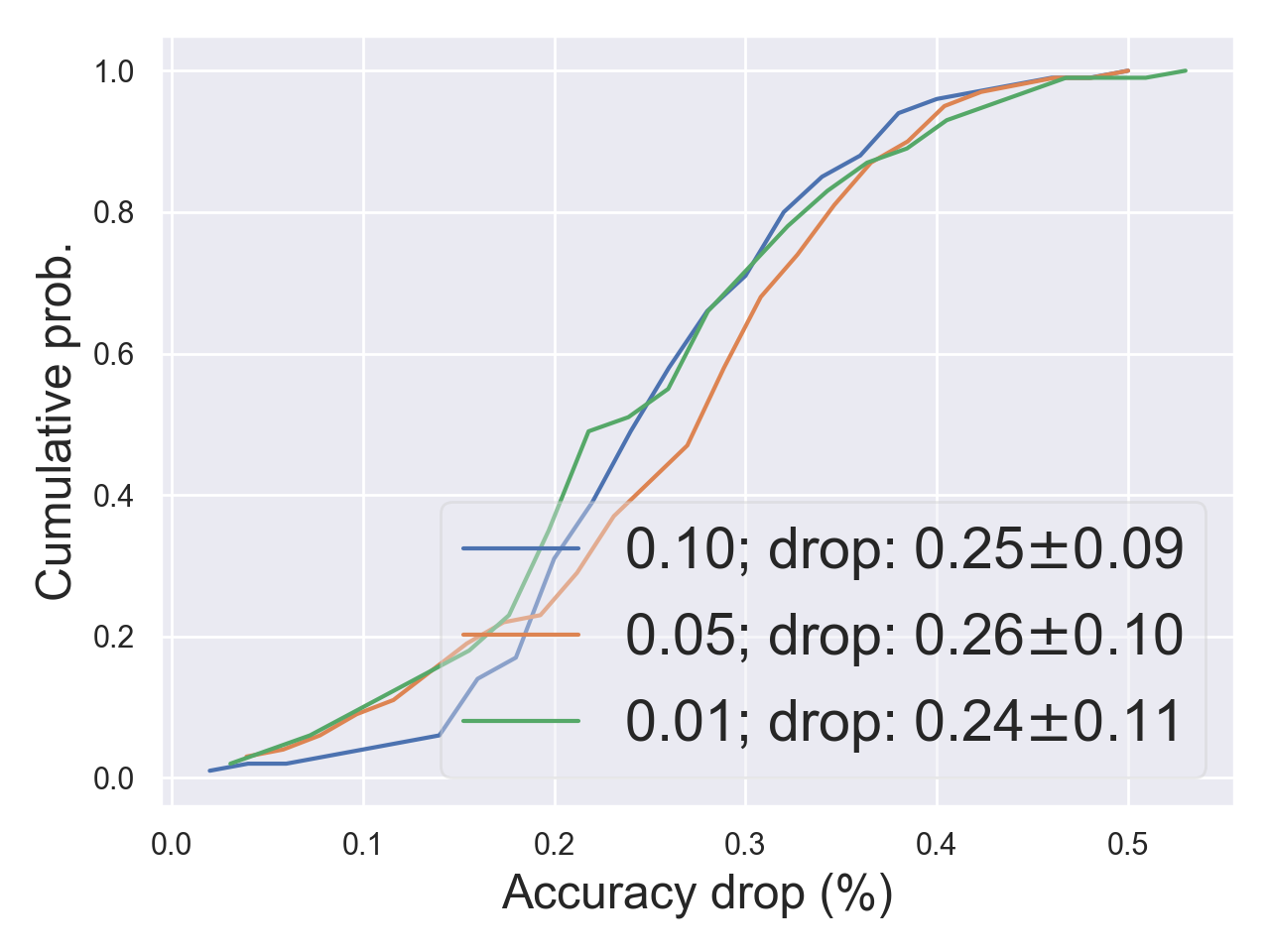}
\end{subfigure}
\begin{subfigure}[b]{\spacesfigwidth\linewidth}
\includegraphics[width=\linewidth]{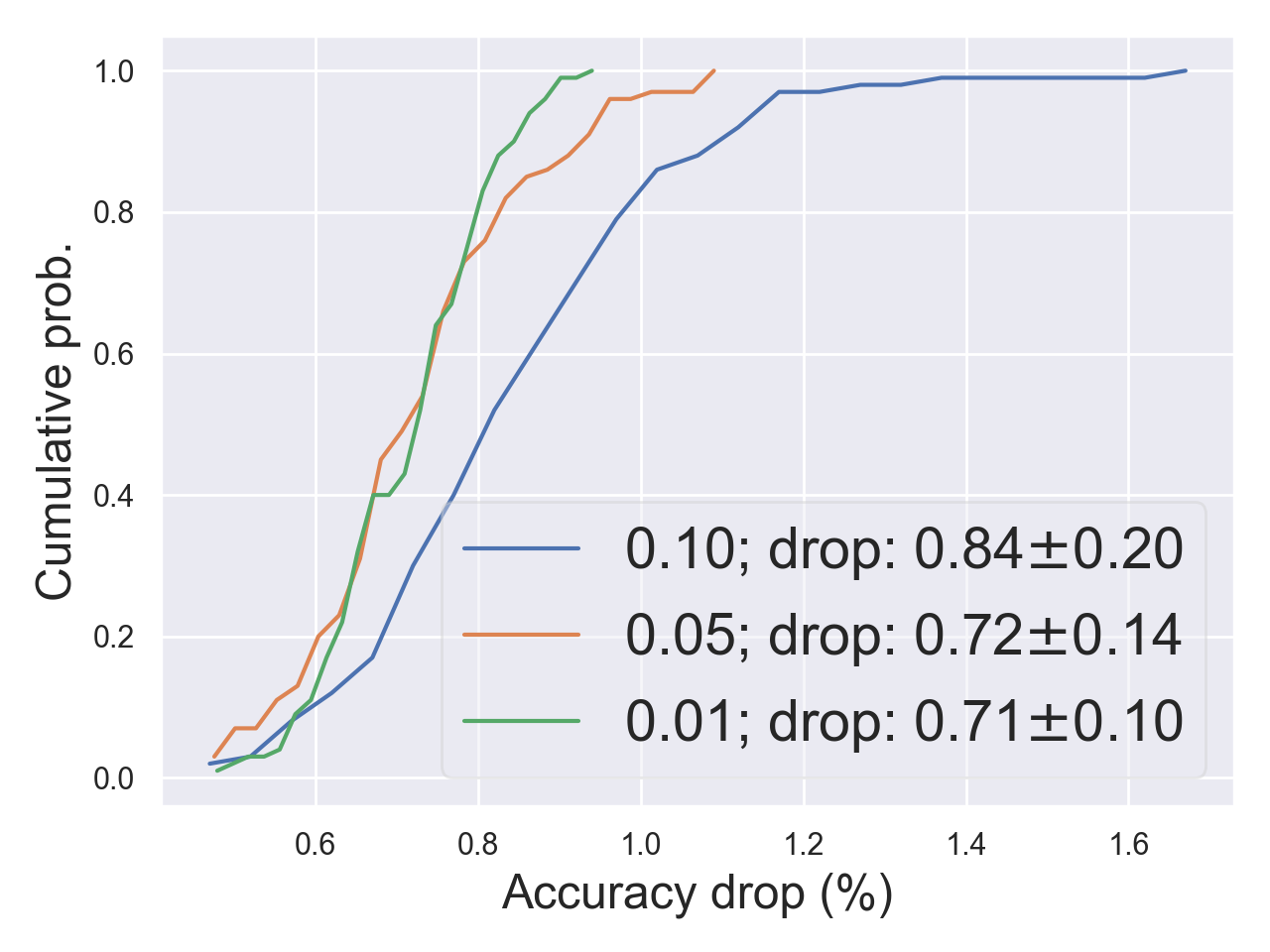}
\end{subfigure}
\begin{subfigure}[b]{\spacesfigwidth\linewidth}
\includegraphics[width=\linewidth]{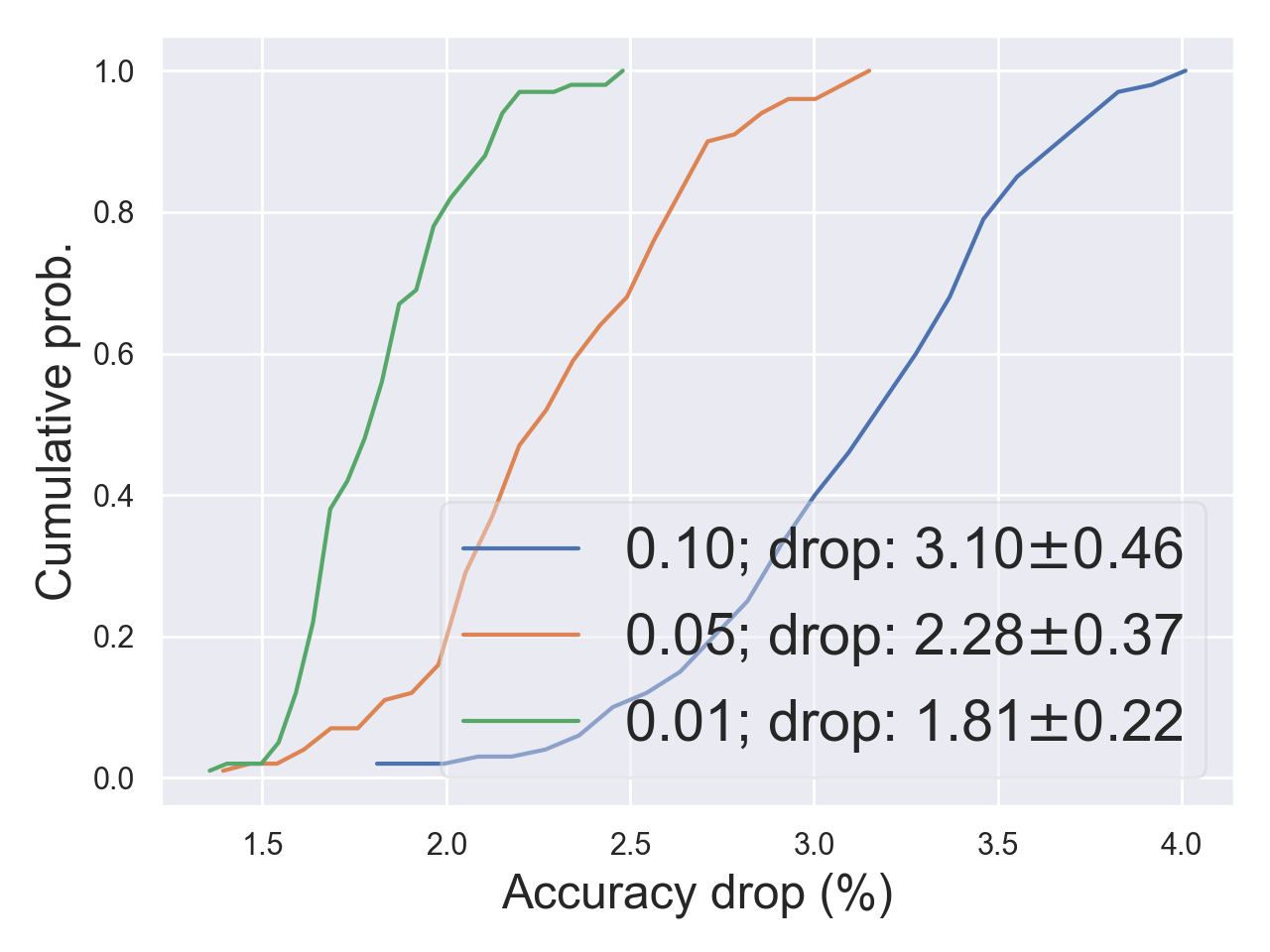}
\end{subfigure}
\begin{subfigure}[b]{\spacesfigwidth\linewidth}
\includegraphics[width=\linewidth]{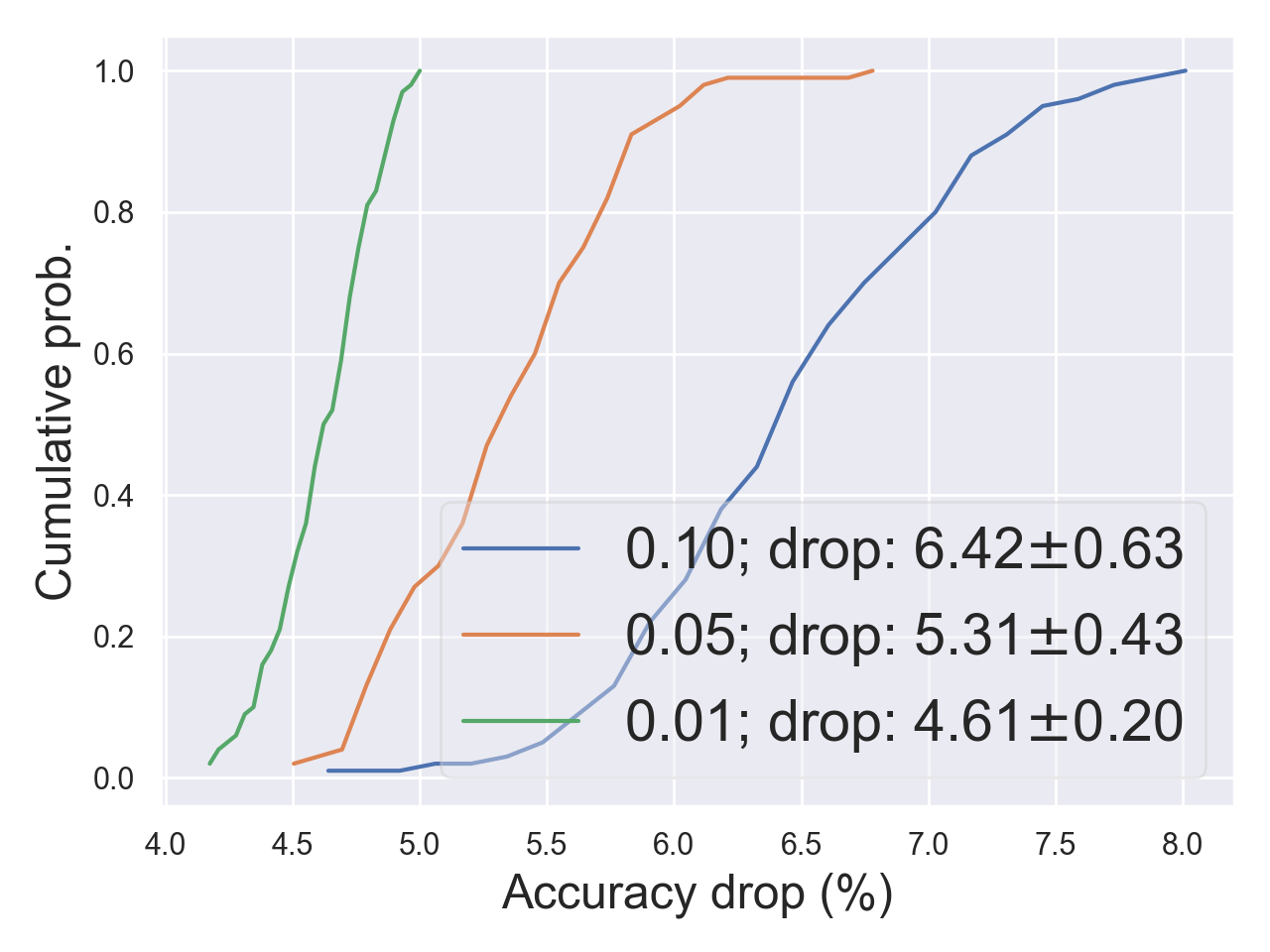}
\end{subfigure}
\begin{subfigure}[b]{\spacesfigwidth\linewidth}
\includegraphics[width=\linewidth]{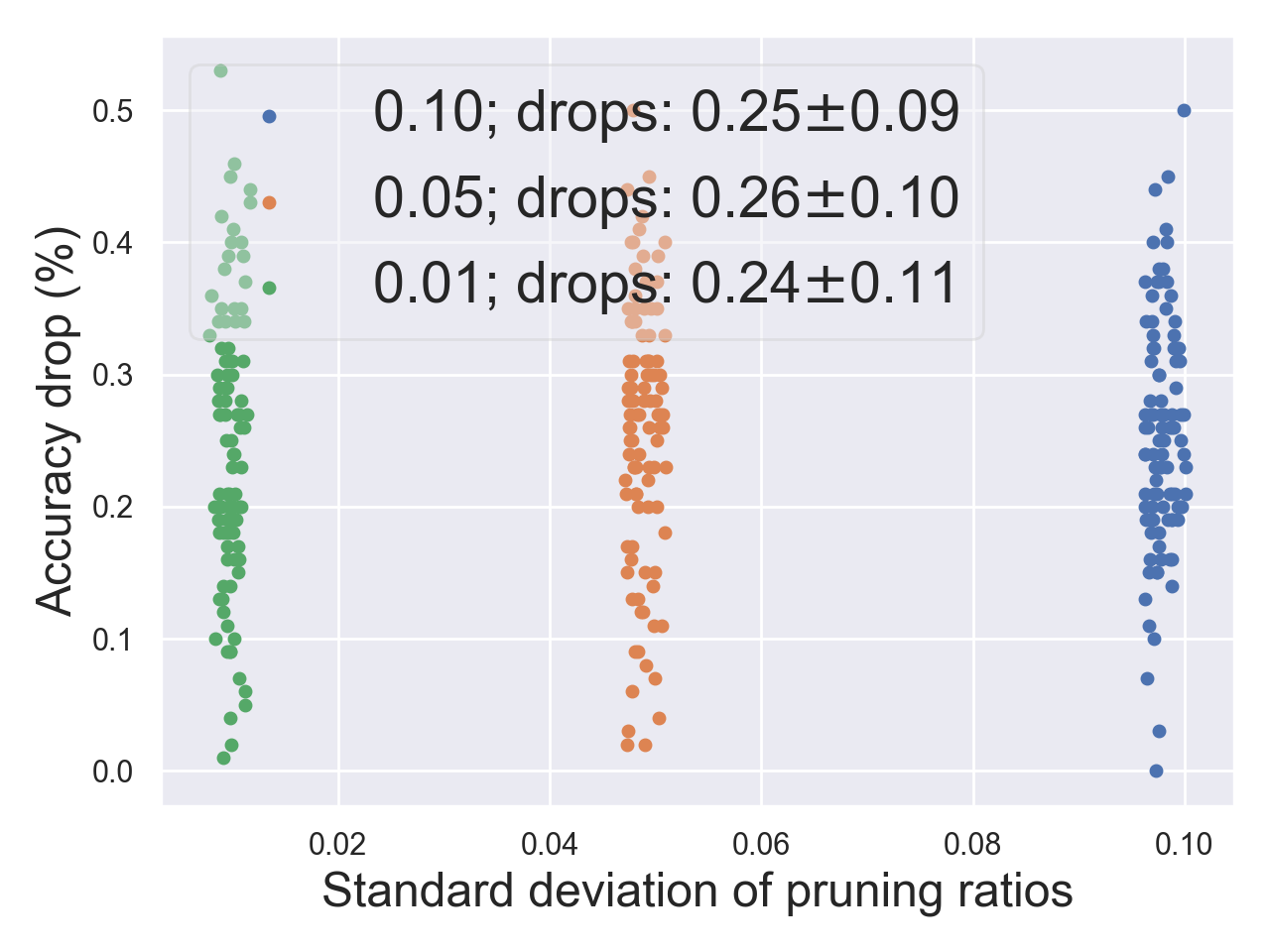}
\end{subfigure}
\begin{subfigure}[b]{\spacesfigwidth\linewidth}
\includegraphics[width=\linewidth]{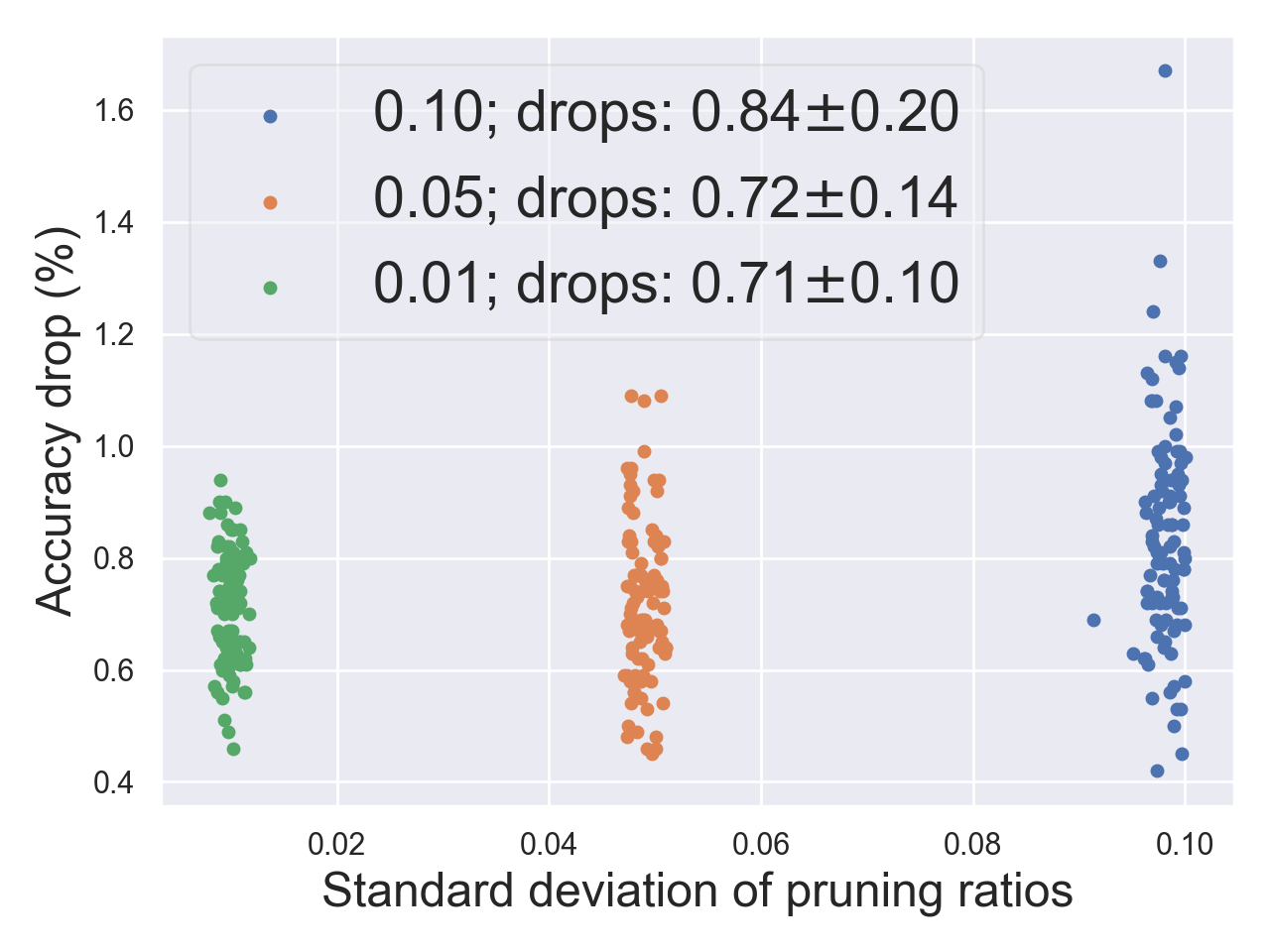}
\end{subfigure}
\begin{subfigure}[b]{\spacesfigwidth\linewidth}
\includegraphics[width=\linewidth]{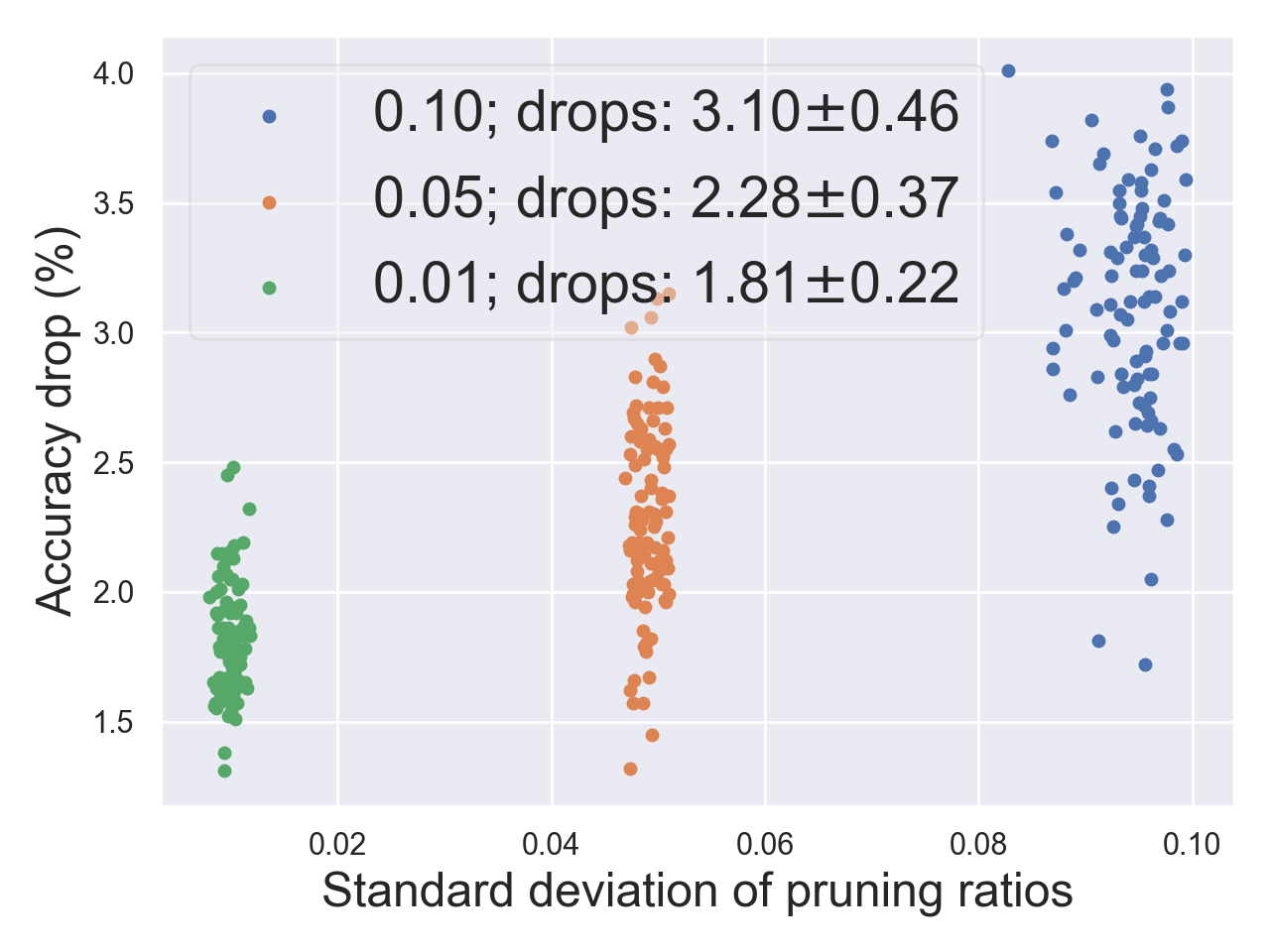}
\end{subfigure}
\begin{subfigure}[b]{\spacesfigwidth\linewidth}
\includegraphics[width=\linewidth]{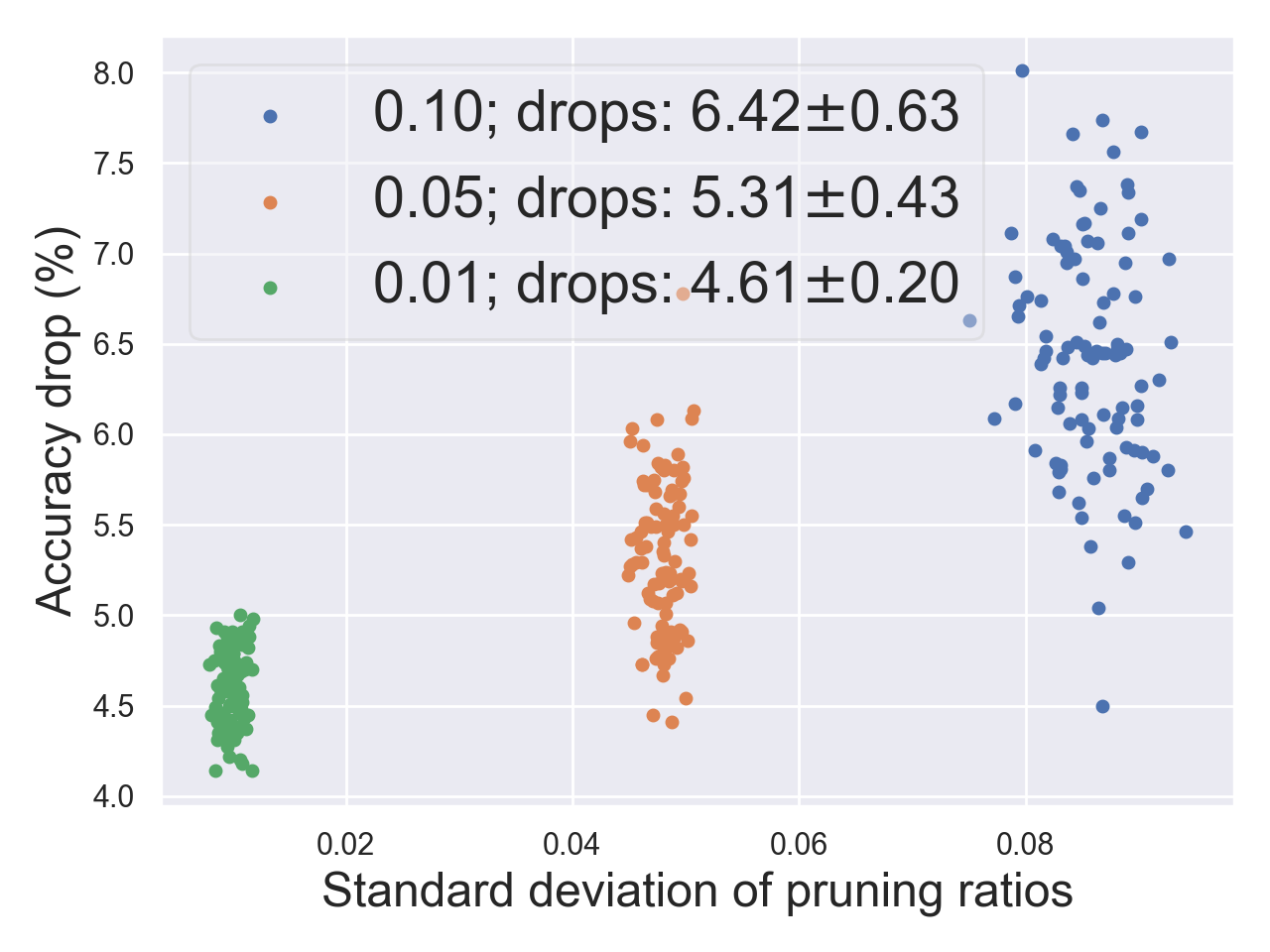}
\end{subfigure}
\begin{subfigure}[b]{\spacesfigwidth\linewidth}
\includegraphics[width=\linewidth]{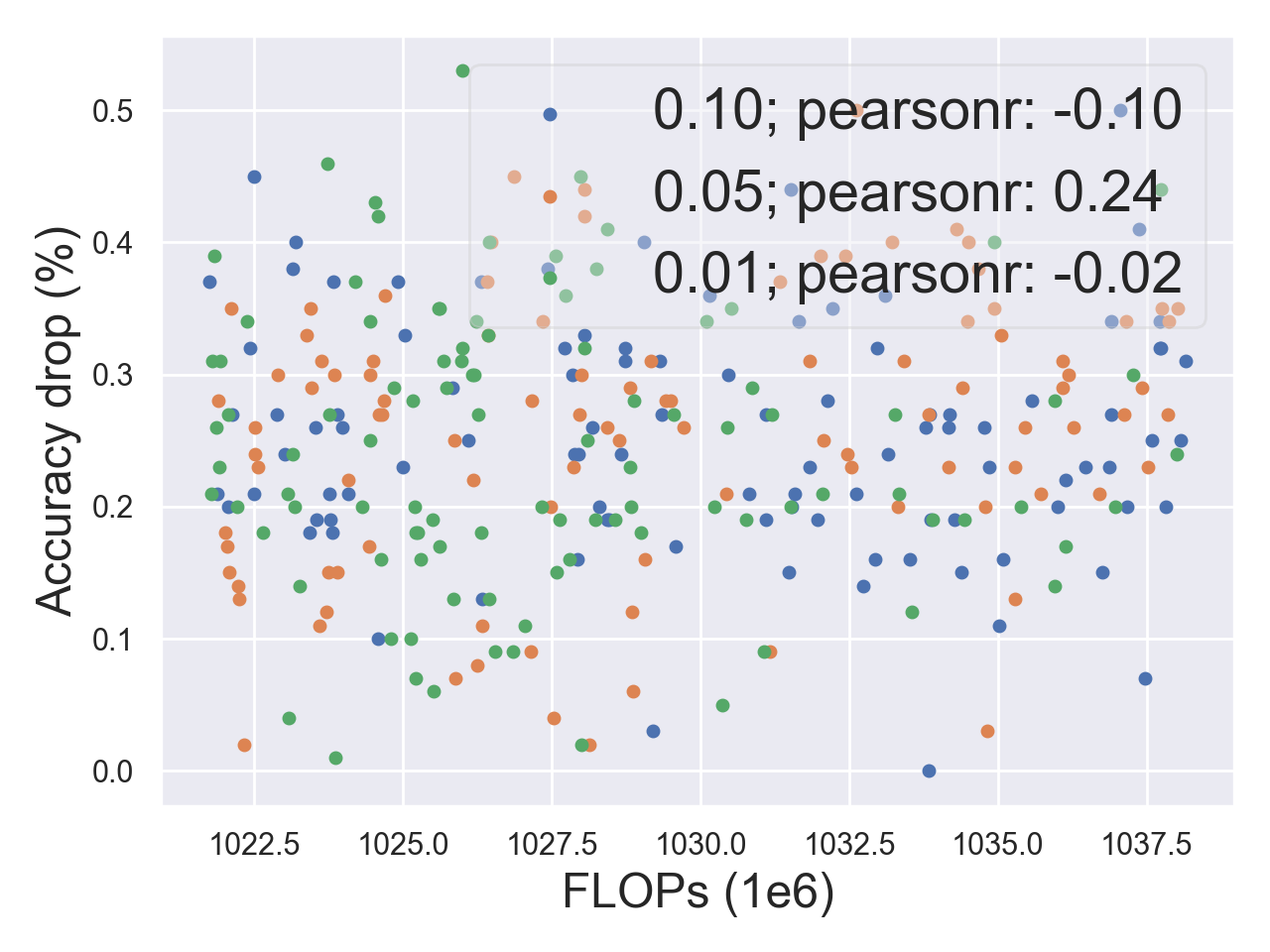}
\end{subfigure}
\begin{subfigure}[b]{\spacesfigwidth\linewidth}
\includegraphics[width=\linewidth]{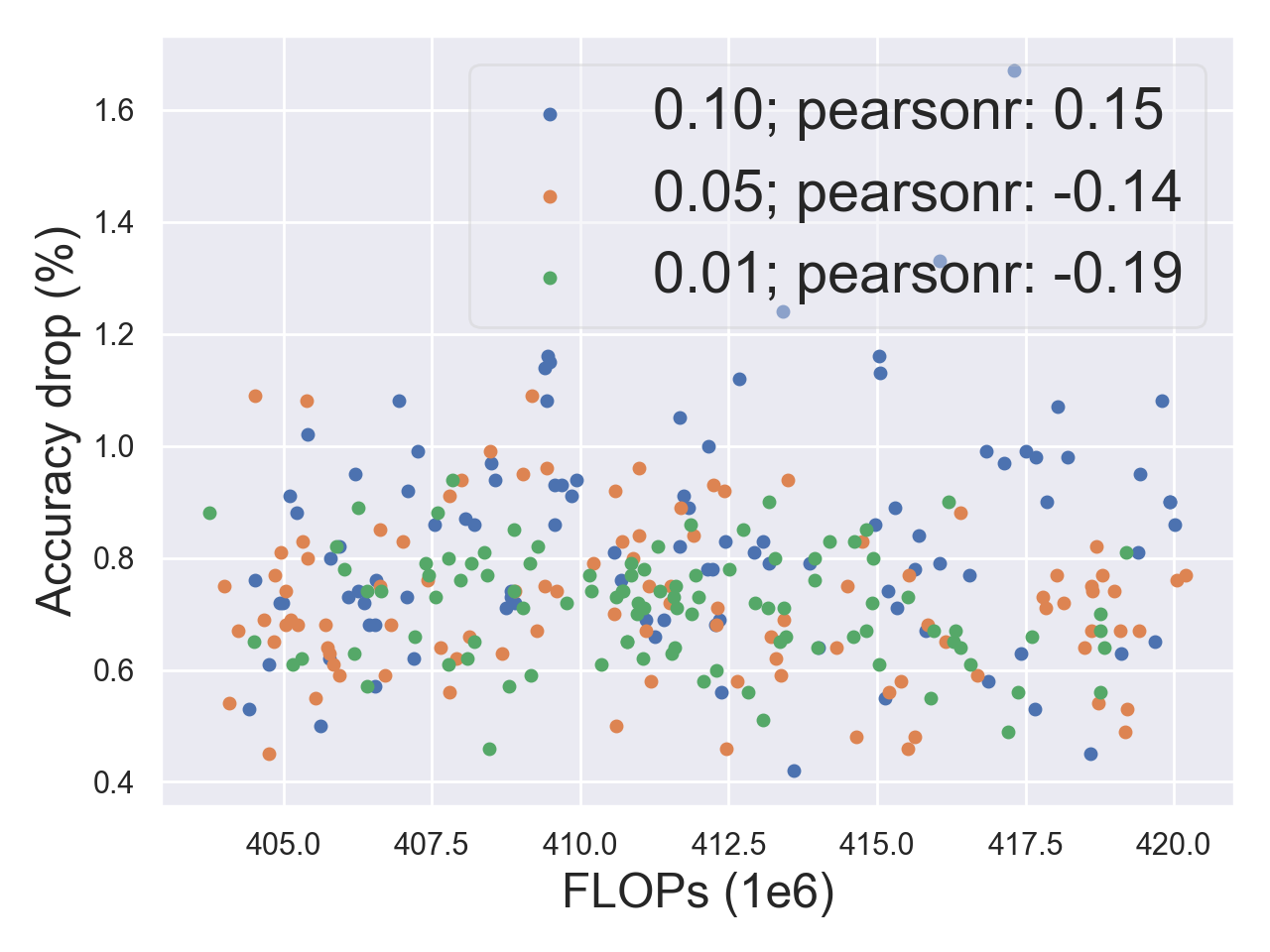}
\end{subfigure}
\begin{subfigure}[b]{\spacesfigwidth\linewidth}
\includegraphics[width=\linewidth]{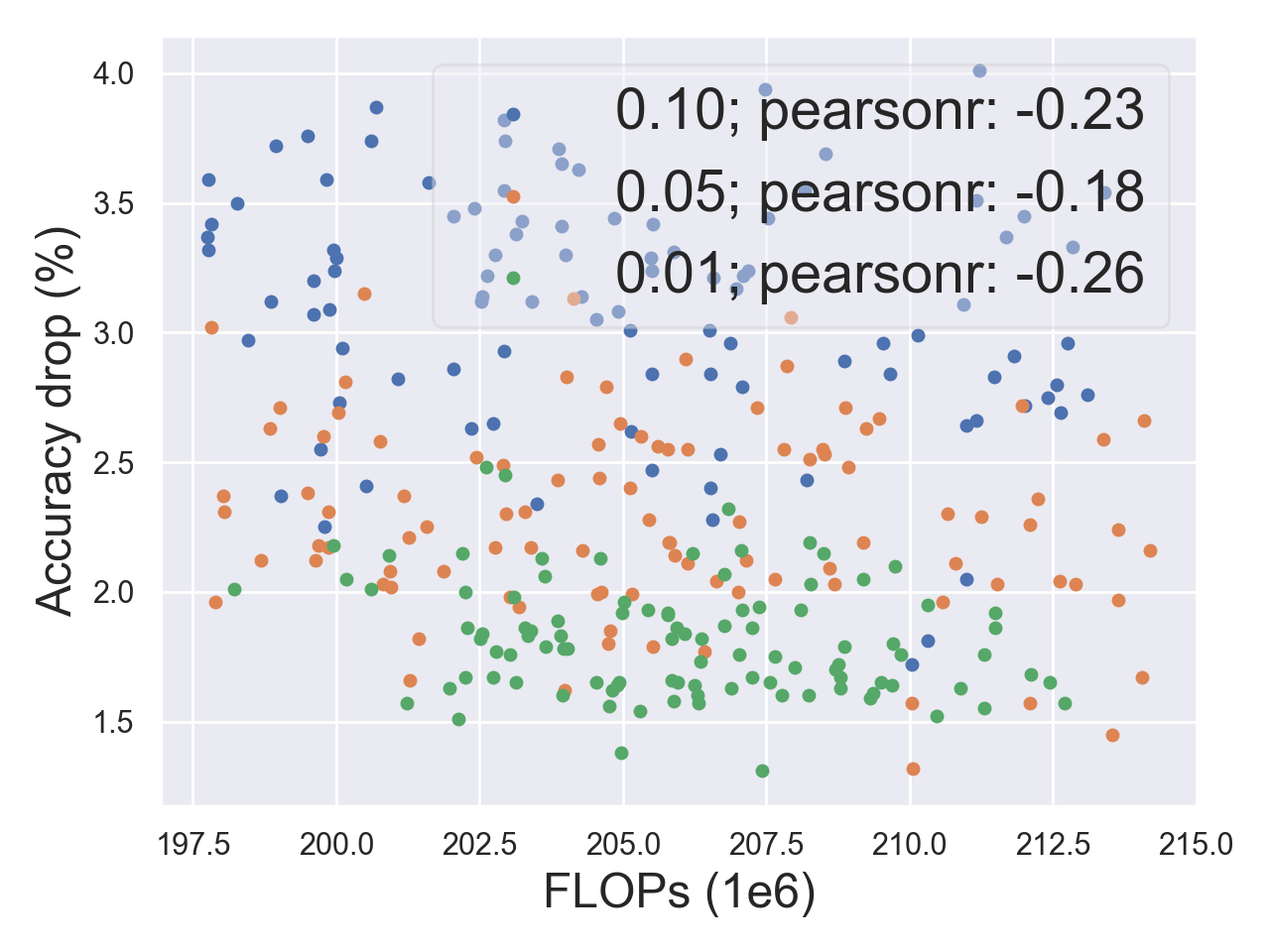}
\end{subfigure}
\begin{subfigure}[b]{\spacesfigwidth\linewidth}
\includegraphics[width=\linewidth]{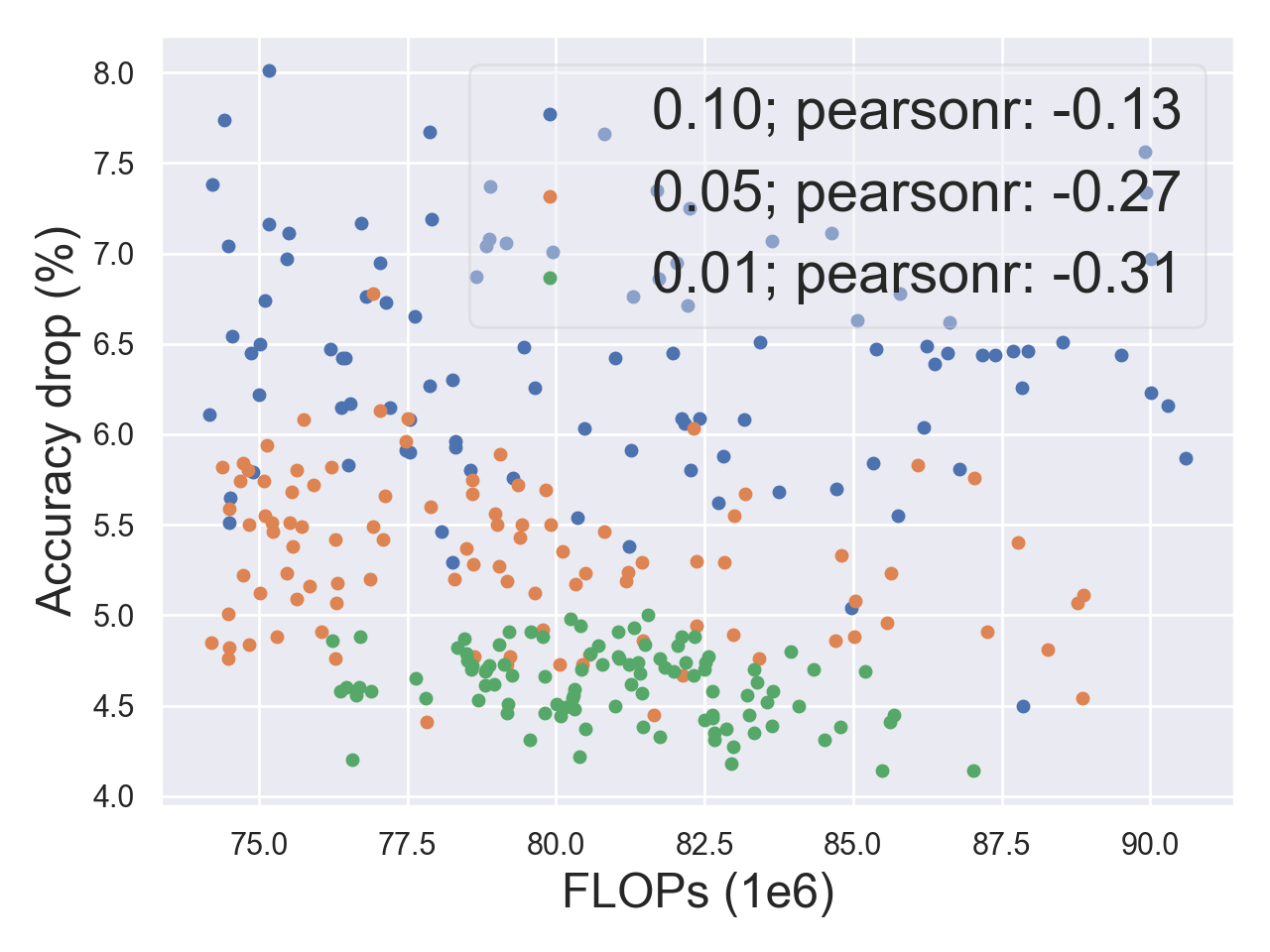}
\end{subfigure}
\begin{subfigure}[b]{\spacesfigwidth\linewidth}
\includegraphics[width=\linewidth]{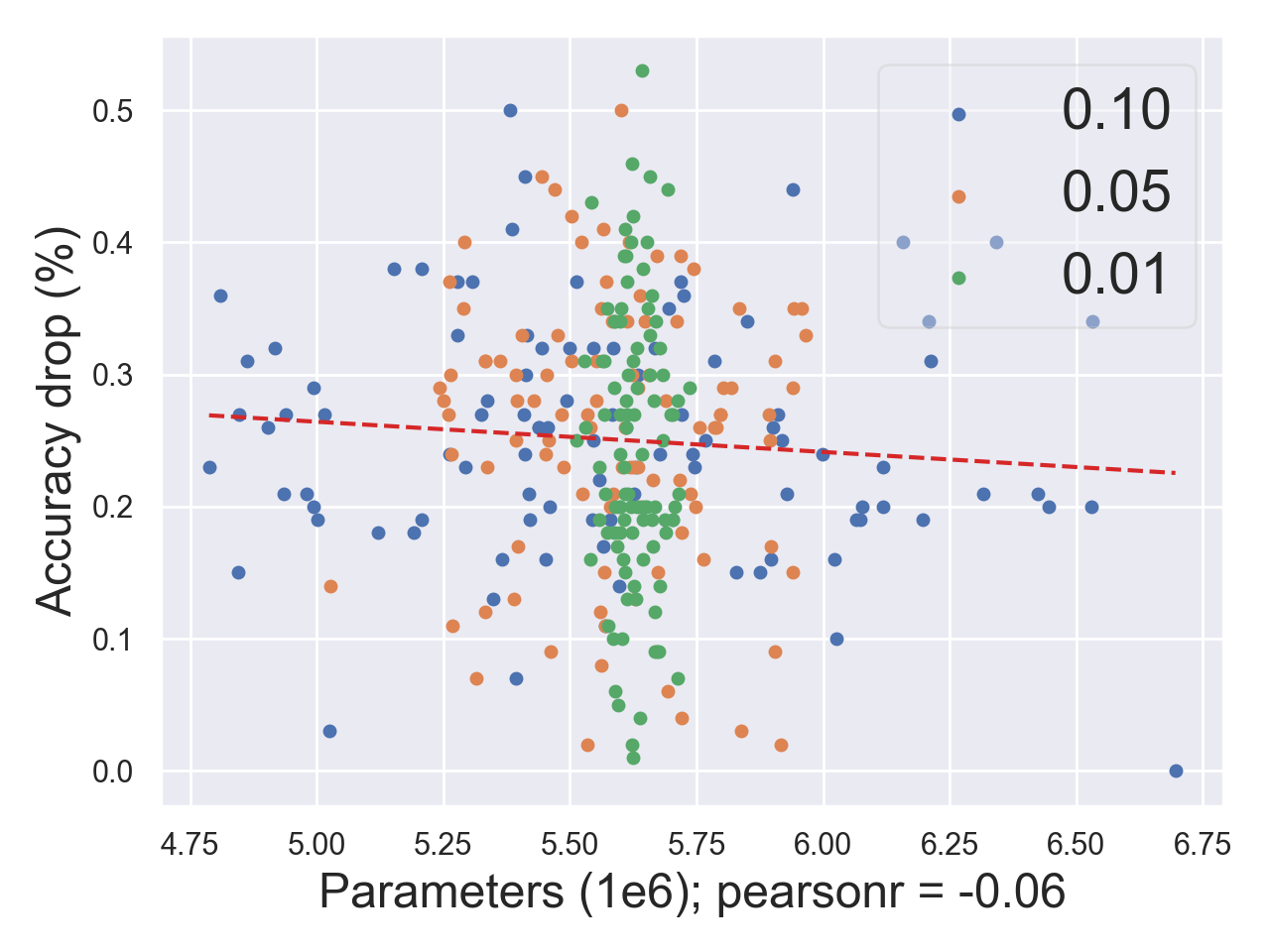}
\end{subfigure}
\begin{subfigure}[b]{\spacesfigwidth\linewidth}
\includegraphics[width=\linewidth]{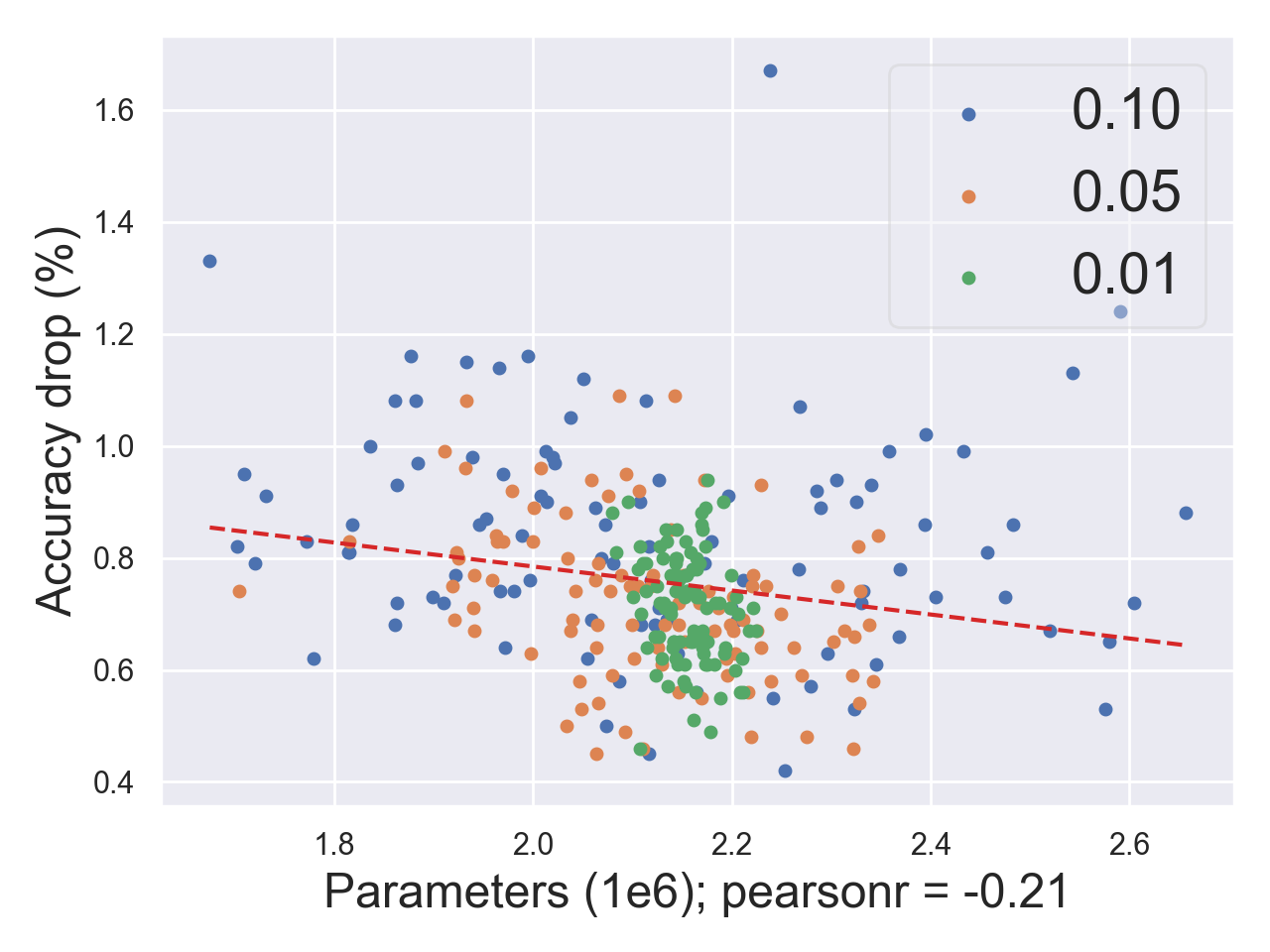}
\end{subfigure}
\begin{subfigure}[b]{\spacesfigwidth\linewidth}
\includegraphics[width=\linewidth]{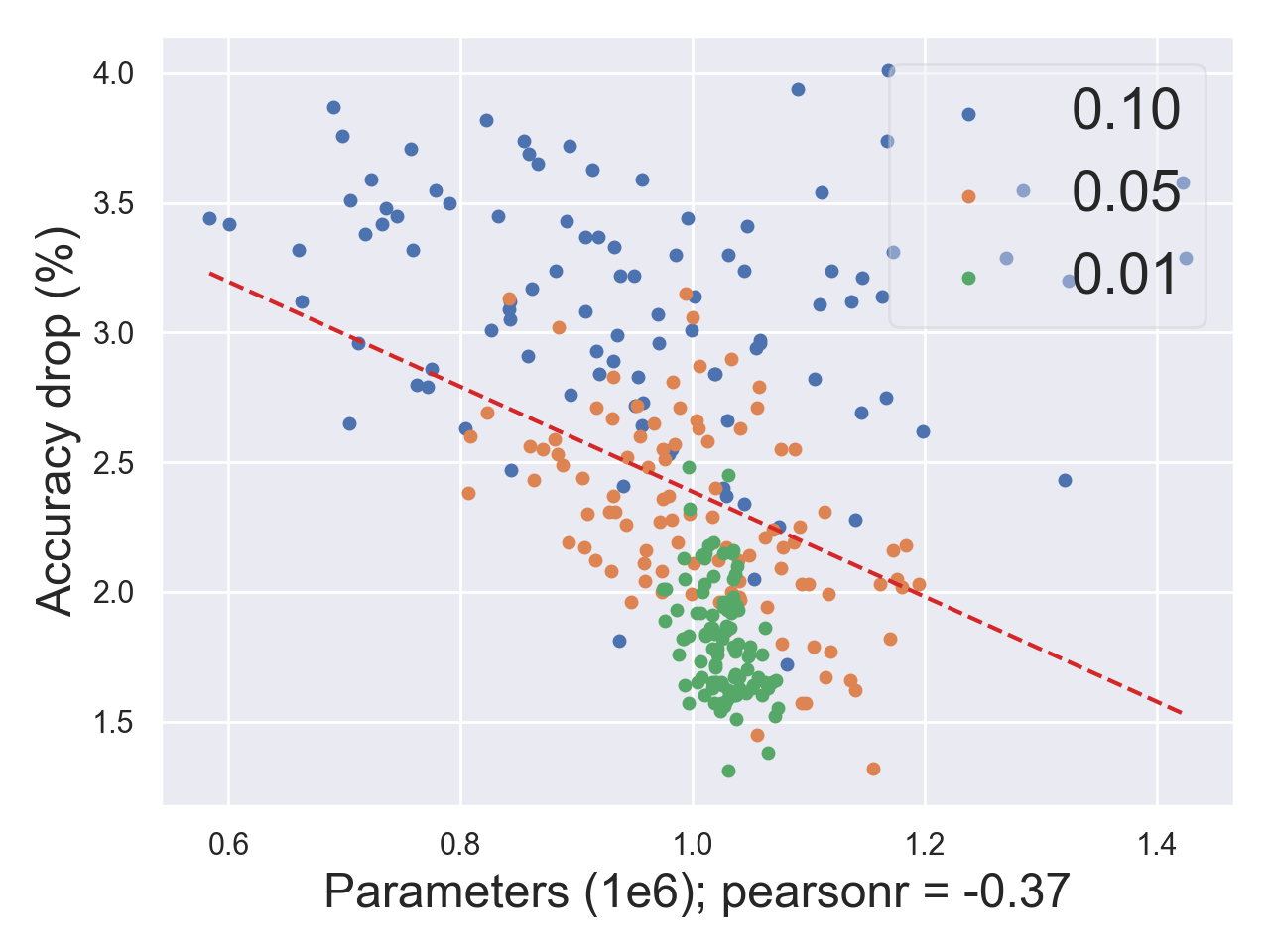}
\end{subfigure}
\begin{subfigure}[b]{\spacesfigwidth\linewidth}
\includegraphics[width=\linewidth]{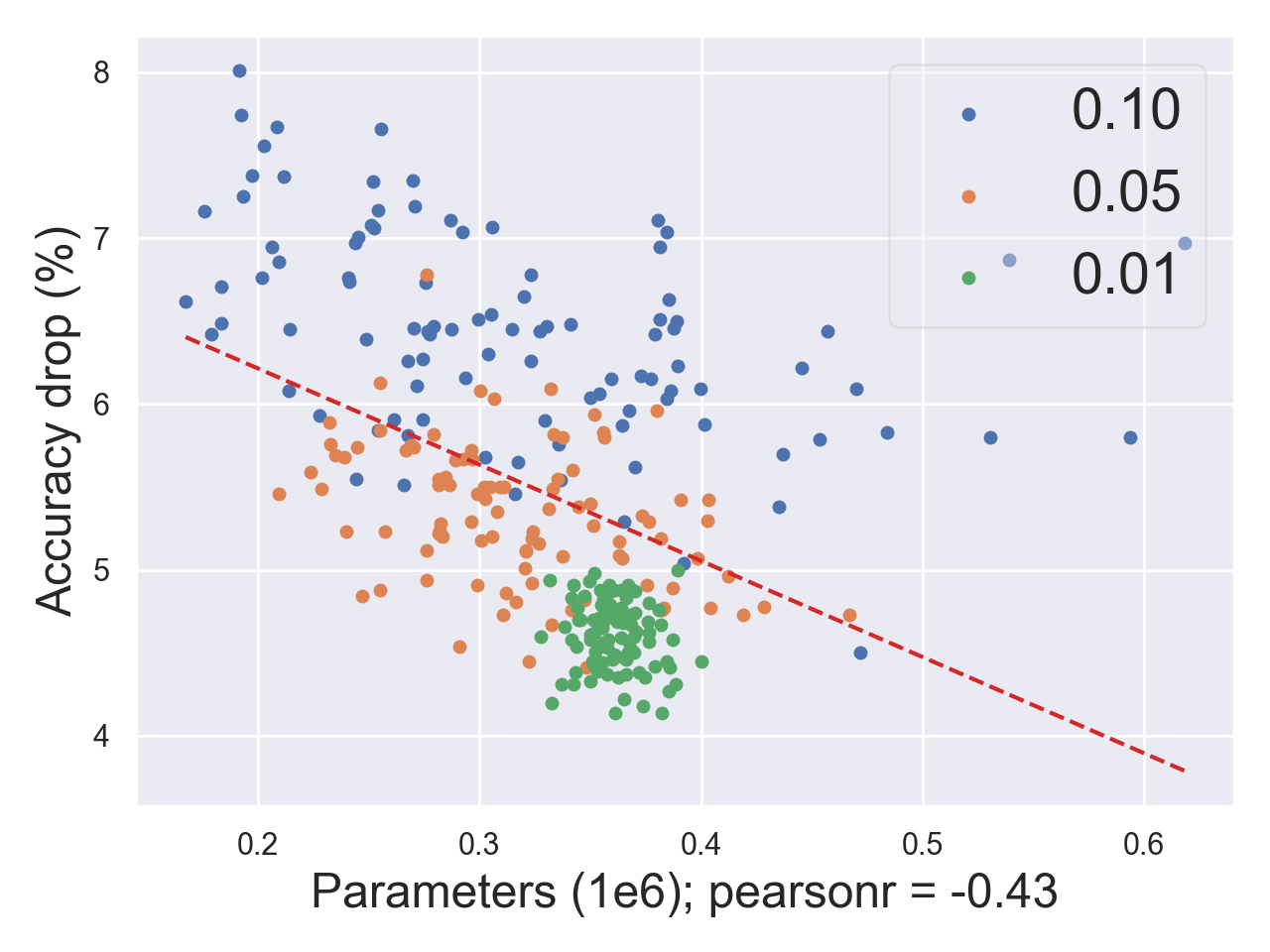}
\end{subfigure}
\begin{subfigure}[b]{\spacesfigwidth\linewidth}
\includegraphics[width=\linewidth]{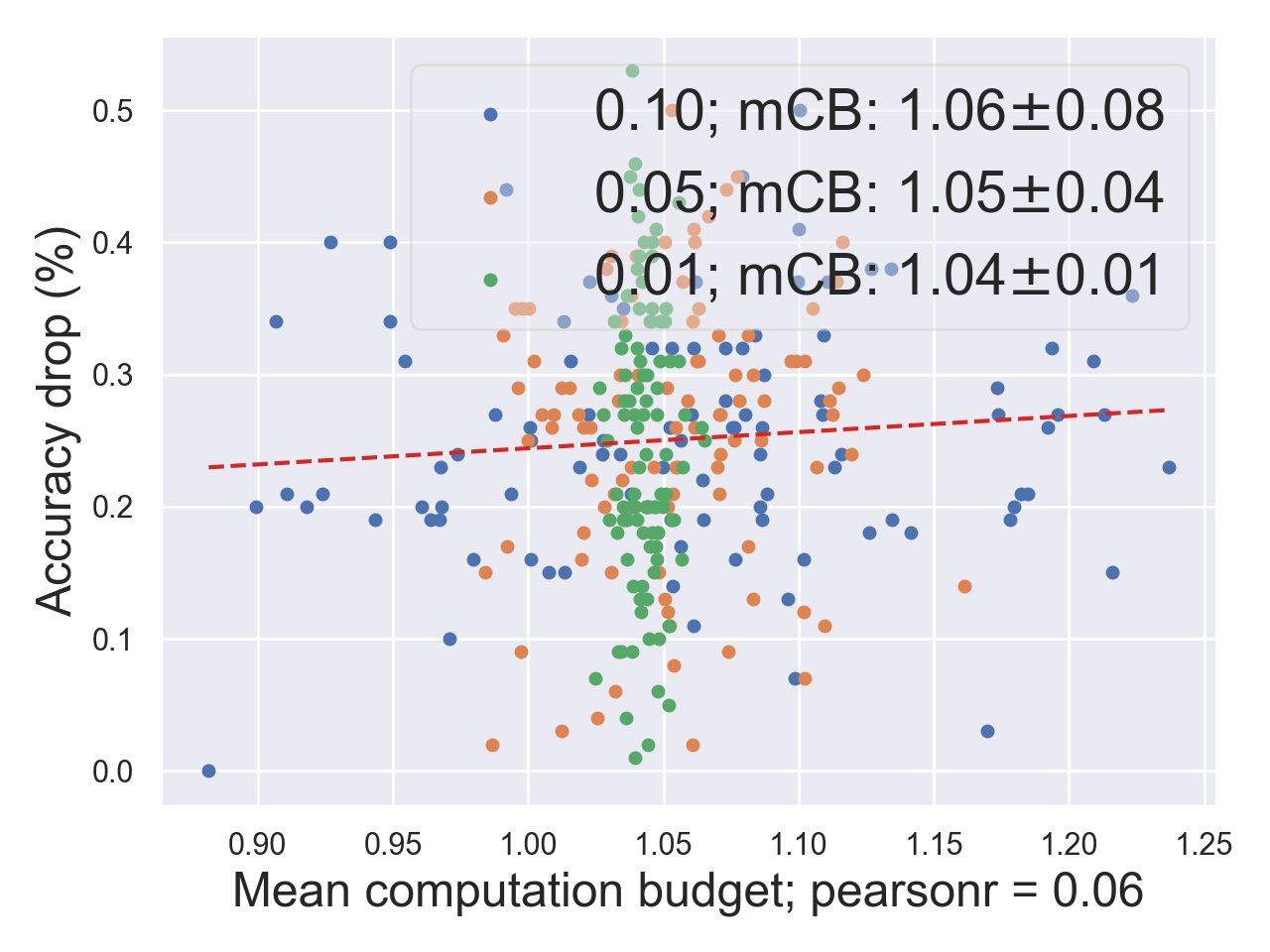}
\end{subfigure}
\begin{subfigure}[b]{\spacesfigwidth\linewidth}
\includegraphics[width=\linewidth]{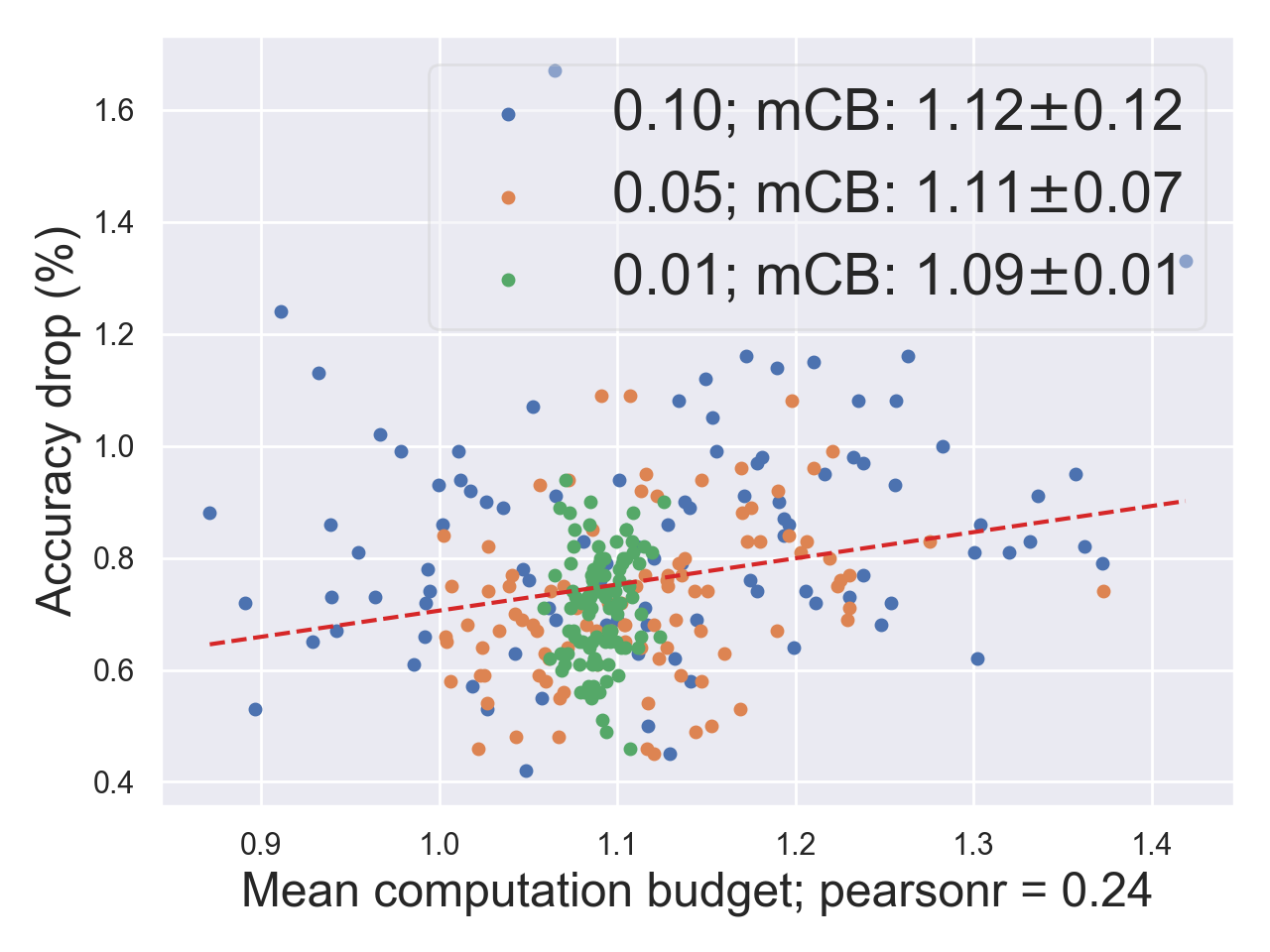}
\end{subfigure}
\begin{subfigure}[b]{\spacesfigwidth\linewidth}
\includegraphics[width=\linewidth]{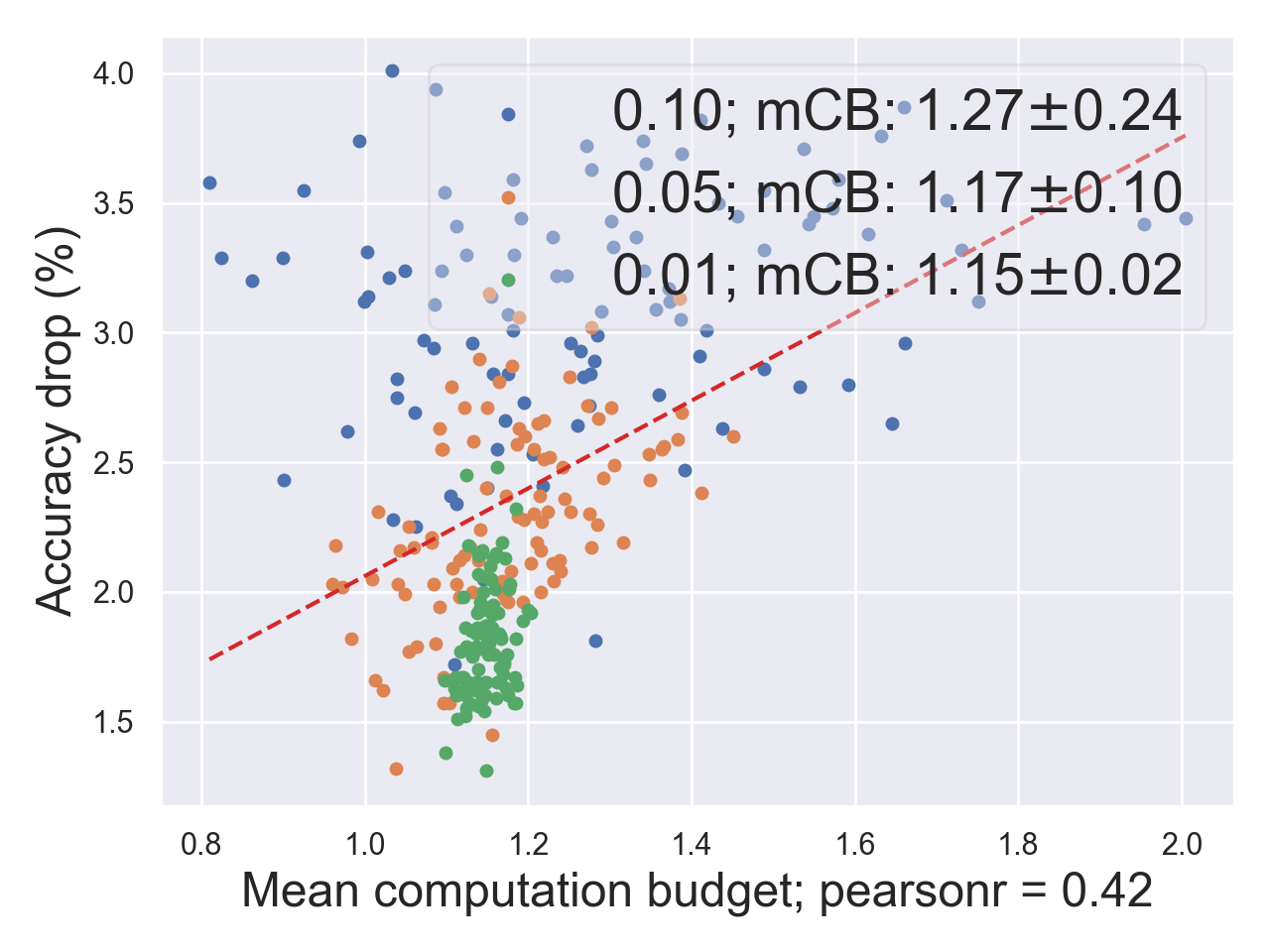}
\end{subfigure}
\begin{subfigure}[b]{\spacesfigwidth\linewidth}
\includegraphics[width=\linewidth]{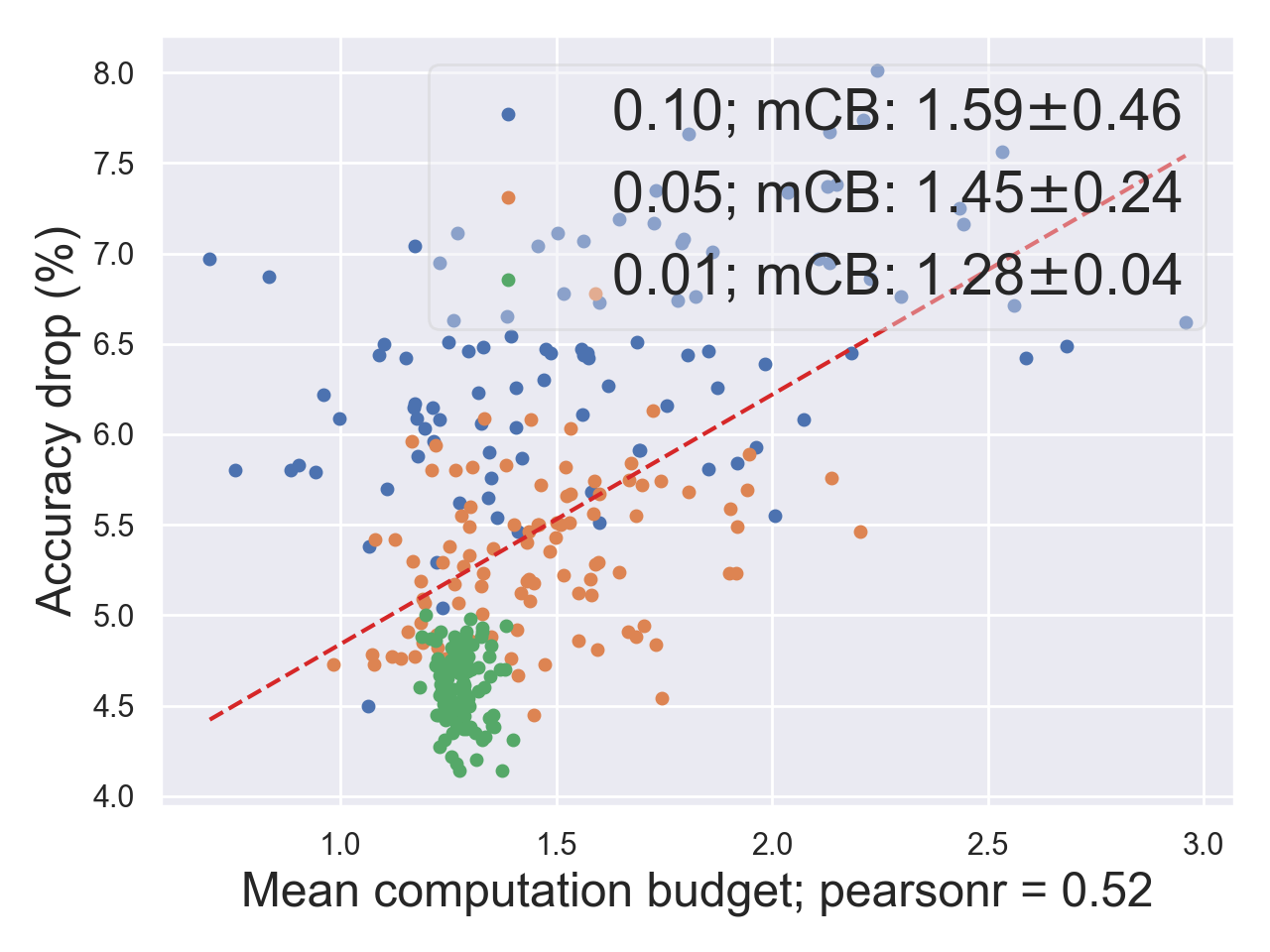}
\end{subfigure}
\begin{subfigure}[b]{\spacesfigwidth\linewidth}
\includegraphics[width=\linewidth]{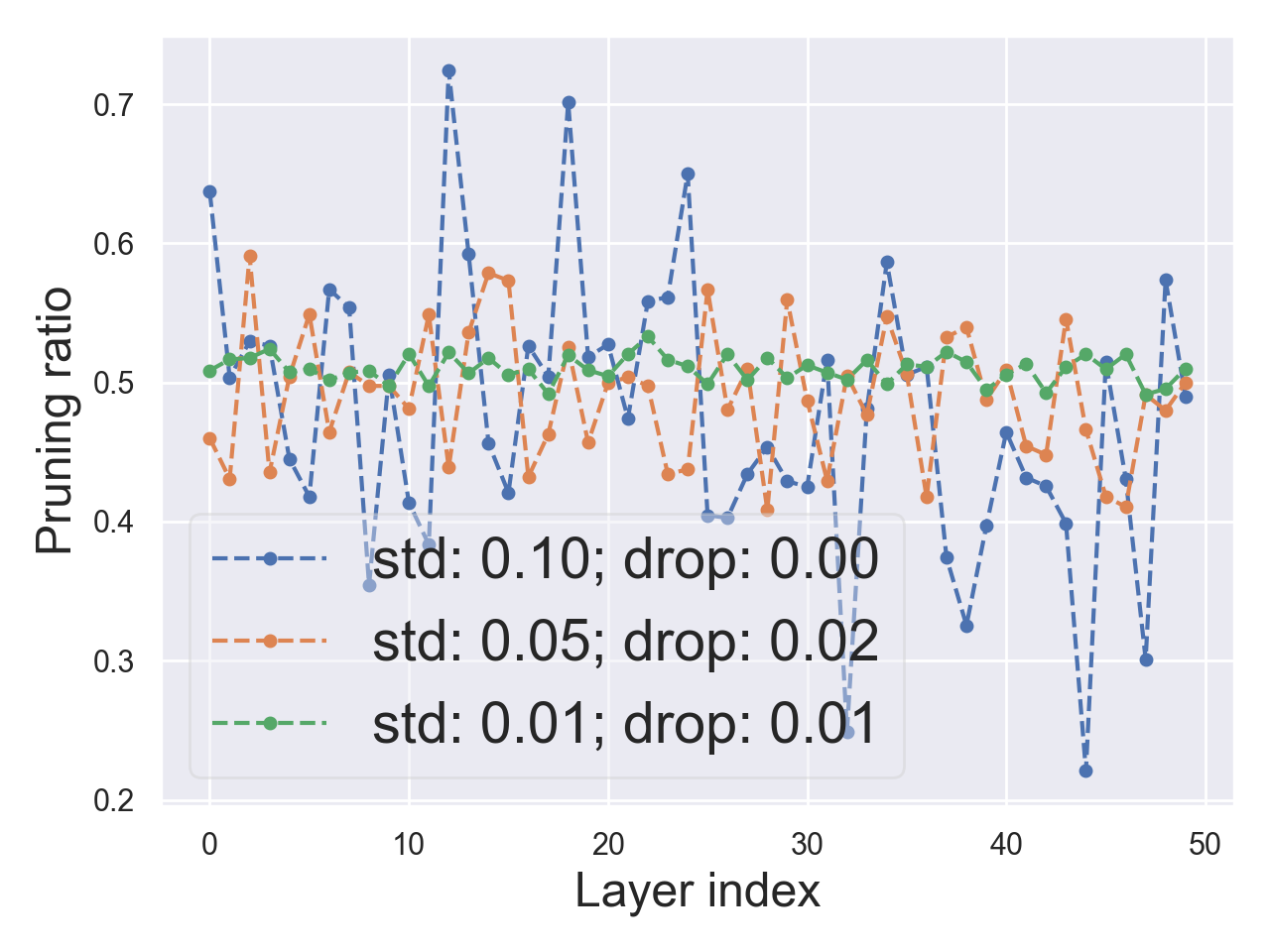}
\caption{$c_{\textnormal{flops}} = 0.25$}
\label{fig:spaces.f75}
\end{subfigure}
\begin{subfigure}[b]{\spacesfigwidth\linewidth}
\includegraphics[width=\linewidth]{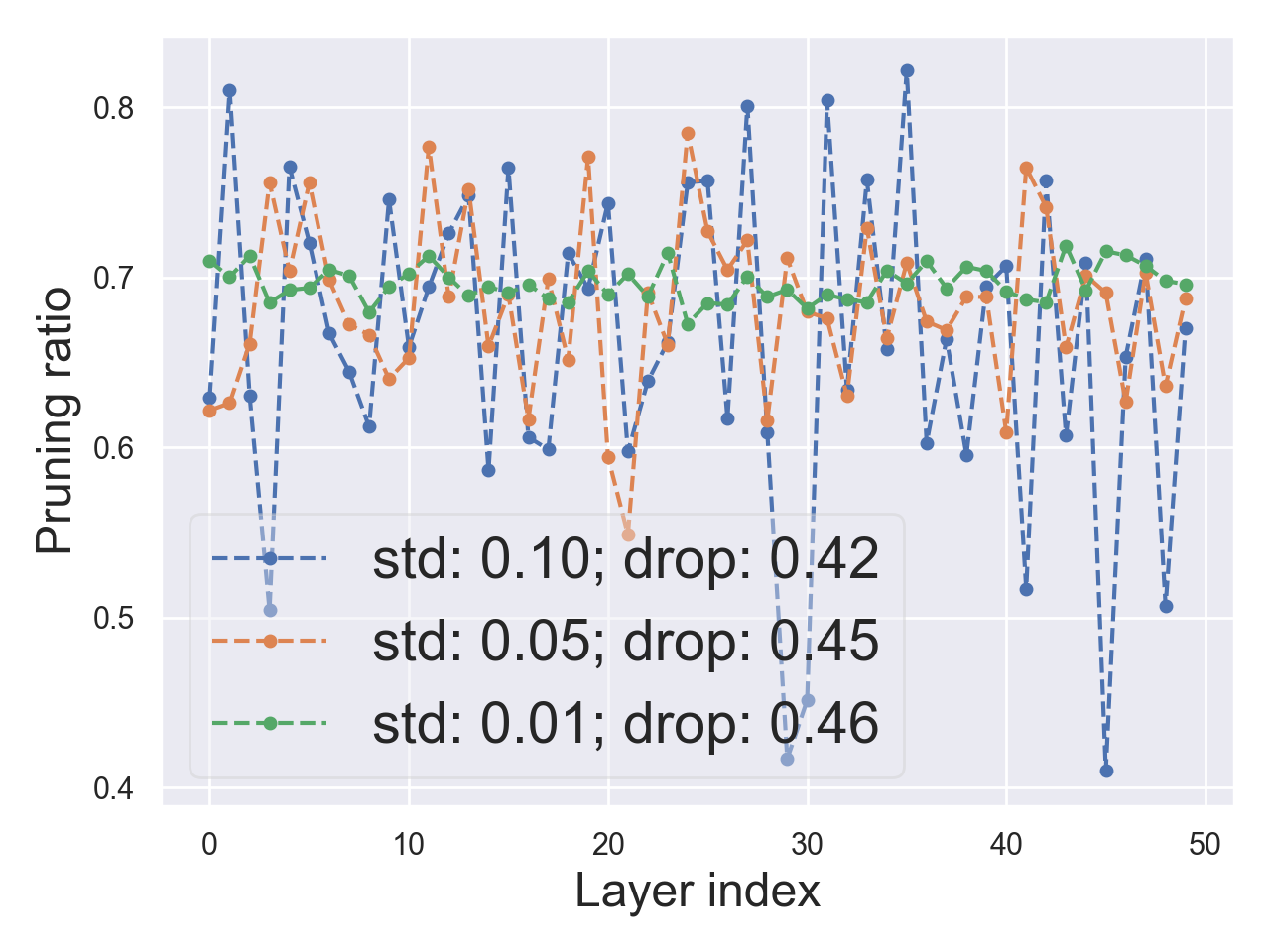}
\caption{$c_{\textnormal{flops}} = 0.1$}
\label{fig:spaces.f90}
\end{subfigure}
\begin{subfigure}[b]{\spacesfigwidth\linewidth}
\includegraphics[width=\linewidth]{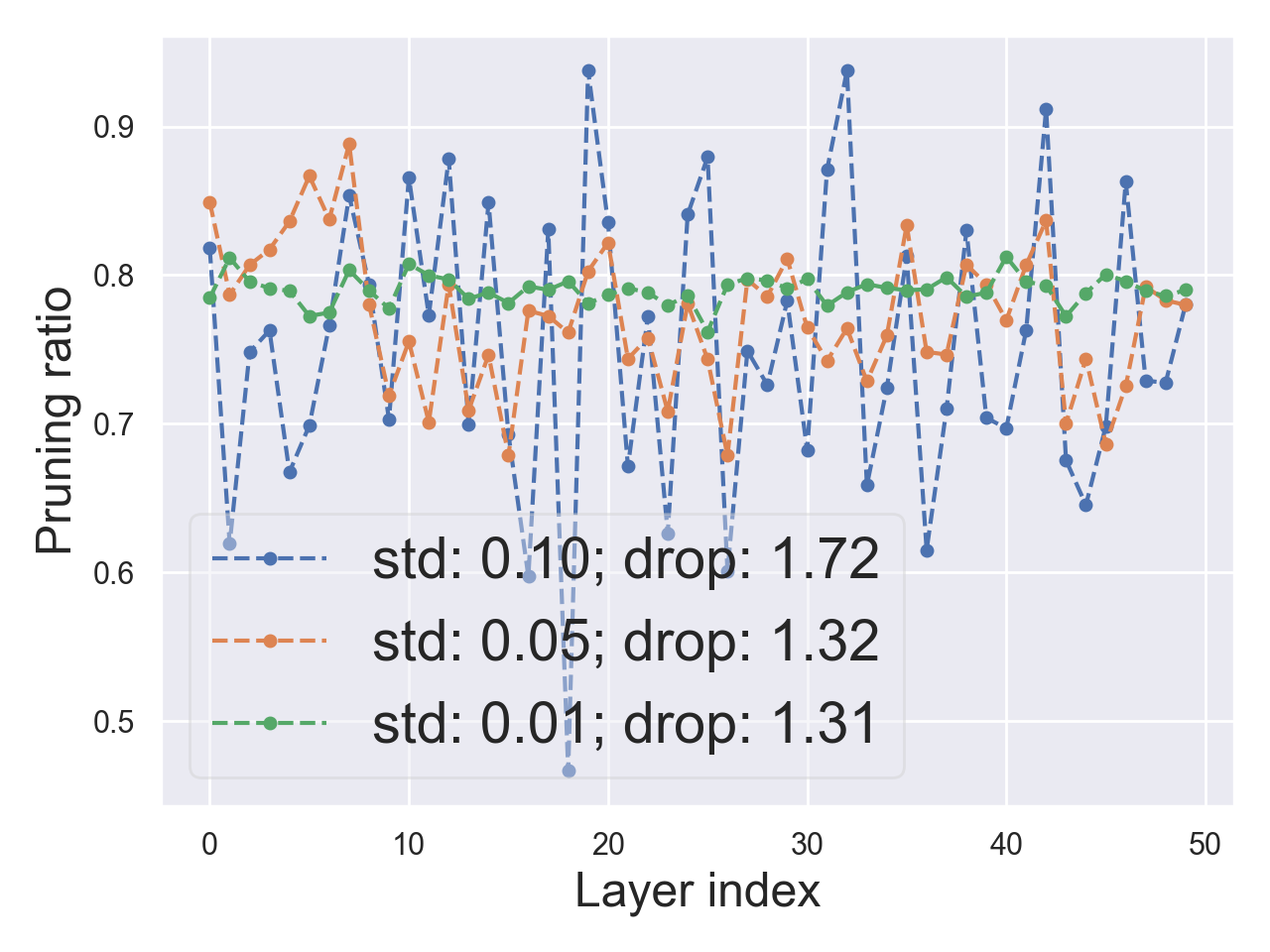}
\caption{$c_{\textnormal{flops}} = 0.05$}
\label{fig:spaces.f95}
\end{subfigure}
\begin{subfigure}[b]{\spacesfigwidth\linewidth}
\includegraphics[width=\linewidth]{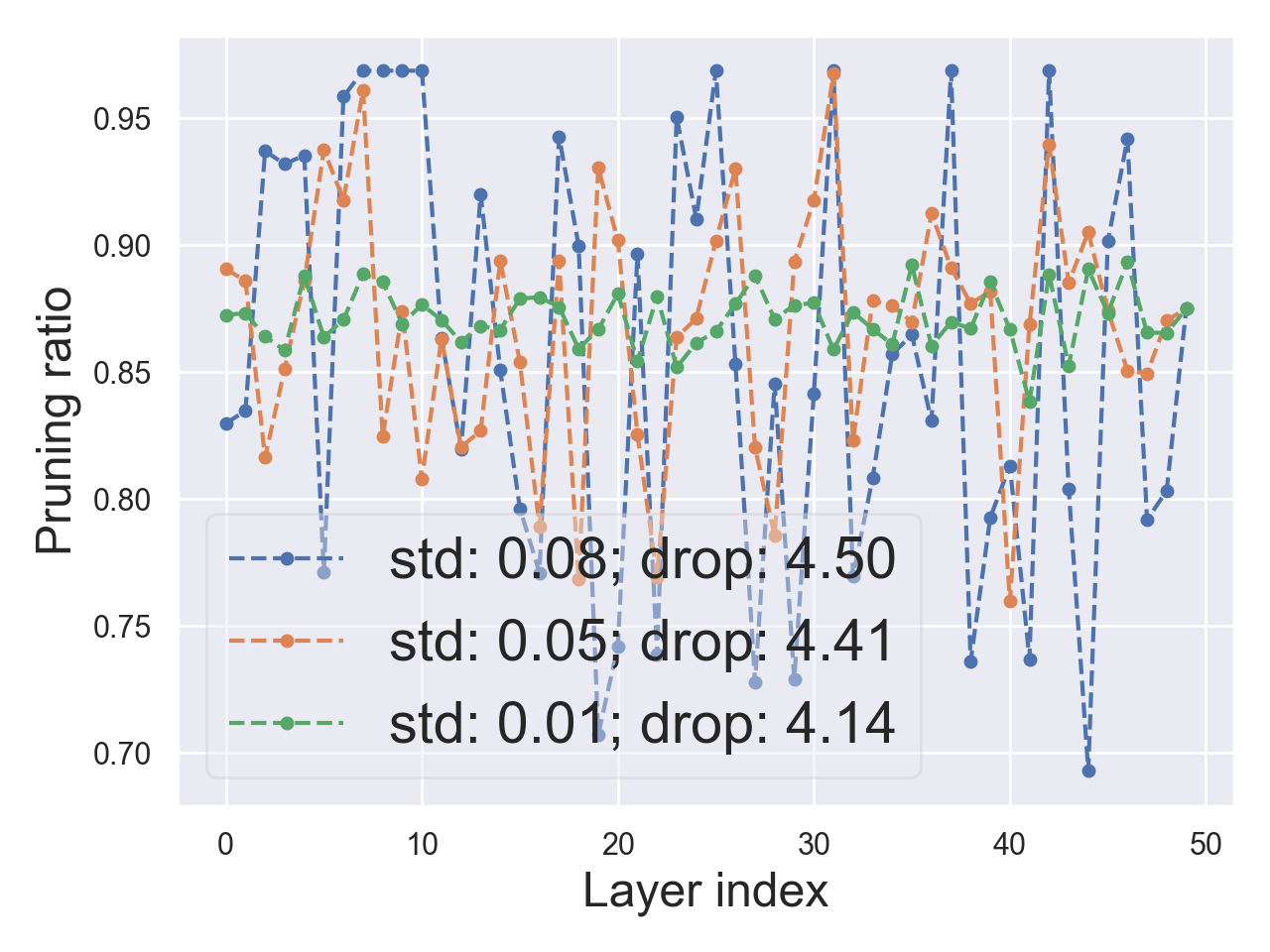}
\caption{$c_{\textnormal{flops}} = 0.02$}
\label{fig:spaces.f98}
\end{subfigure}
\caption{
\texttt{STD} pruning spaces under $c_{\textnormal{flops}} = \{0.25, 0.1, 0.05, 0.02\}$.
In the figure, we study three \texttt{STD} pruning spaces, including \texttt{STD-0.1}, \texttt{STD-0.05} and \texttt{STD-0.01} spaces.
(\emph{Row 1}) Accuracy drop EDFs on \texttt{STD} spaces.
(\emph{Row 2}) Standard deviation of $\vr$ distribution.
(\emph{Row 3}) Remaining FLOPs distribution.
(\emph{Row 4}) Remaining parameters distribution.
(\emph{Row 5}) Mean computation budget distribution.
(\emph{Row 6}) The winning pruning recipe on each \texttt{STD} pruning space.
Note that the scales of accuracy drop under different $c_{\textnormal{flops}}$ varies.
}
\label{fig:spaces}
\end{figure}

We start with the initial pruning space $\sS$. Any recipe generation strategy, such as evolutionary algorithm and reinforcement learning~\citep{he2018amc}, can be used when sampling subnetworks.
Random sampling is the most straightforward one, which randomly generates $N$ real numbers from a given range $[ 0, R ]$ as a pruning recipe $\vr$~\citep{li2020eagleeye}.\footnote{
$R$ is the largest pruning ratio applied to a layer.}

Although a random recipe that meets the requirement $c_{\textnormal{flops}}$ can be a valid one, we observe an interesting problem when sampling subnetworks from $\sS$ following~\citet{li2020eagleeye}.
This random sampling process is not effective enough, taking many times for a valid recipe, as we have a compact constraint (\ie, small $\delta$).
It is much easier to collect enough recipes when we start with an uniform pruning ratio and modify $\{r_i\}$ with $N$ small random numbers.
Therefore, we take a refinement step and come to \texttt{STD} pruning spaces.

We additionally use a constraint $c_{\textnormal{std}}$ on the standard deviation of the recipe $\vr$ and refer to the pruning spaces with this constraint as \texttt{STD} pruning spaces.
By controlling the standard deviation of $\vr$, we divide the initial space $\sS$ into \texttt{STD} subspaces.
In this work, we mainly study three \texttt{STD} pruning spaces with $c_{\textnormal{std}} =\{0.1, 0.05, 0.01\}$, denoted as \texttt{STD-0.1}, \texttt{STD-0.05} and \texttt{STD-0.01} spaces respectively.

We focus on exploring the structure aspect of winning subnetworks that have the best performance after retraining.
With our baseline setting, we sample $n = 100$ subnetworks from \texttt{STD} pruning spaces and fine-tune these subnetworks for $t = 50$ epochs.
Although better performance can be achieved when we retrain subnetworks for more epochs, we argue that fine-tuning for 50 epochs is more efficient for most settings.

\textbf{Medium-pruning-ratio regime.}
\Figref{fig:spaces.f75} (Row 1) illustrates accuracy drop EDFs for three \texttt{STD} pruning spaces in a medium-pruning-ratio regime.
As a baseline, the pruned subnetwork with an uniform pruning ratio of 0.5 reduces $74.1\%$ FLOPs (corresponds to $c_{\textnormal{flops}} = 0.259$) from the original network and has an accuracy drop of $0.3\%$ after fine-tuning for 50 epochs.
On each \texttt{STD} space, we can easily find many subnetworks (about a fraction of $60\%$) that outperform the baseline subnetwork.

The winning subnetworks in this regime have significantly different recipes (Row 6) and all of them lead to pruned subnetworks without any accuracy drop.\footnote{
A small loss of 0.02\% is ingored here, as better performance can be achieved when retrained for more epochs.}
From accuracy drop EDFs and standard deviation of $\vr$ (Row 2), the cost to find such a good subnetwork on each \std space is similar, suggesting that there is no difference between \std spaces in medium-pruning-ratio regimes.

\begin{figure}[t]
\centering
\begin{subfigure}[b]{0.45\linewidth}
\includegraphics[width=\linewidth]{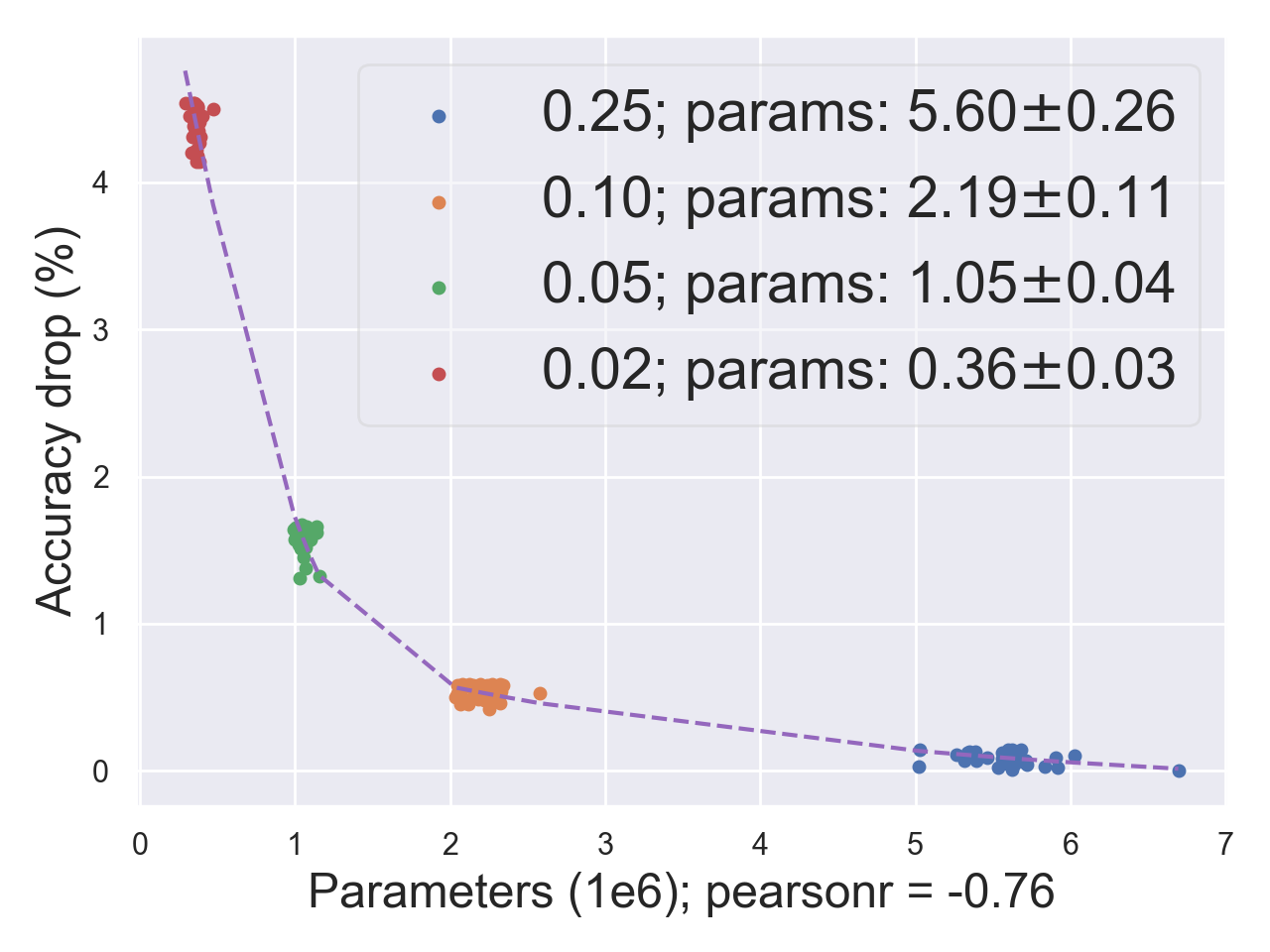}
\end{subfigure}
\begin{subfigure}[b]{0.45\linewidth}
\includegraphics[width=\linewidth]{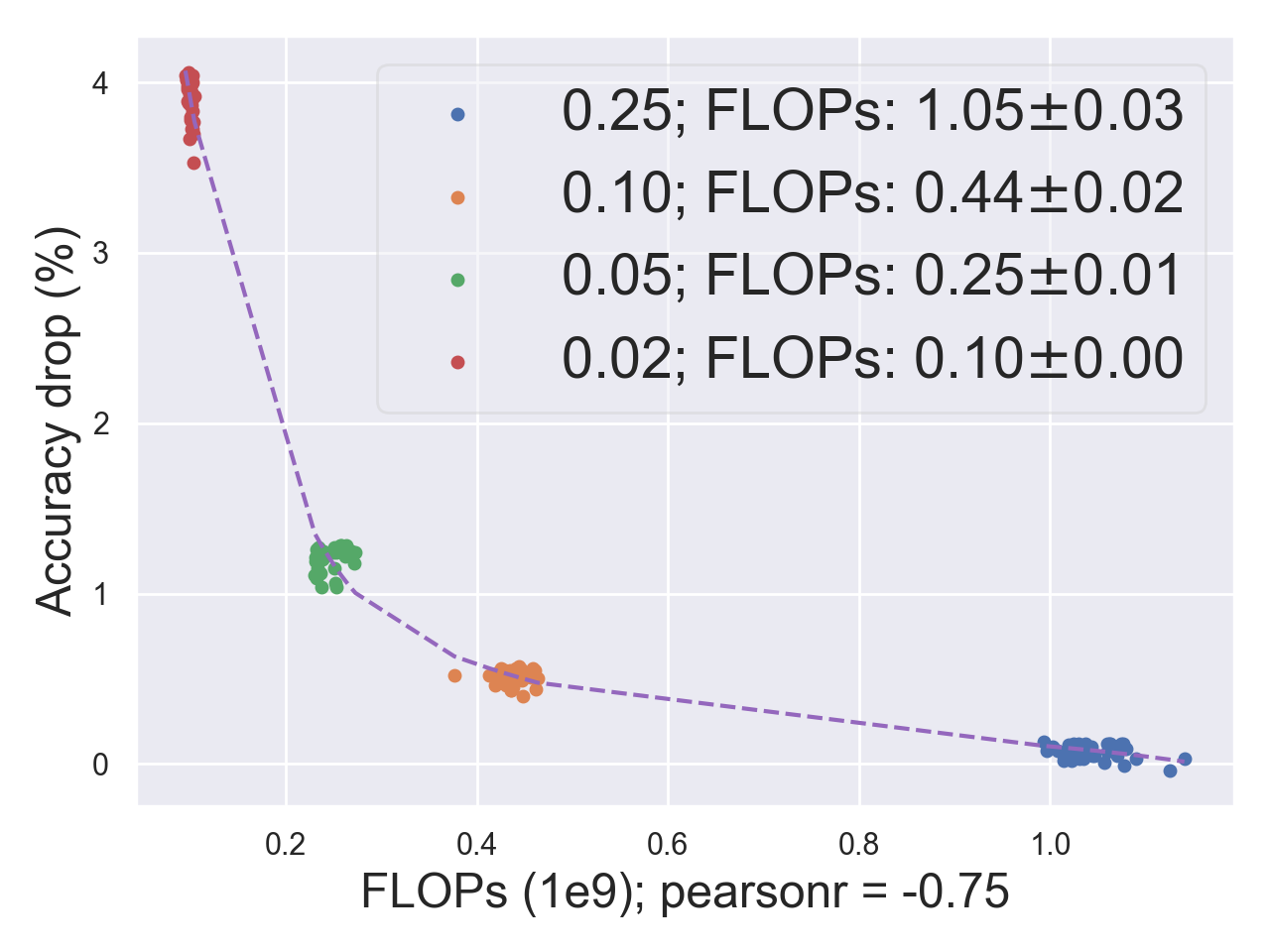}
\end{subfigure}
\caption{
We present winning subnetworks on all \std spaces with $c_{\textnormal{flops}}$ and $c_{\textnormal{params}}$ respectively.
(\emph{Left}) Parameters distribution of winning subnetworks under $c_{\textnormal{flops}} = \{0.25, 0.1, 0.05, 0.02\}$.
(\emph{Right}) FLOPs distribution of winning subnetworks under $c_{\textnormal{params}} = \{0.25, 0.1, 0.05, 0.02\}$.}
\label{fig:best40}
\end{figure}

\textbf{High-pruning-ratios regime.}
Next, we present \std spaces in high-pruning-ratio regimes, where more than 90\% FLOPs are reduced after pruning.
In Figure~\ref{fig:spaces.f90}, \ref{fig:spaces.f95} and \ref{fig:spaces.f98} (Row 1) we find that the accuracy drop EDFs for \std spaces progressively differ from each other.
From $c_{\textnormal{flops}} = 0.1$ to $c_{\textnormal{flops}} = 0.02$, \texttt{STD-0.01} spaces become higher than \texttt{STD-0.05} and \texttt{STD-0.1} spaces consistently.
In other words, \texttt{STD-0.01} space has a higher fraction of better subnetworks at every accuracy drop threshold when we prune more filters.
This clear qualitative difference suggests that the cost to find a good subnetwork on \texttt{STD-0.01} spaces is less than the costs on others.
Moreover, under an extreme constraint such as $c_{\textnormal{flops}} = 0.02$, we can easily find many subnetworks on \texttt{STD-0.01} space that outperform the winning one on \texttt{STD-0.1} space.
There might exist a winning subnetwork on \texttt{STD-0.1} space that is comparable to the best subnetworks on others, as we only sample $n = 100$ recipes in the experiments.
However, it does not alter our conclusion that \texttt{STD-0.01} space is a more efficient network pruning space across pruning regimes, especially in high-pruning-ratio regimes.

According to our experimental results, in these pruning regimes, there seems to be no shared pattern from the perspective of winning subnetwork structures.
However, we observed an interesting trend that good pruning recipes always have a small standard deviation, although those with a large standard deviation can also produce good subnetworks with a small probability.
Upon this discovery, one might ask: \emph{What makes these recipes with a small standard deviation outperform others?}

\textbf{Distribution comparison.}
In~\Figref{fig:spaces} we present FLOPs distribution (Row 3) and parameter bucket distribution (Row 4).
A pattern emerges: the amount of remaining parameters of subnetworks on \texttt{STD-0.01} spaces is much closer to a certain number than those on other \std spaces in each pruning regime.
Since we produce pruning recipes $\vr$ with a constraint on FLOPs, we have the following conjecture:

\textbf{Conjecture 1.}
\emph{
(\textbf{a}) Given a target FLOPs reduction, there exists an optimal parameter bucket for pruning filters;
(\textbf{b}) Subnetworks that have the optimal parameter bucket will be the winning ones on $\sS$;
(\textbf{c}) A good structure of pruned subnetwork guarantees an approximately optimal parameter bucket under its FLOPs.}

With Conjecture 1, we present the best 40 subnetworks on all \std spaces under each $c_{\textnormal{flops}}$ in~\Figref{fig:best40} (left).
Note that the 40 winning subnetworks have different standard deviations of $\vr$, which empirically proves our Conjecture 1c.

\textbf{Constraint on parameter bucket.}
Consider a simple question: \emph{What kind of subnetworks will be winning ones under a constraint on parameter bucket?}
Next, we present \std pruning spaces with a constraint on parameter bucket $c_{\textnormal{params}}$ in~\Figref{fig:spaces_params}.
We observe a consistent trend that \texttt{STD-0.01} space is higher than others in high-pruning-ratio regimes (\eg, $c_{\textnormal{params}} = \{0.05, 0.02\}$).
Similarly, in~\Figref{fig:best40} (right), we present the best 40 subnetworks on all \std spaces under each $c_{\textnormal{params}}$ and have the following conjecture:

\textbf{Conjecture 2.}
\emph{
(\textbf{a}) Given a target parameter bucket reduction, there exists an optimal FLOPs for pruning filters;
(\textbf{b}) Subnetworks that have the optimal FLOPs will be winning ones on $\sS$;
(\textbf{c}) A good structure of pruned subnetwork guarantees an approximately optimal FLOPs under its parameter bucket.}

\def\spacesparamsfigwidth{0.245}
\begin{figure}[t]
\centering
\begin{subfigure}[b]{\spacesparamsfigwidth\linewidth}
\includegraphics[width=\linewidth]{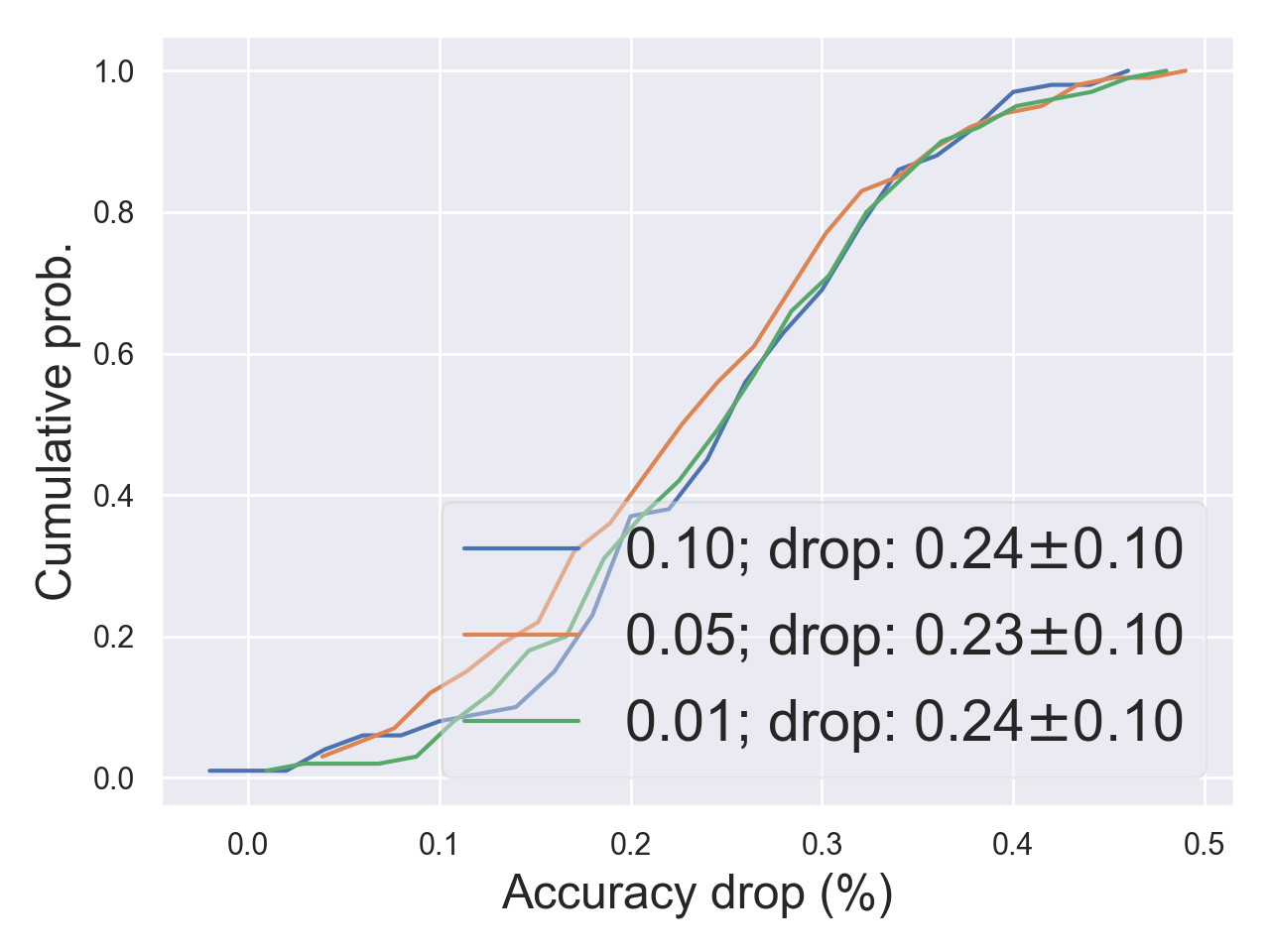}
\end{subfigure}
\begin{subfigure}[b]{\spacesparamsfigwidth\linewidth}
\includegraphics[width=\linewidth]{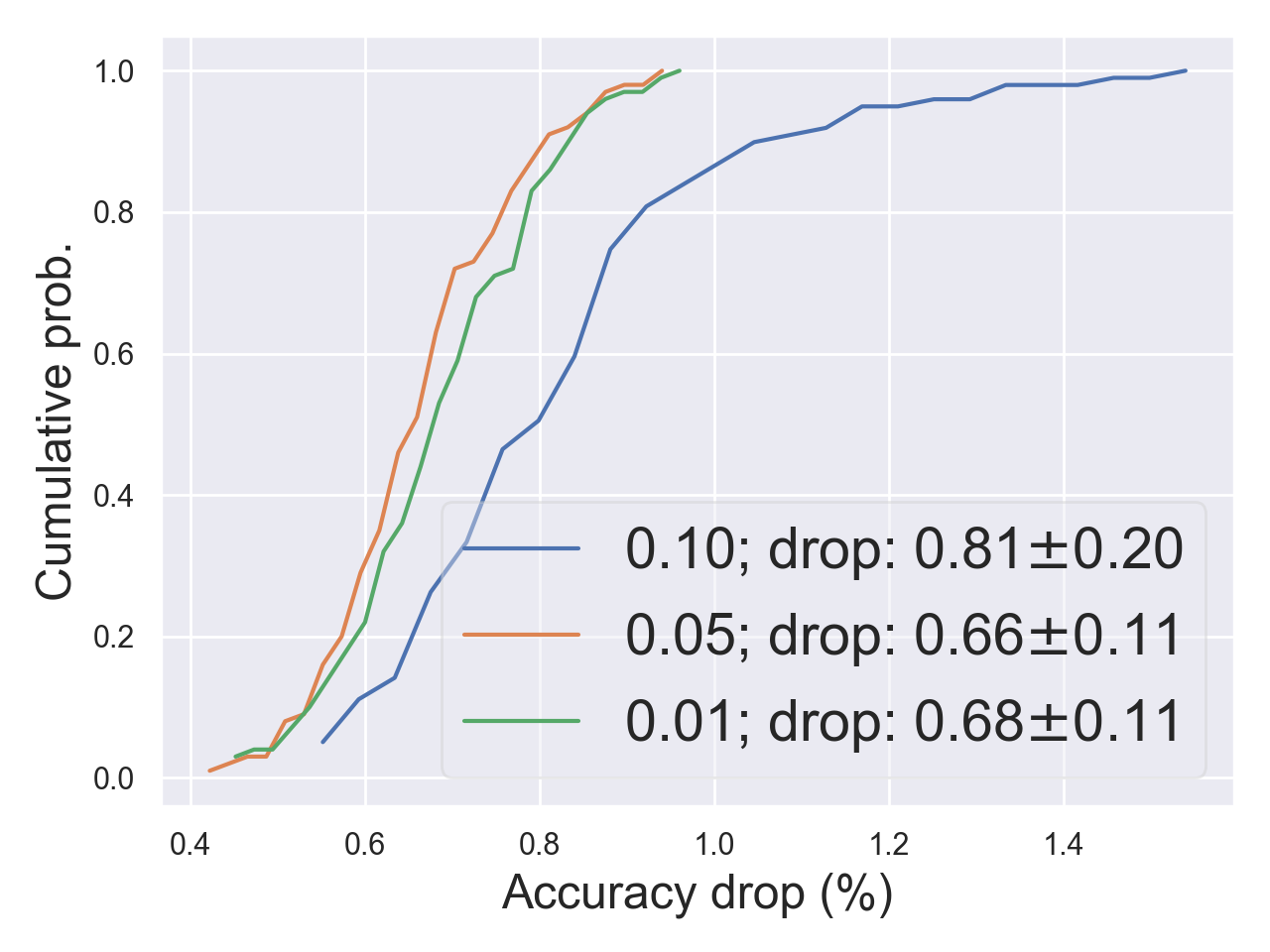}
\end{subfigure}
\begin{subfigure}[b]{\spacesparamsfigwidth\linewidth}
\includegraphics[width=\linewidth]{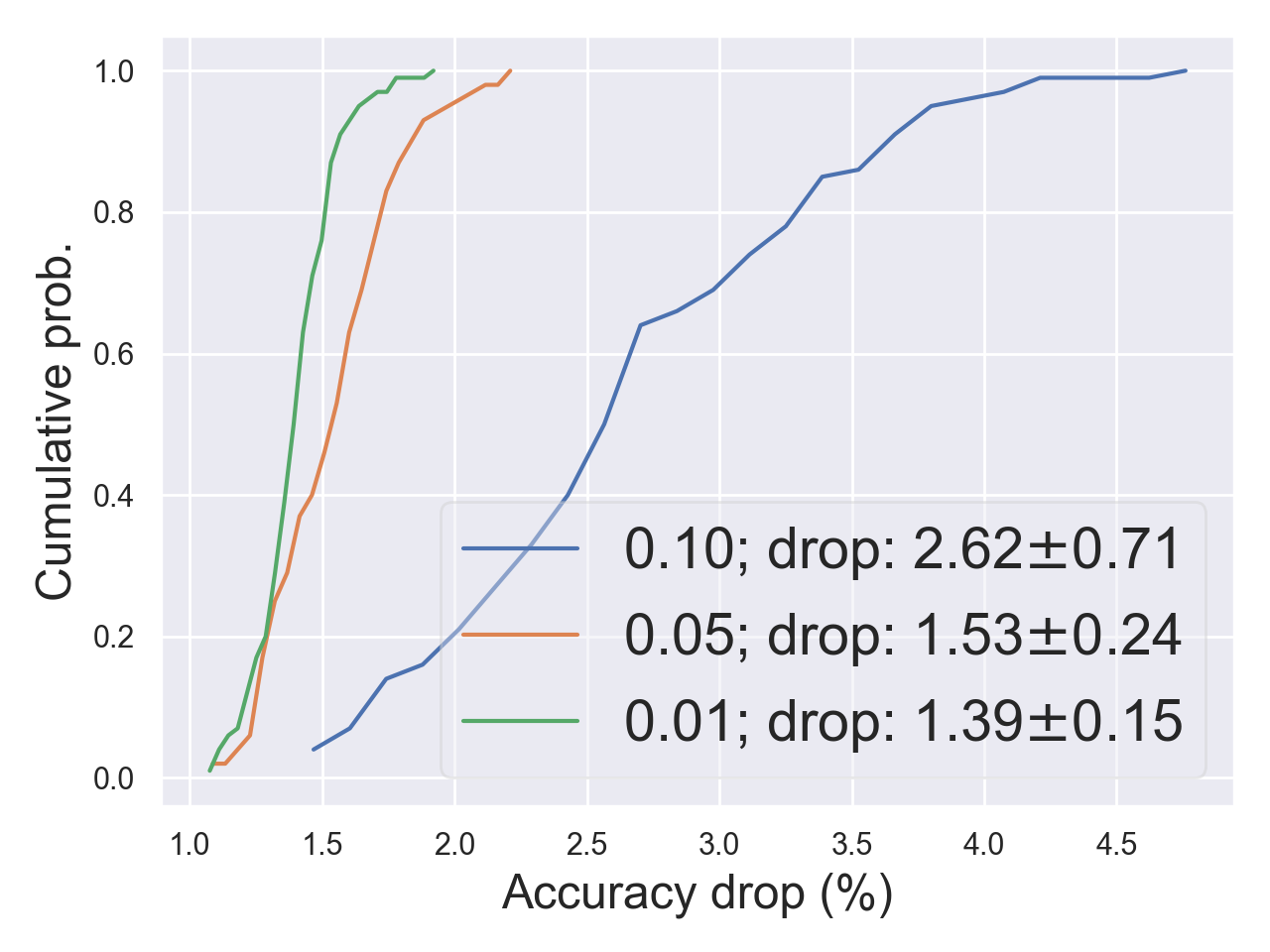}
\end{subfigure}
\begin{subfigure}[b]{\spacesparamsfigwidth\linewidth}
\includegraphics[width=\linewidth]{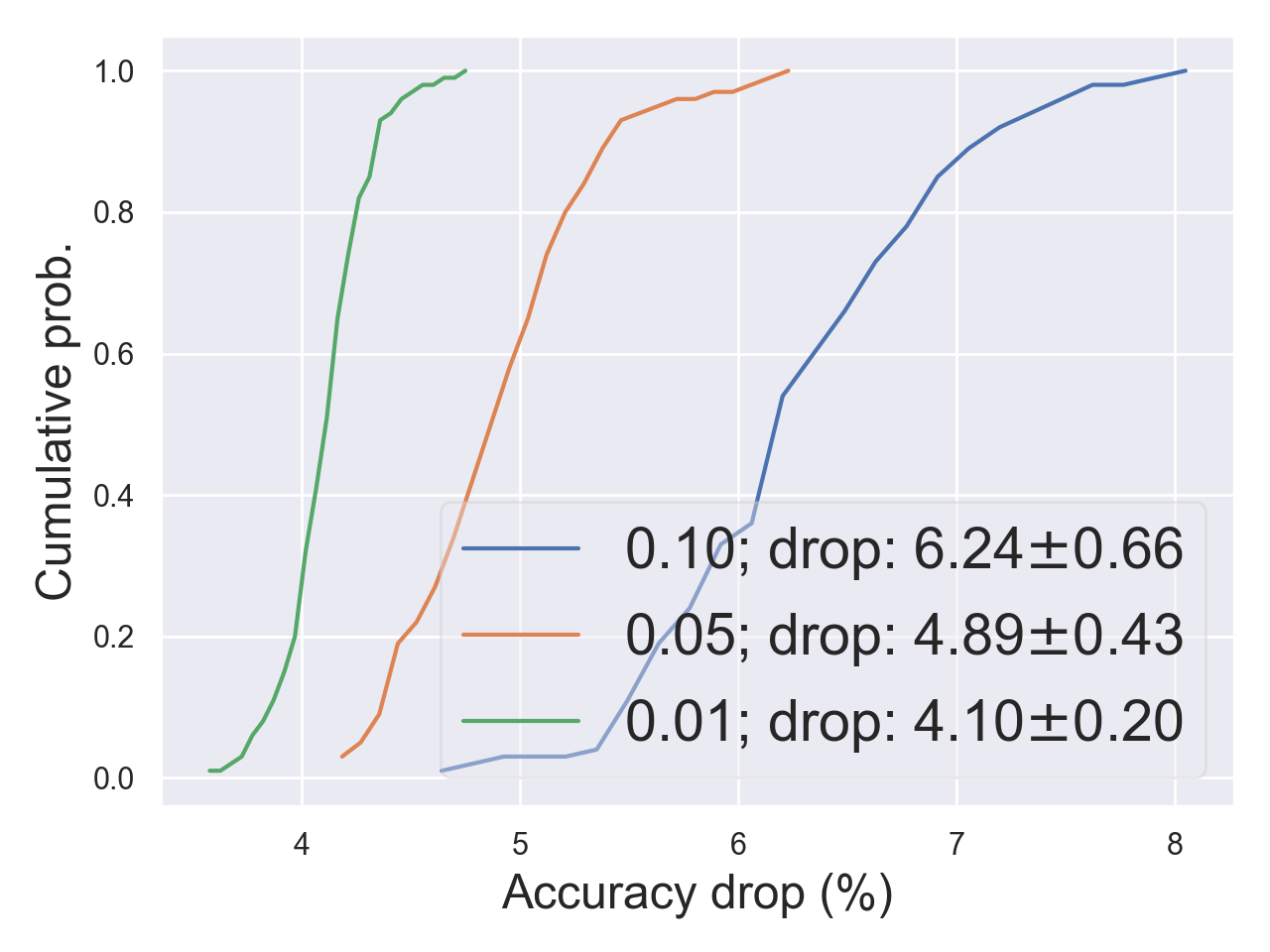}
\end{subfigure}
%
%
\begin{subfigure}[b]{\spacesparamsfigwidth\linewidth}
\includegraphics[width=\linewidth]{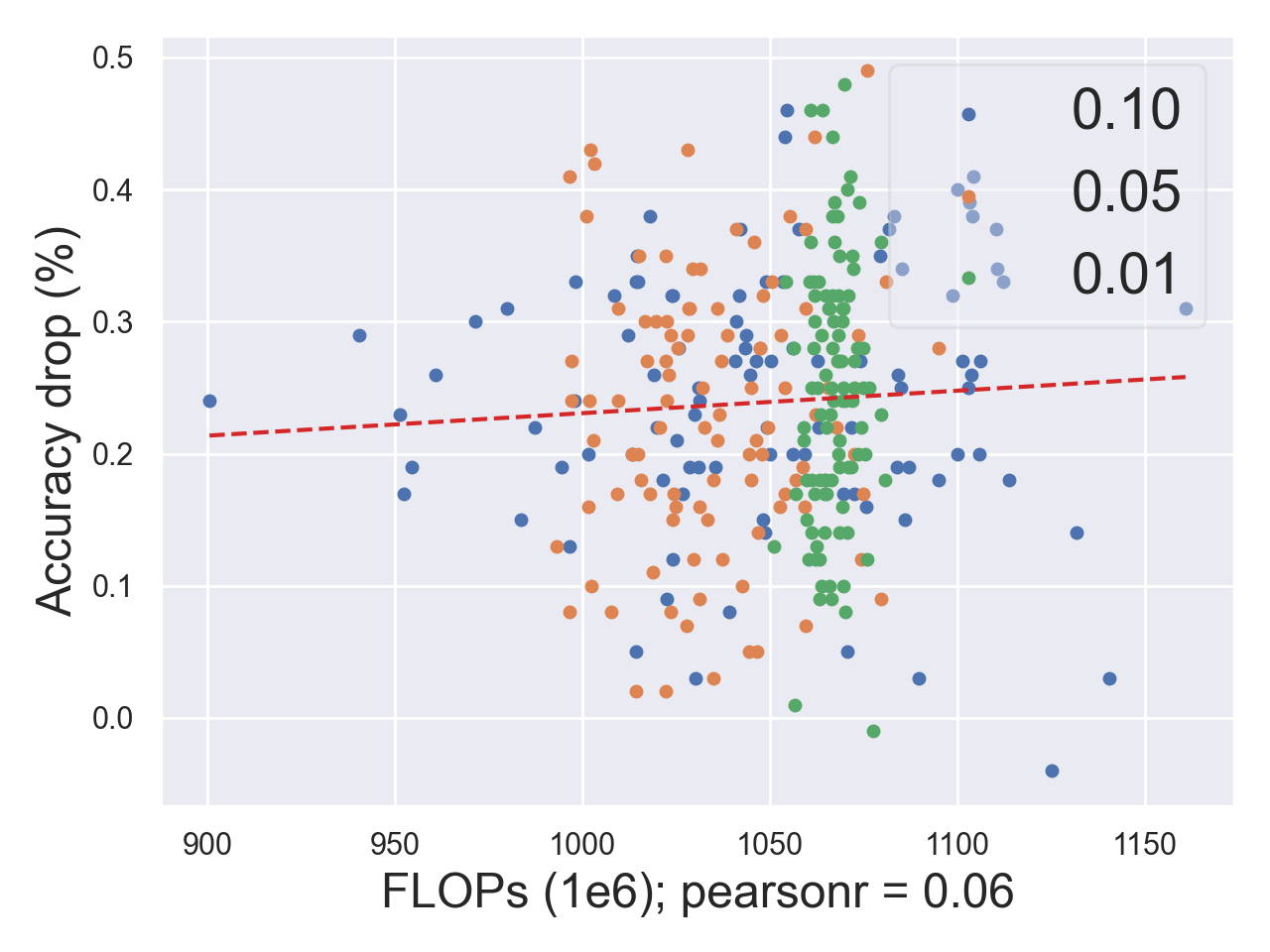}
\end{subfigure}
\begin{subfigure}[b]{\spacesparamsfigwidth\linewidth}
\includegraphics[width=\linewidth]{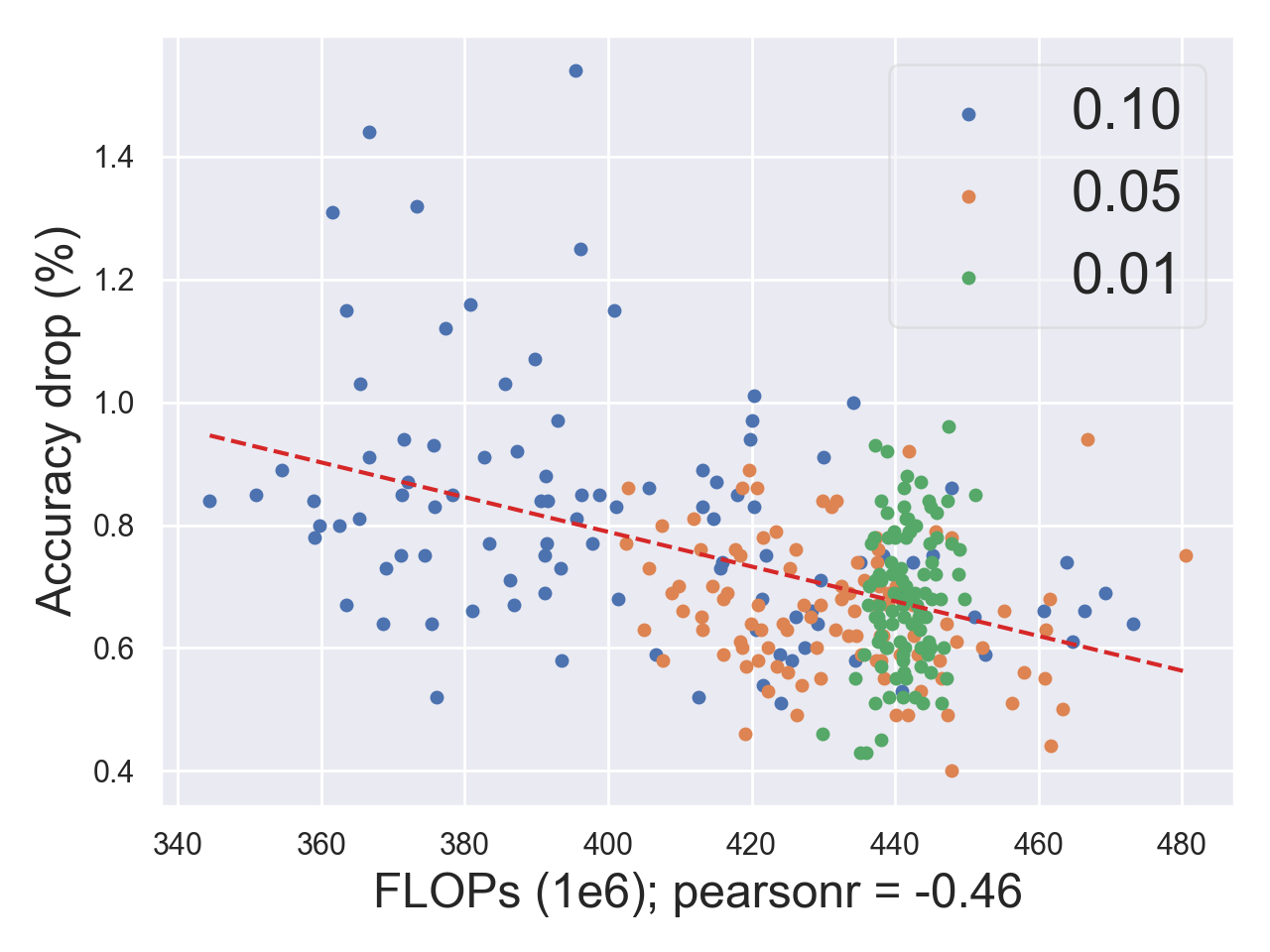}
\end{subfigure}
\begin{subfigure}[b]{\spacesparamsfigwidth\linewidth}
\includegraphics[width=\linewidth]{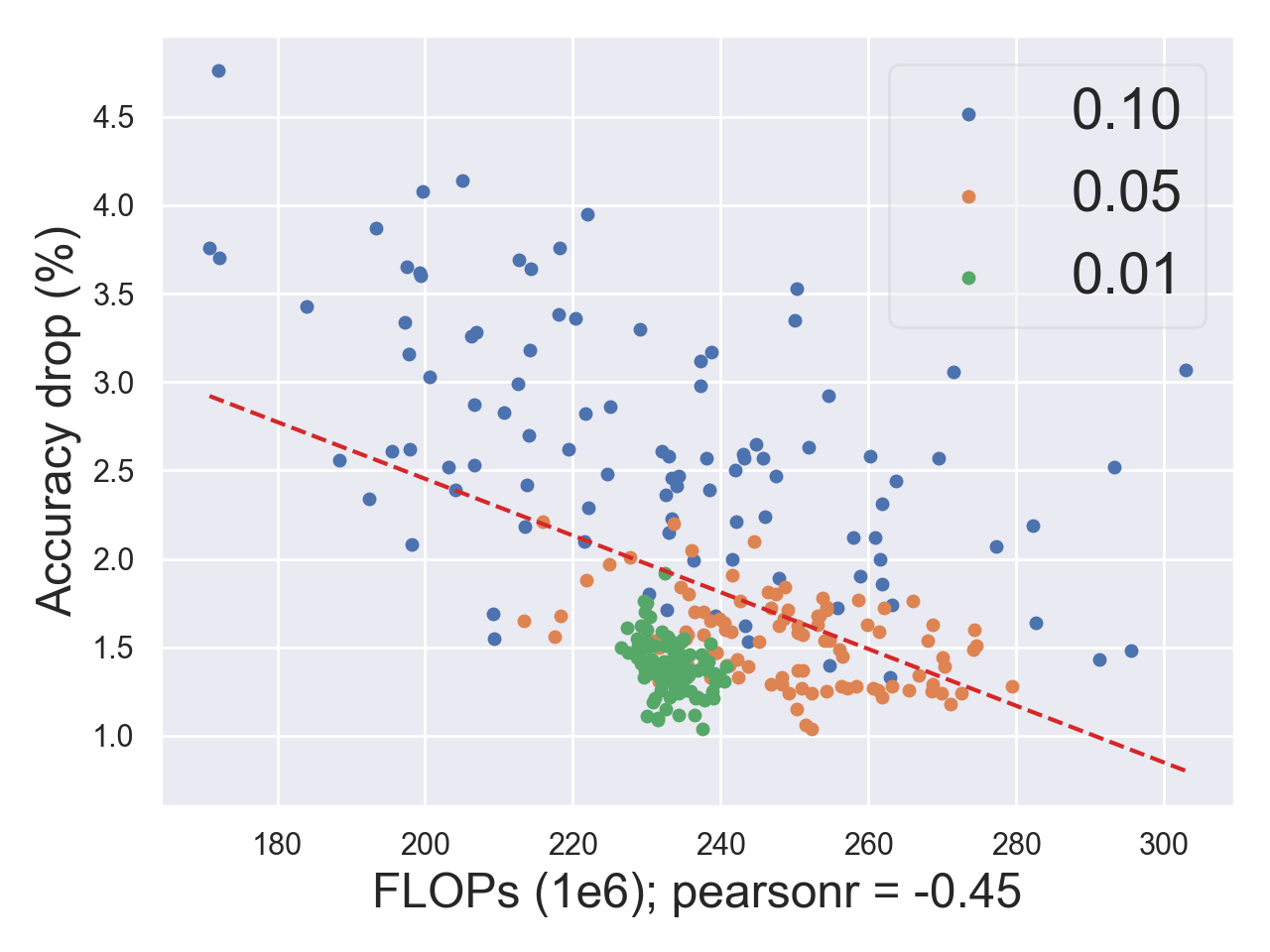}
\end{subfigure}
\begin{subfigure}[b]{\spacesparamsfigwidth\linewidth}
\includegraphics[width=\linewidth]{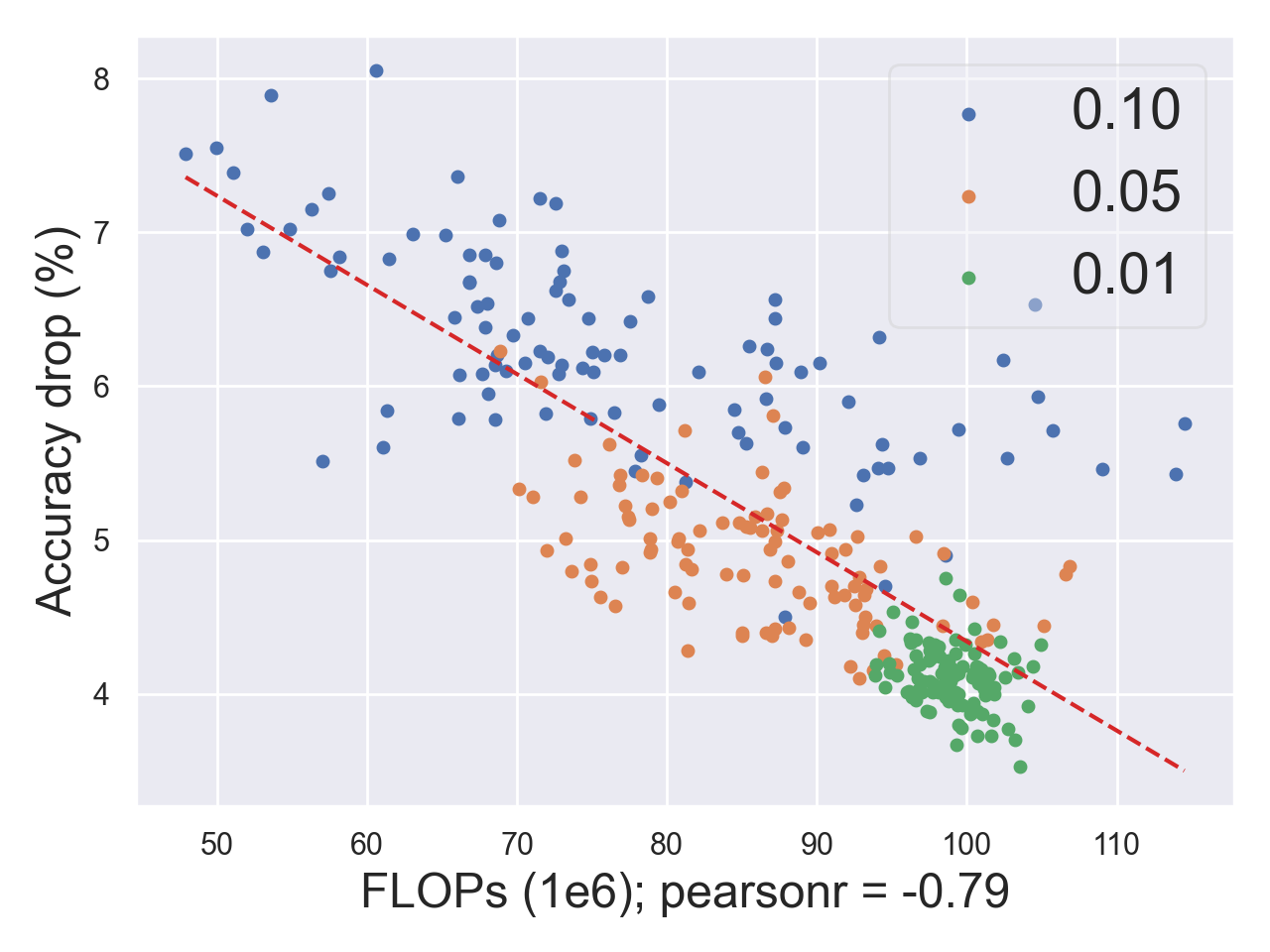}
\end{subfigure}
\begin{subfigure}[b]{\spacesparamsfigwidth\linewidth}
\includegraphics[width=\linewidth]{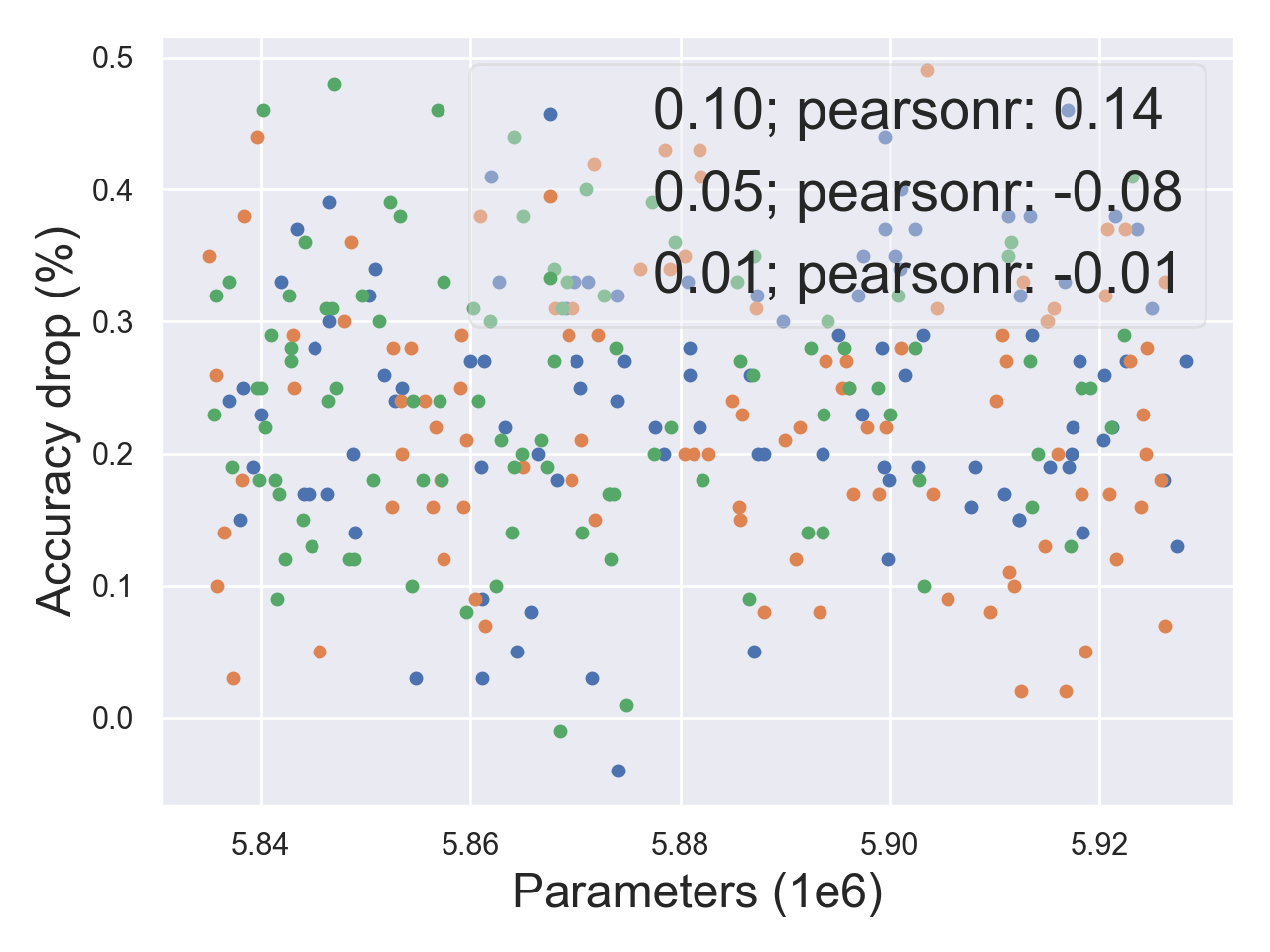}
\end{subfigure}
\begin{subfigure}[b]{\spacesparamsfigwidth\linewidth}
\includegraphics[width=\linewidth]{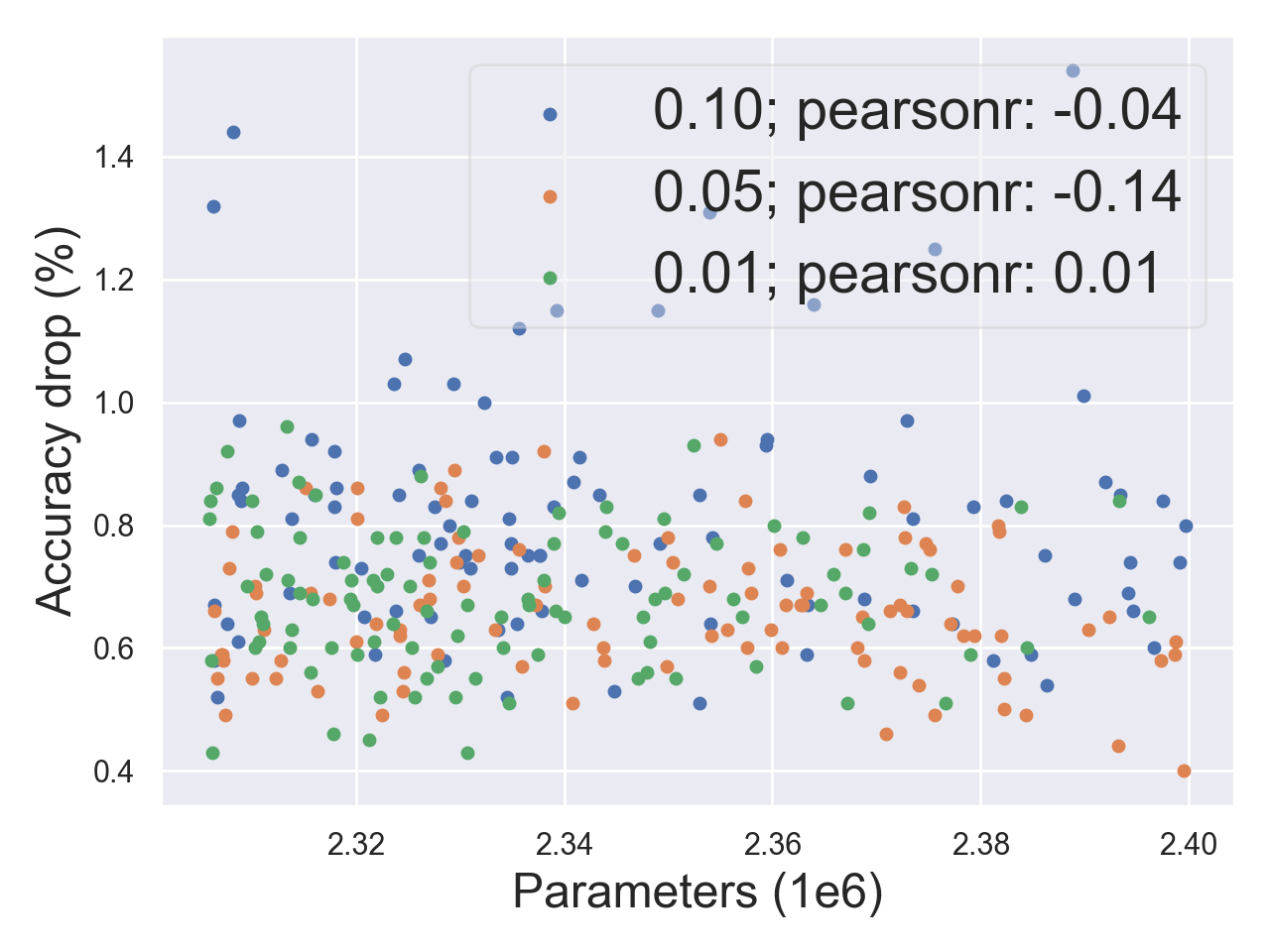}
\end{subfigure}
\begin{subfigure}[b]{\spacesparamsfigwidth\linewidth}
\includegraphics[width=\linewidth]{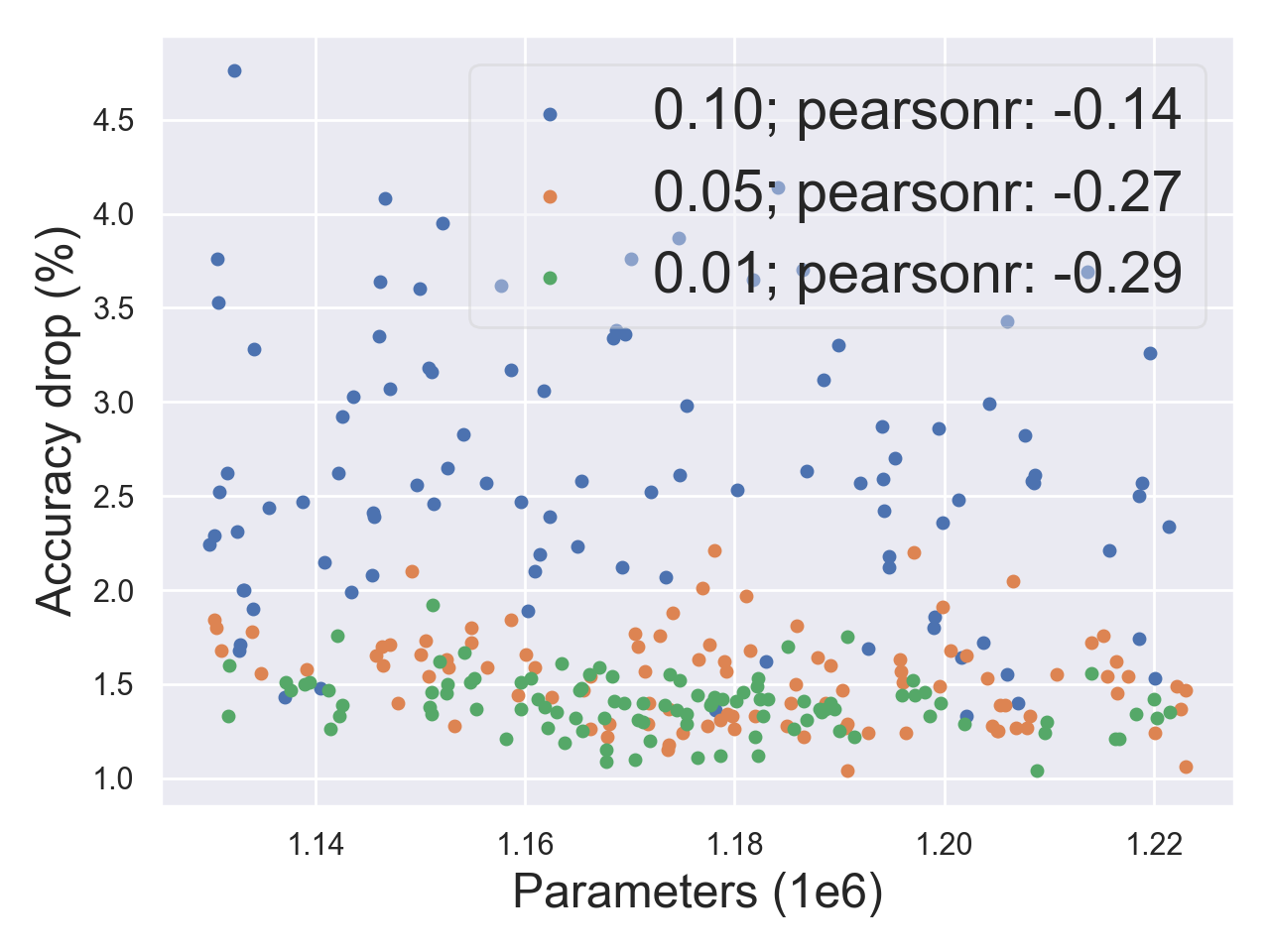}
\end{subfigure}
\begin{subfigure}[b]{\spacesparamsfigwidth\linewidth}
\includegraphics[width=\linewidth]{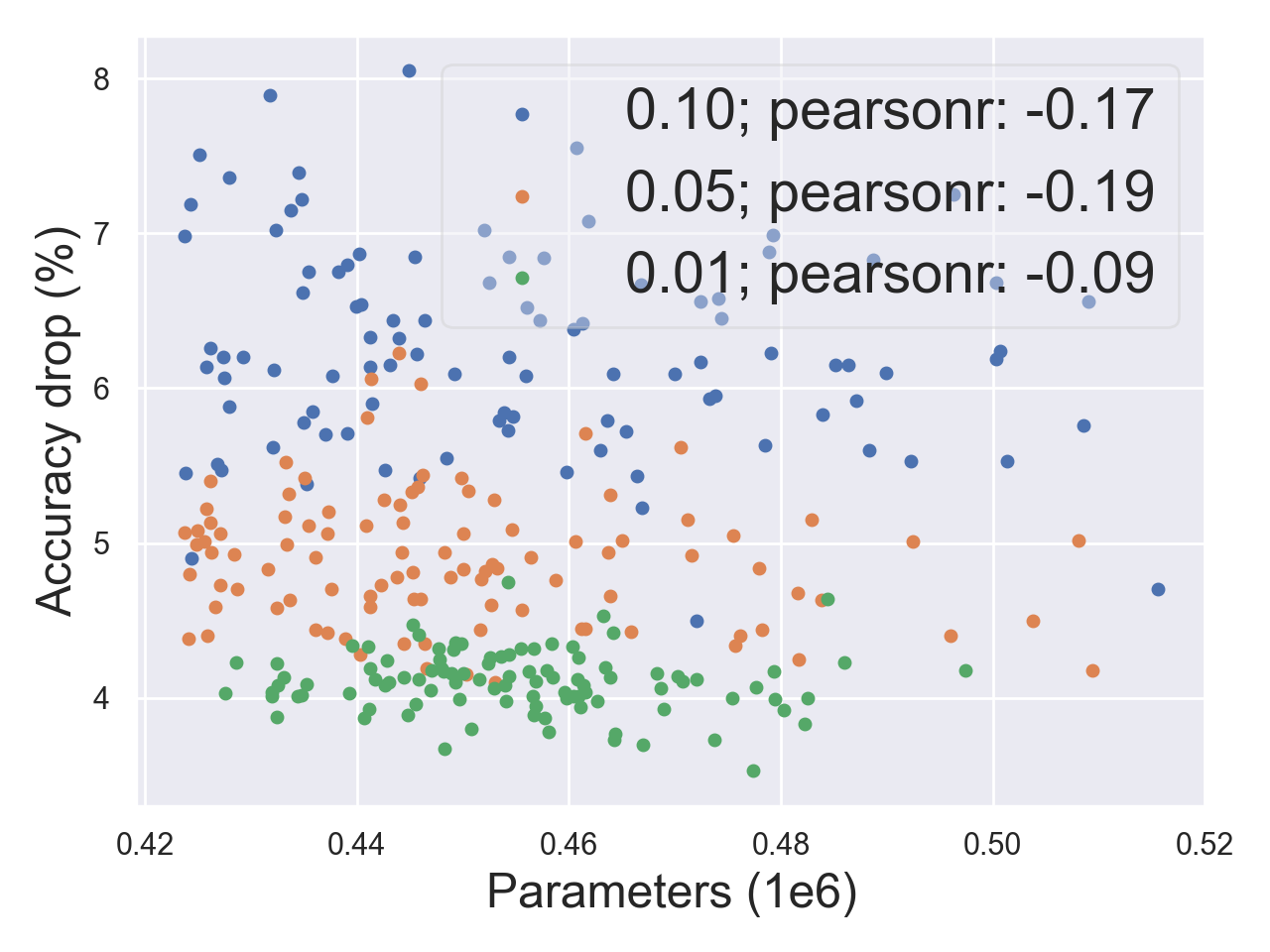}
\end{subfigure}
\begin{subfigure}[b]{\spacesparamsfigwidth\linewidth}
\includegraphics[width=\linewidth]{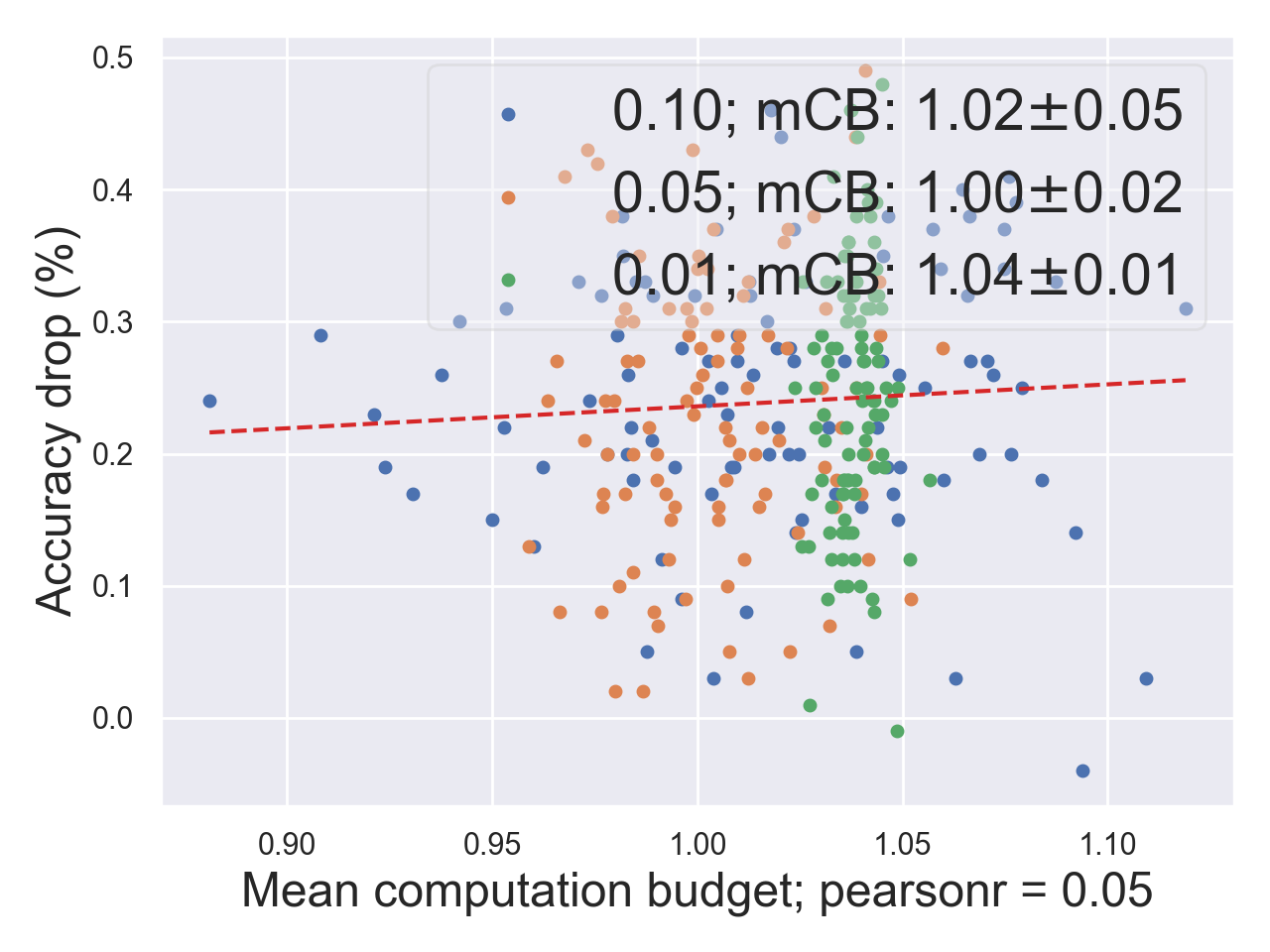}
\end{subfigure}
\begin{subfigure}[b]{\spacesparamsfigwidth\linewidth}
\includegraphics[width=\linewidth]{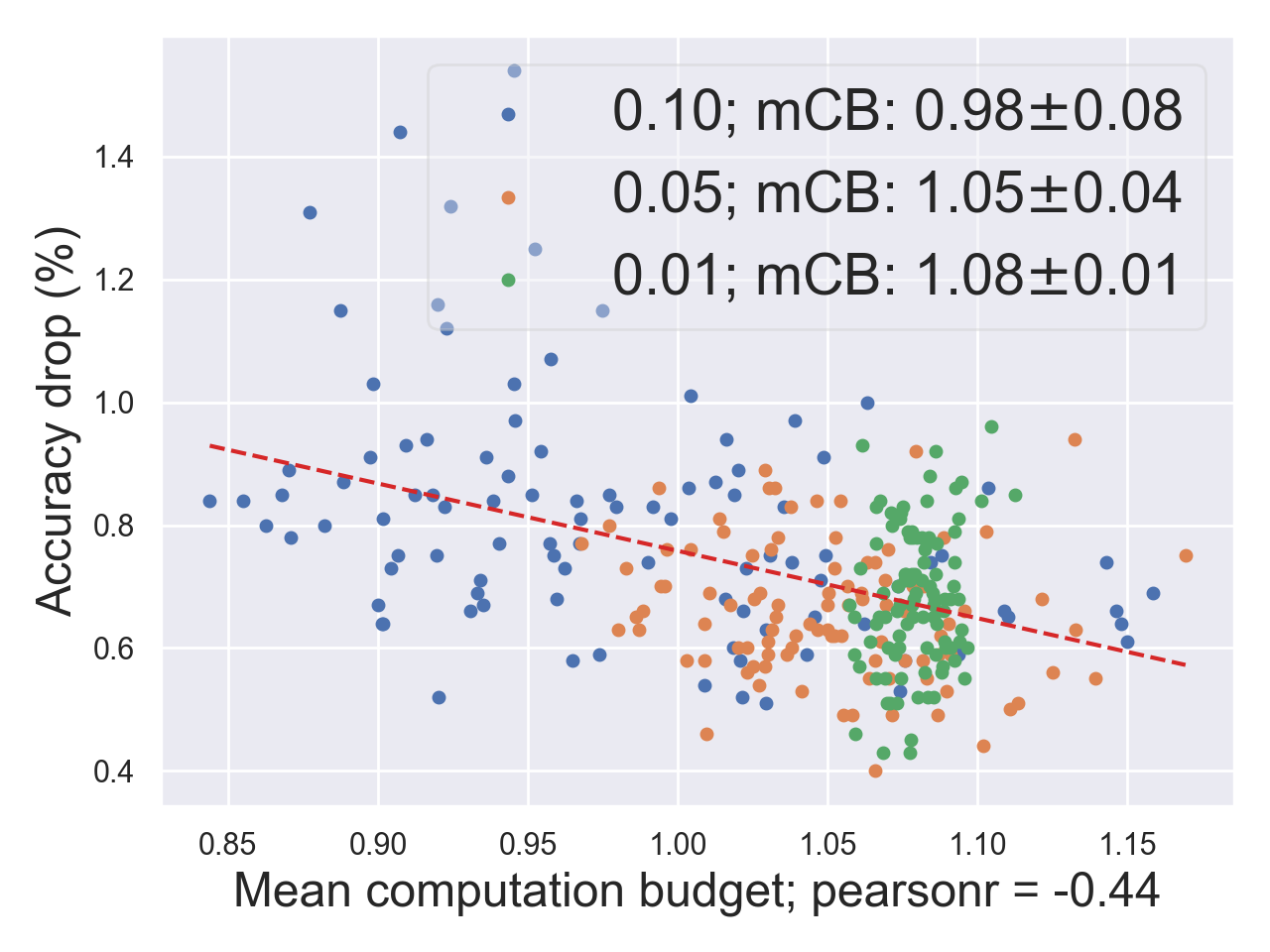}
\end{subfigure}
\begin{subfigure}[b]{\spacesparamsfigwidth\linewidth}
\includegraphics[width=\linewidth]{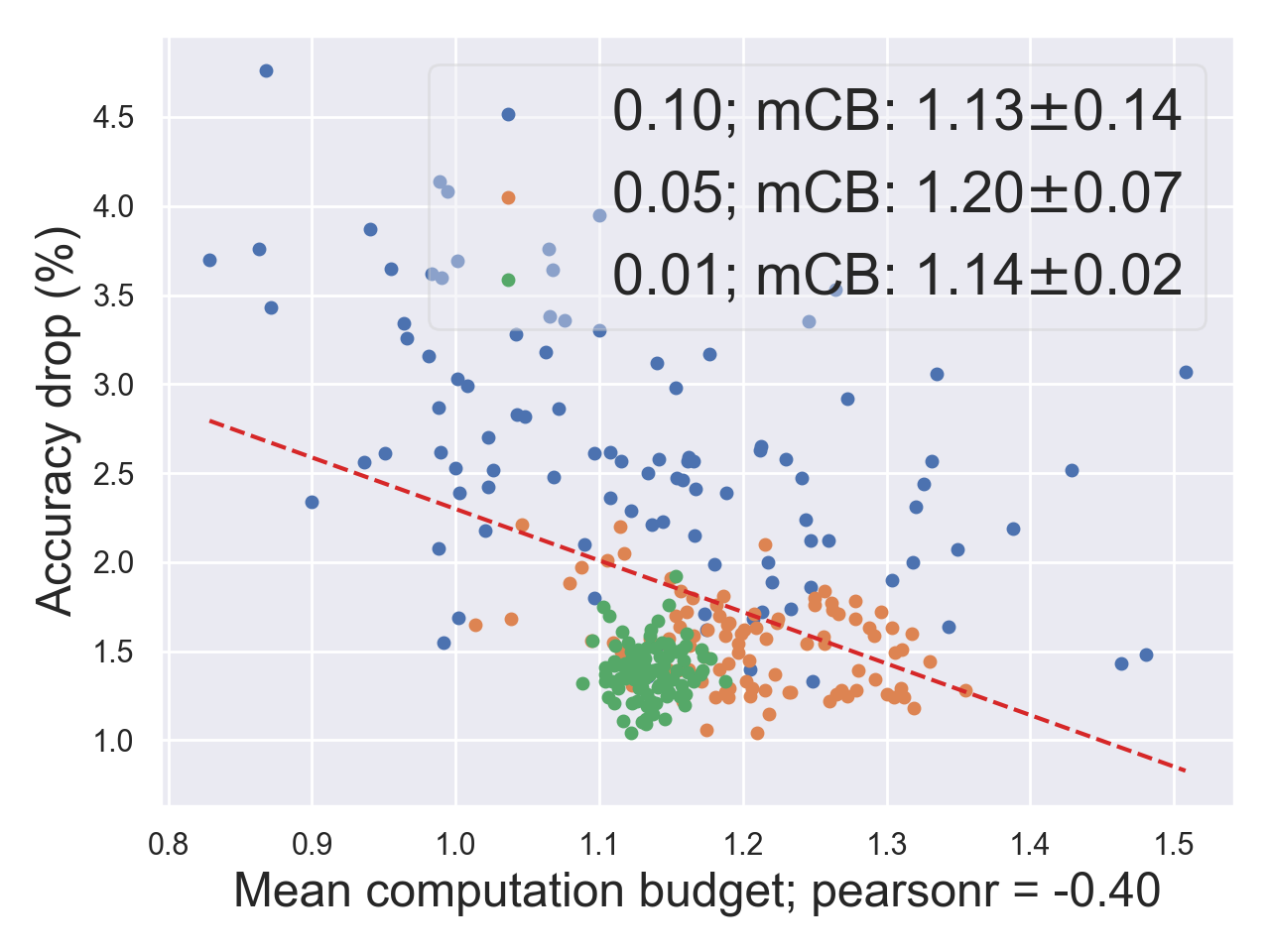}
\end{subfigure}
\begin{subfigure}[b]{\spacesparamsfigwidth\linewidth}
\includegraphics[width=\linewidth]{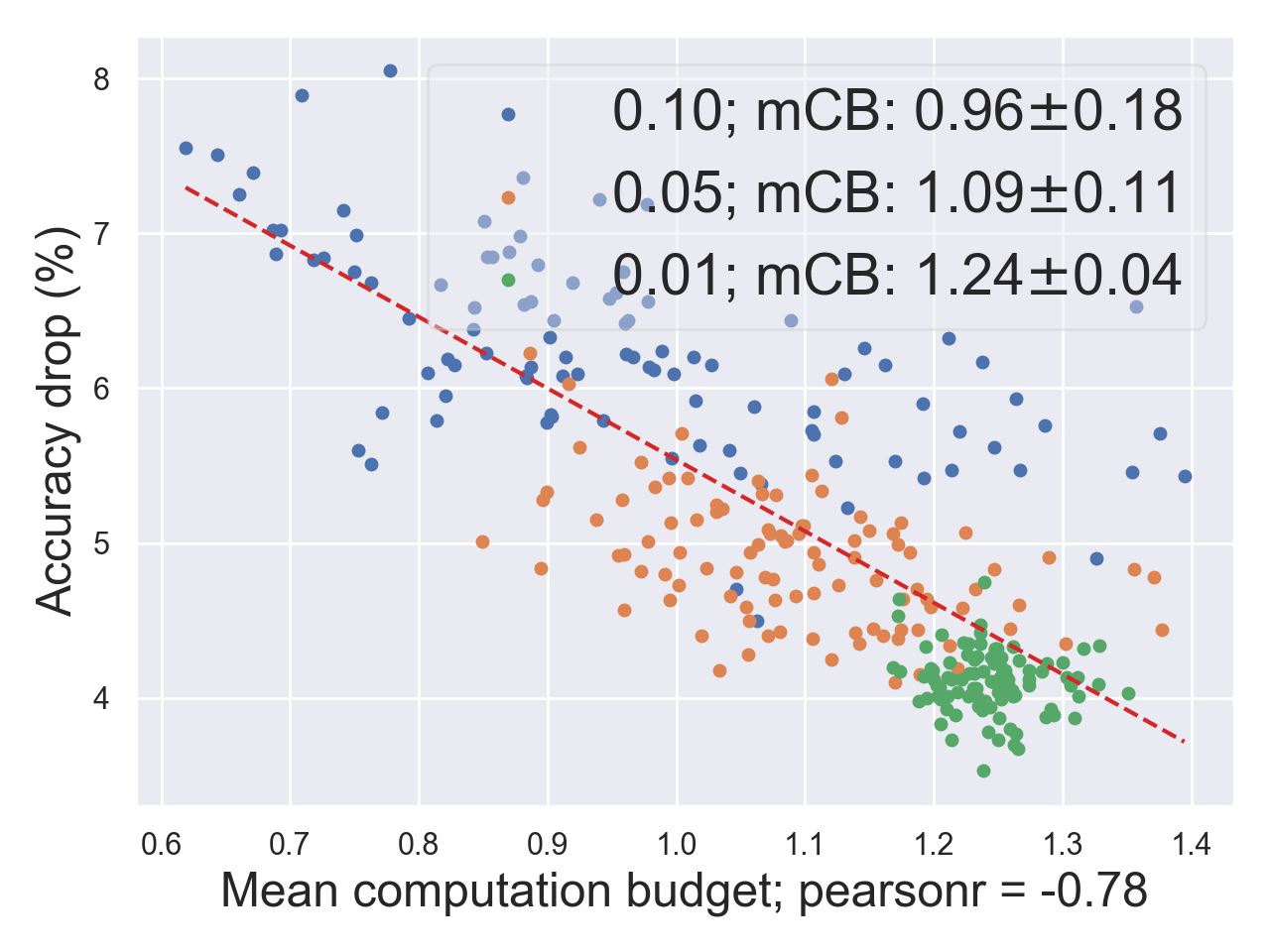}
\end{subfigure}
\begin{subfigure}[b]{\spacesparamsfigwidth\linewidth}
\includegraphics[width=\linewidth]{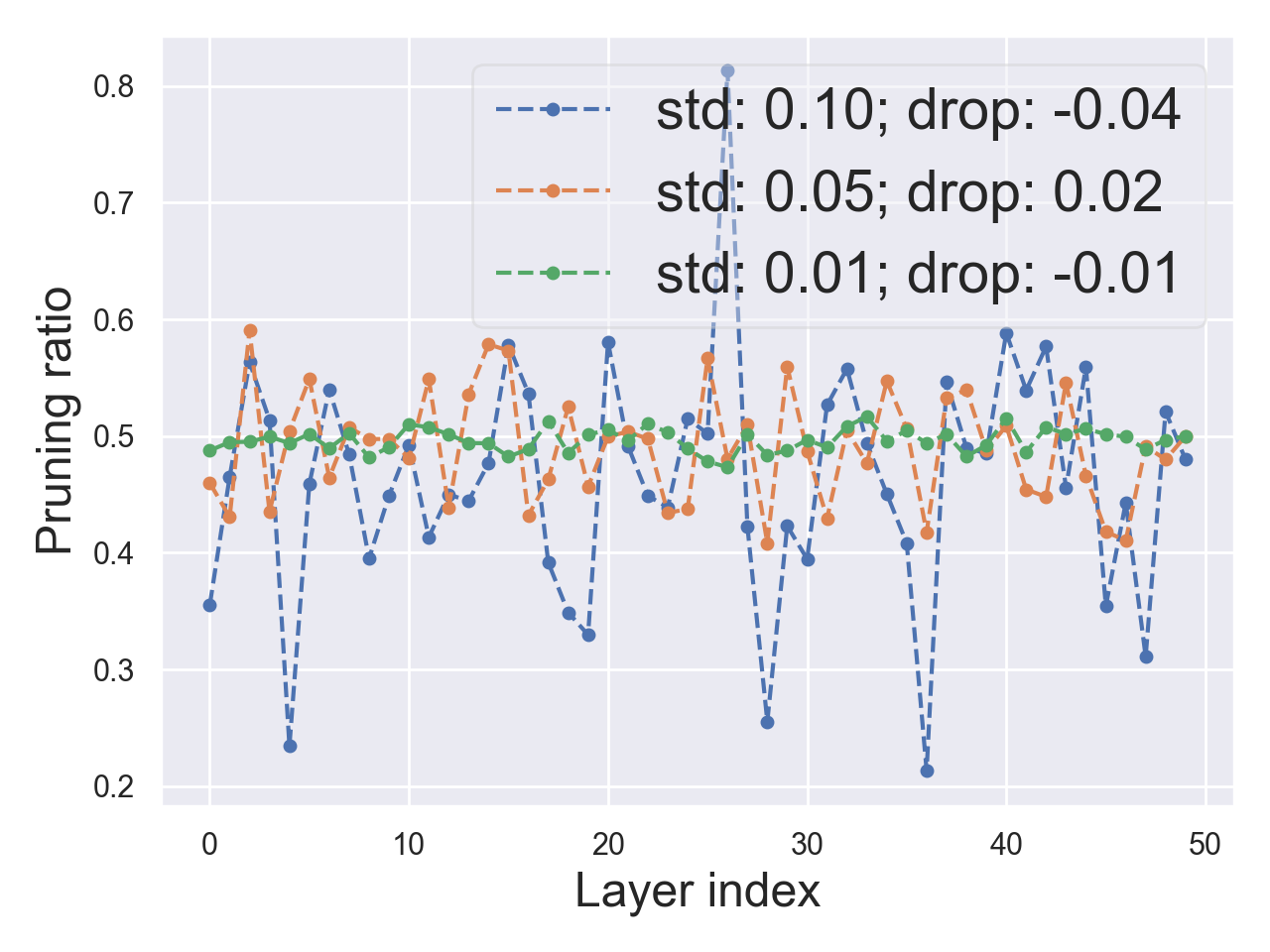}
\caption{$c_{\textnormal{params}} = 0.25$}
\label{fig:spaces_params.p75}
\end{subfigure}
\begin{subfigure}[b]{\spacesparamsfigwidth\linewidth}
\includegraphics[width=\linewidth]{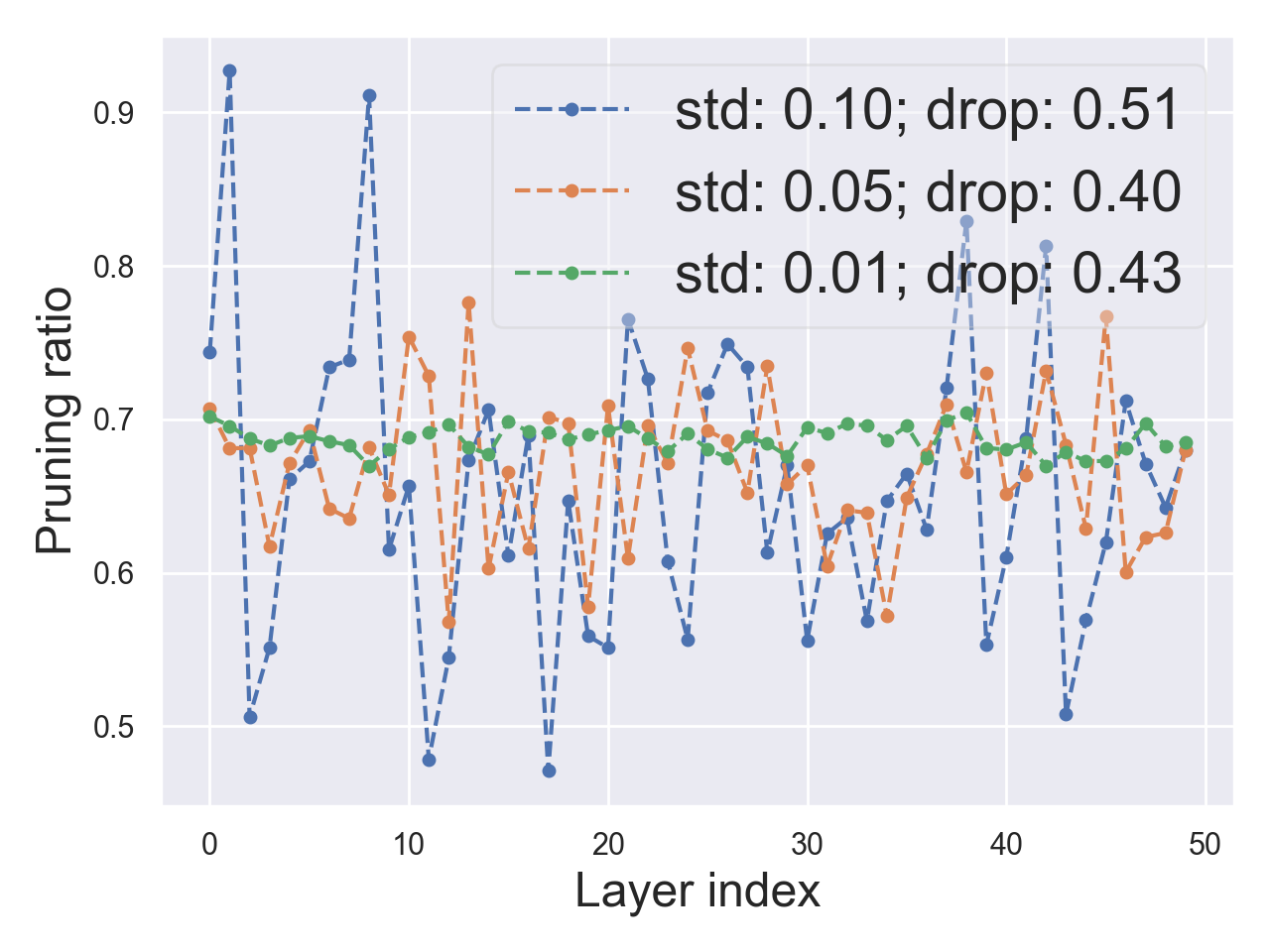}
\caption{$c_{\textnormal{params}} = 0.1$}
\label{fig:spaces_params.p90}
\end{subfigure}
\begin{subfigure}[b]{\spacesparamsfigwidth\linewidth}
\includegraphics[width=\linewidth]{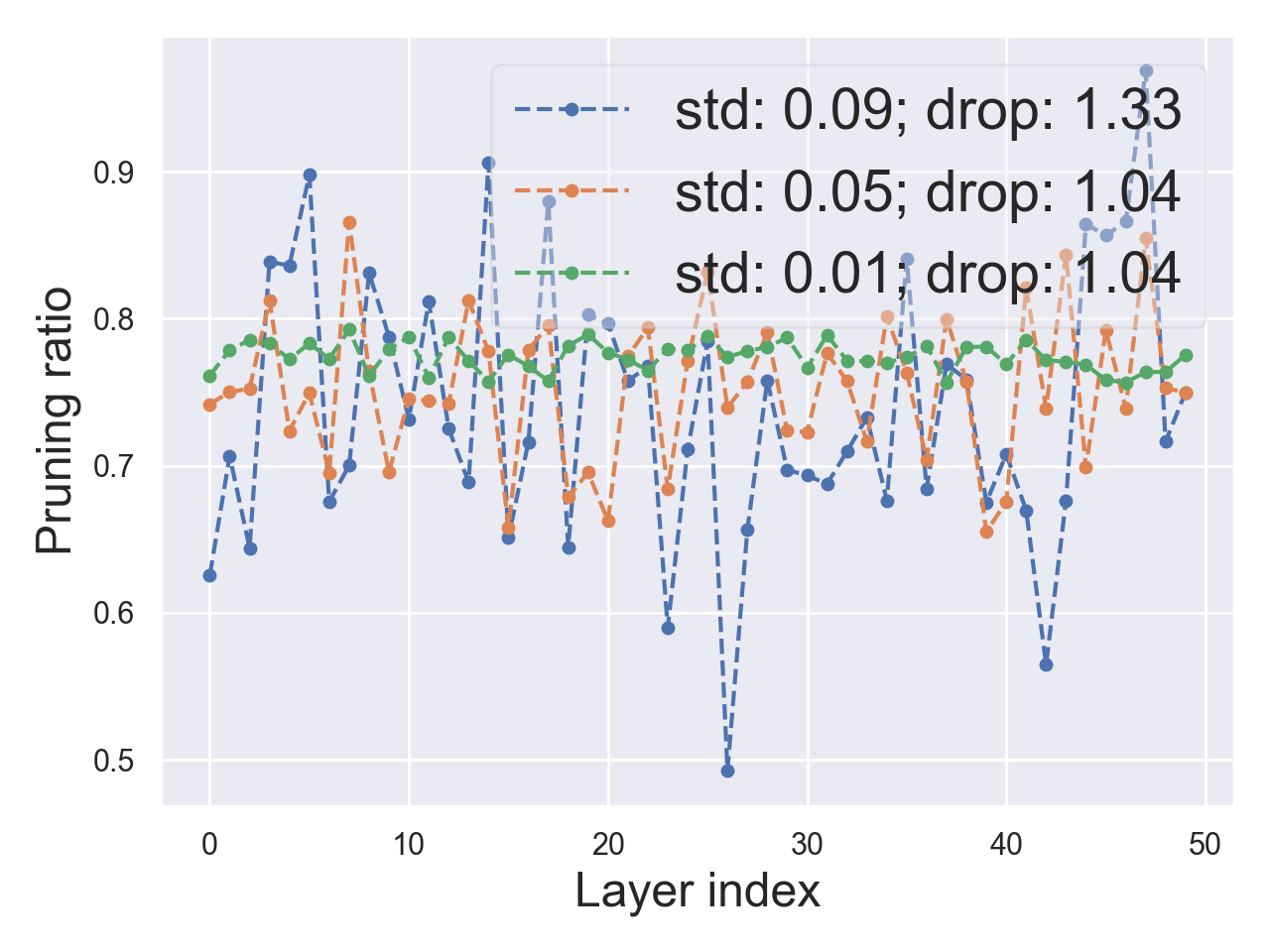}
\caption{$c_{\textnormal{params}} = 0.05$}
\label{fig:spaces_params.p95}
\end{subfigure}
\begin{subfigure}[b]{\spacesparamsfigwidth\linewidth}
\includegraphics[width=\linewidth]{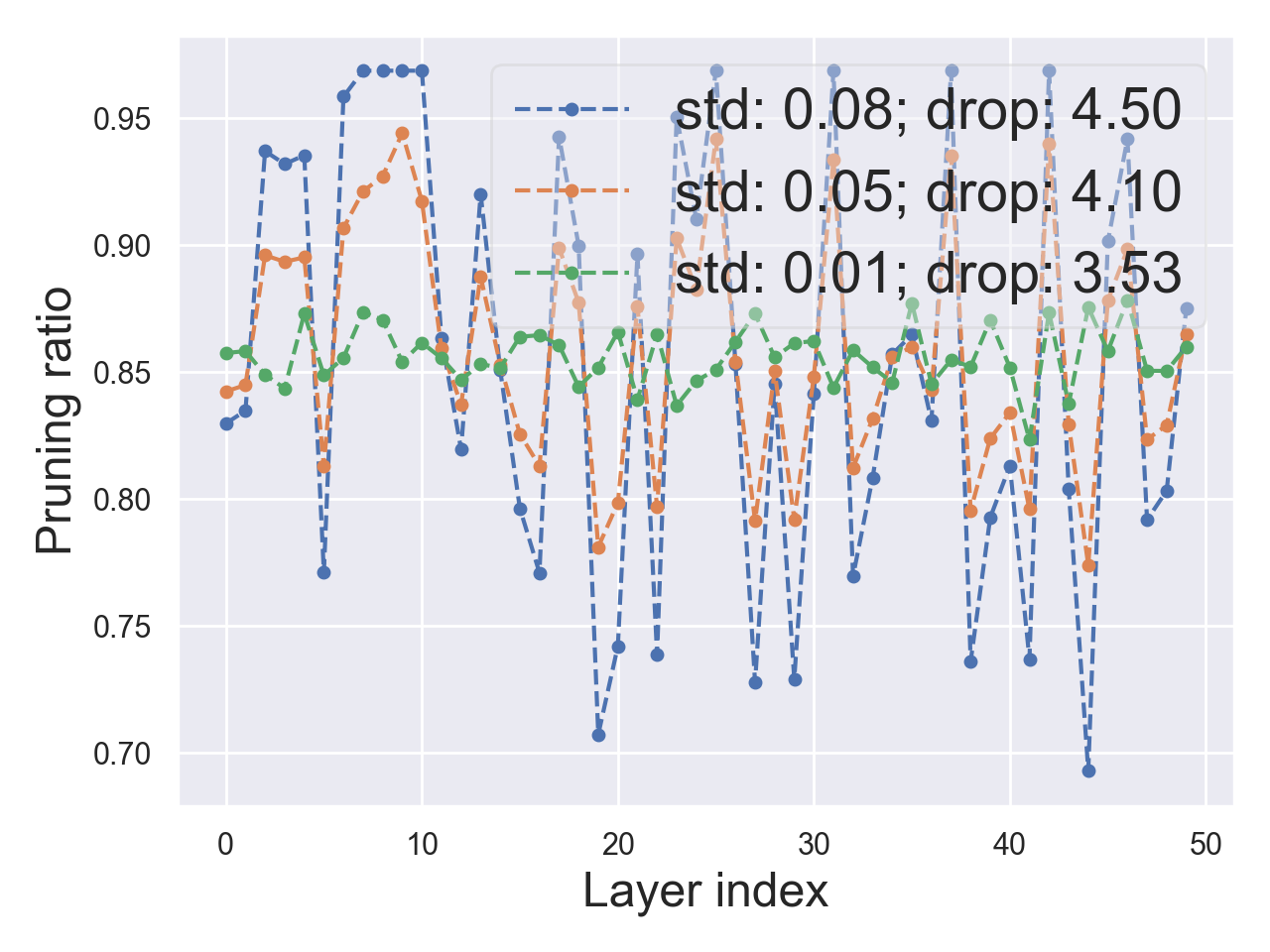}
\caption{$c_{\textnormal{params}} = 0.02$}
\label{fig:spaces_params.p98}
\end{subfigure}
\caption{
\texttt{STD} pruning spaces under $c_{\textnormal{params}} = \{0.25, 0.1, 0.05, 0.02\}$.
In the figure, we study three \texttt{STD} pruning spaces, including \texttt{STD-0.1}, \texttt{STD-0.05} and \texttt{STD-0.01} spaces with a constraint on parameter bucket.
(\emph{Row 1}) Accuracy drop EDFs on \texttt{STD} spaces.
(\emph{Row 2}) Remaining FLOPs distribution.
(\emph{Row 3}) Remaining parameters distribution.
(\emph{Row 4}) Mean computation budget distribution.
(\emph{Row 5}) The winning pruning recipe on each \texttt{STD} pruning space.
}
\label{fig:spaces_params}
\end{figure}

Based on Conjectures 1 and 2, we believe that the winning subnetworks in different pruning regimes have optimal configurations of \emph{(FLOPs, parameter bucket)}, which are achieved by their pruning recipes.
To further reduce the degrees of freedom, we introduce a tool that represents the FLOPs-to-parameter-bucket ratio compared to the original network, namely \emph{mean computation budget} (mCB):
\begin{equation}
\textnormal{mCB} = \frac{
\textnormal{FLOPs}(f_{\theta_{\textnormal{sub}}}) / \textnormal{Params}(f_{\theta_{\textnormal{sub}}})
}{
\textnormal{FLOPs}(f_\theta) / \textnormal{Params}(f_\theta)
} .
\end{equation}
In \Figref{fig:spaces} (Row 5) and \Figref{fig:spaces_params} (Row 4) we present mCB distribution of subnetworks with $c_{\textnormal{flops}}$ and $c_{\textnormal{params}}$ respectively.
With this tool, we combine Conjectures 1 and 2 into one and have the following conjecture:

\textbf{Conjecture 3.}
\emph{
(\textbf{a}) In a pruning regime, there exists an optimal mean computation budget for pruning filters;
(\textbf{b}) Subnetworks that have the optimal mean computation budget will be the winning ones on $\sS$;
(\textbf{c}) A good structure of pruned subnetwork guarantees an approximately optimal mean computation budget in this pruning regime.}

\begin{figure}[t]
\centering
\begin{subfigure}{0.45\linewidth}
\includegraphics[width=\linewidth]{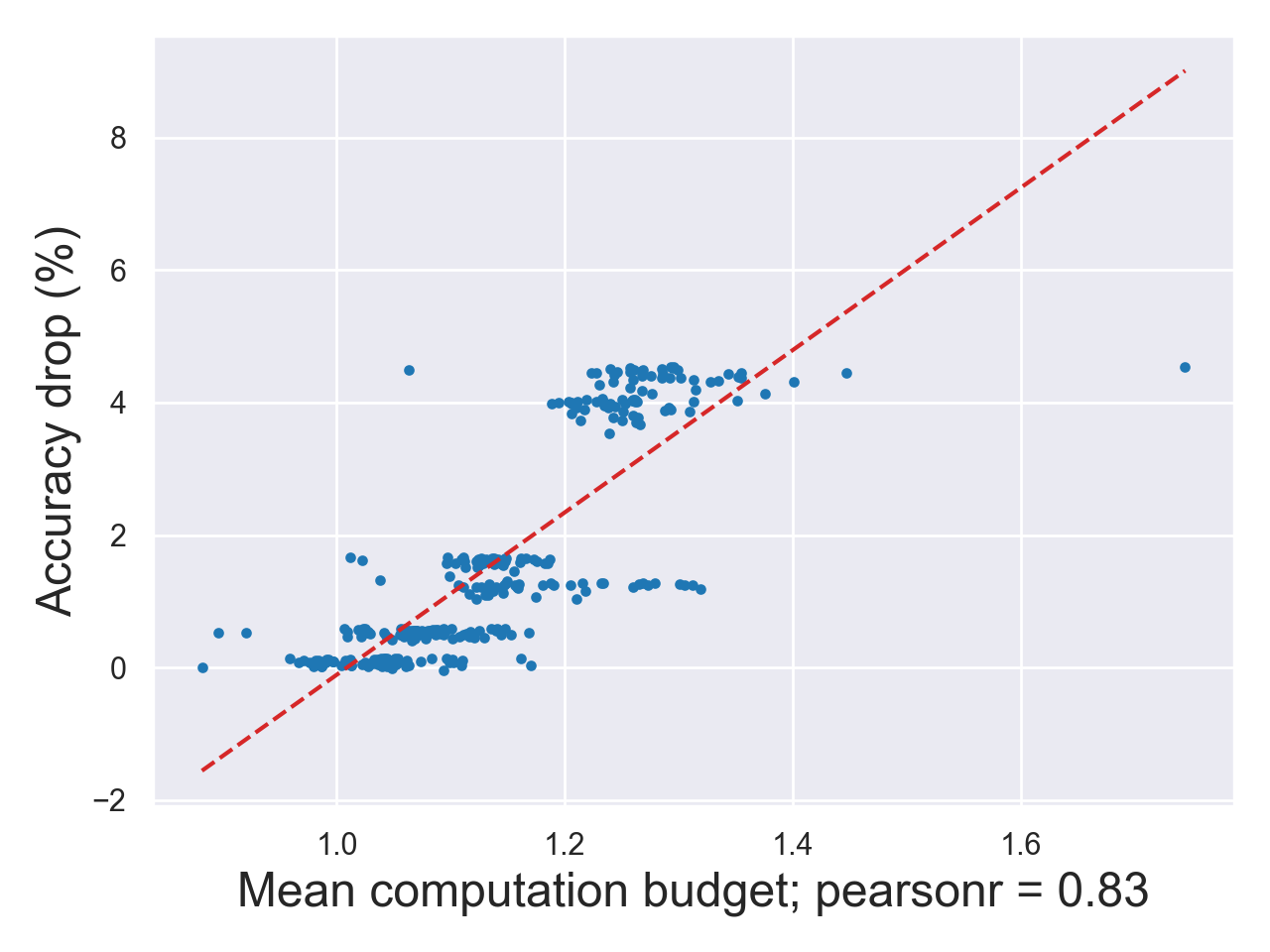}
\end{subfigure}
\begin{subfigure}{0.45\linewidth}
\includegraphics[width=\linewidth]{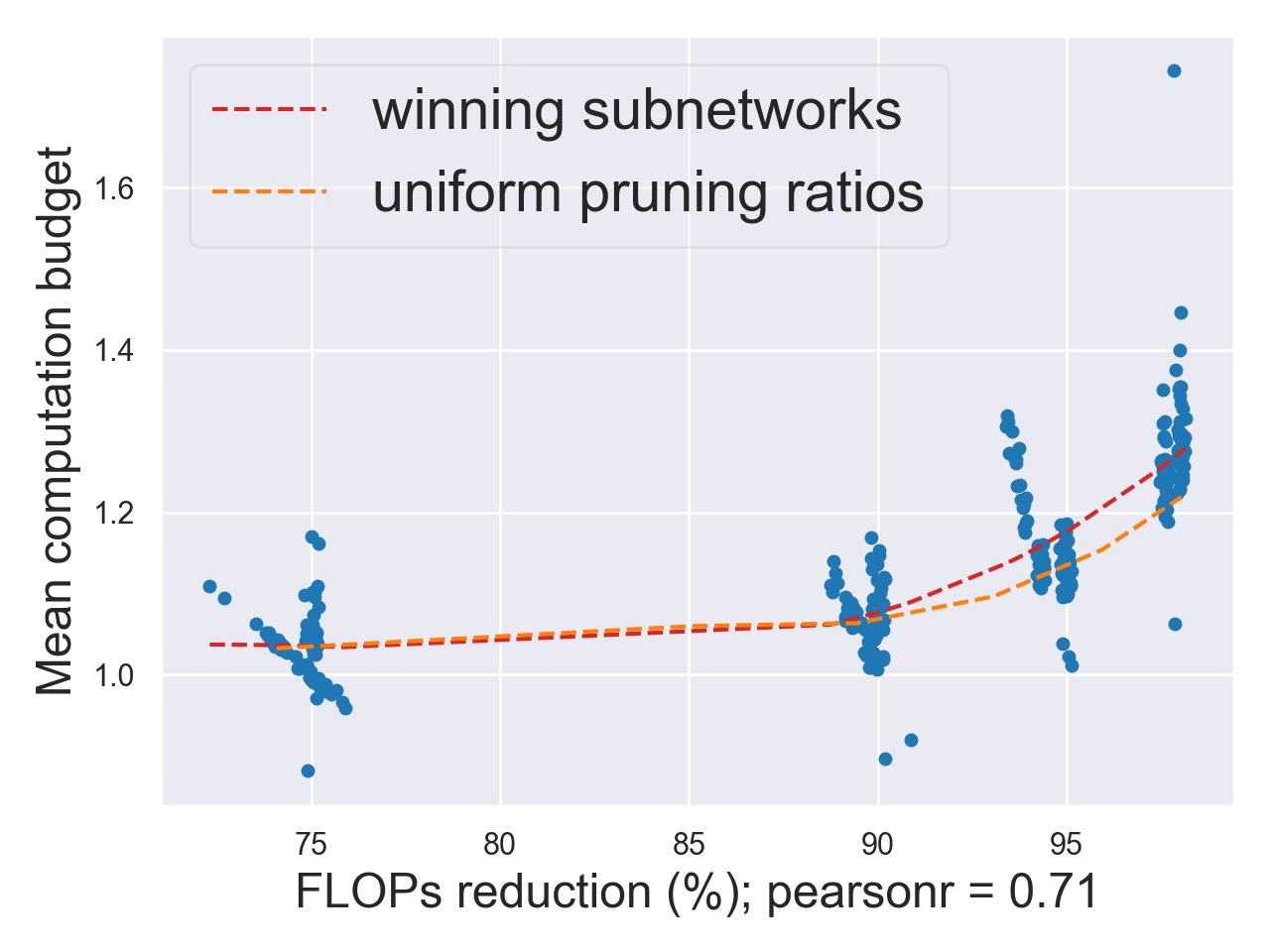}
\end{subfigure}
\caption{
(\emph{Left}) Sort winning subnetworks under different constraints ($c_{\textnormal{flops}}$ and $c_{\textnormal{params}}$) by their accuracy drops.
Each one is the best on all \std spaces under a constraint.
(\emph{Right}) We present the best 40 subnetworks on all \std spaces under each constraint.
}
\label{fig:budget}
\end{figure}


Taking the design of the original network as the optimal configuration of \emph{(FLOPs, parameter bucket)}, we argue that the optimal mCBs in many pruning regimes are roughly equivalent to 1.0.
In~\Figref{fig:budget} (right), we present mCB distribution of winning subnetworks when FLOPs reduction increases.
Statistically, our experimental results show that the optimal mCB for a pruning regime should be a dynamic range and the optimal mCB range is around 1.0 in most pruning regimes.

Moreover, the optimal mCB tends to increase when we reduce more FLOPs by pruning.
We draw a fitted curve of optimal mCB in~\Figref{fig:budget} (right, red dashed line) and the curve rises significantly when the reduction is above $90\%$.
We also draw a curve (orange dashed line) presenting the mCB of subnetworks with uniform pruning ratios.
These two curves are close in most regimes, suggesting that an uniform recipe is a good starting point when we search winning pruning recipes.
That might be the reason why uniform pruning ratios outperform some pruning methods in our extremely pruning experiments.

In~\Figref{fig:best40}, we observed that the optimal parameter bucket, the optimal FLOPs and the final performance of winning subnetworks seemed to be predictable.
This observation is consistent with \citet{rosenfeld2021predictability}, in which a scaling law is developed to estimate the error after weight pruning.
Therefore, we have another conjecture:

\textbf{Conjecture 4.}
\emph{
The limitation of performance in each pruning regime can be predicted by a functional form.}\footnote{
The limitations of performance and the curves in~\Figref{fig:best40} can be further refined, as we only retrain subnetworks for 50 epochs.
}

In~\Figref{fig:budget} (left), we present the relationship between the mCBs of winning subnetworks and their accuracy drops.
We observed that the limitation of performance could be predicted by a function of the optimal mCB in a pruning regime.
Since the goal of network pruning is to increase efficiency while maintaining accuracy, such an empirical function helps a lot when one is making a trade-off.
Moreover, we believe that one might find a winning subnetwork without any accuracy drop in many pruning regimes, where the optimal mCB ranges are around 1.0.\footnote{
We argue that the limitation is related to the dataset size. Our experiments on ImageNet fail to find such a subnetwork without any accuracy drop when we reduce more than 75\% FLOPs.}
\section{ResNet-50 on ImageNet}

In this section, we prune ResNet-50 on ImageNet~\citep{russakovsky2015imagenet} with $c_{\textnormal{flops}} = 0.5$ and $c_{\textnormal{flops}} = 0.25$.
Since ImageNet is a large-scale dataset that takes more time for one epoch than CIFAR-10, we first generate pruning recipes and then fine-tune subnetwork candidates for 5 epochs.
The top-5 ranking candidates will be retrained with a complete schedule.
Specifically, we fine-tune subnetworks with an initial learning rate $lr = 0.01$ for 100 epochs on ImageNet.

In the experiments, we produce pruning recipes with a refined pruning space upon the proposed Conjecture 3.
We empirically use a constraint on mCB $c_{\textnormal{mCB}} = 1.0 \pm 0.1$ instead of a constraint on the standard deviation of recipes $\vr$ for efficiency. This is because all \std spaces have a similar distribution in such a pruning regime, according to our analysis for CIFAR-10.
Note that the excepted parameter bucket range is decided by $c_{\textnormal{mCB}}$ when a target FLOPs reduction is set.
We generate $n = 300$ subnetwork candidates and finally keep the top-5 ranking candidates.
In Table~\ref{tab:resnet50_imagenet}, we compare the winning subnetwork to the state-of-the-art pruning methods.
The comparison shows that our analysis for network pruning spaces guides us to better pruning results.

\begin{table}[t]
\centering
\caption{Pruning ResNet-50 on ImageNet.
$\dagger$ denotes our re-implementation.}
\label{tab:resnet50_imagenet}
\begin{tabular}{@{}clccc@{}}
\toprule
FLOPs zone & method & FLOPs (G) & Top-1 (\%) & Top-5 (\%) \\
\midrule
- & ResNet-50~\citep{he2016deep}$^\dagger$ &
4.12 & 76.88 & 93.44 \\
\midrule
\multirow{7}{*}{2G}
& Thinet-50~\citep{luo2017thinet} &
2.10 & 74.70 & 90.02 \\
& EagleEye~\citep{li2020eagleeye} &
2.00 & 76.40 & 92.89 \\
& MetaPruning~\citep{liu2019metapruning} &
2.00 & 75.40 & -     \\
& AutoSlim~\citep{yu2019autoslim} &
1.00 & 75.60 & -     \\
& HRank~\citep{lin2020hrank} &
2.30 & 74.98 & 92.33 \\
& FPGM~\citep{he2019filter} &
1.92 & 74.83 & 92.32 \\
& \textbf{Ours} ($c_{\textnormal{flops}} = 0.5$) &
2.06 & 76.90 & 93.55 \\
\midrule
\multirow{6}{*}{1G}
& ThiNet-30~\citep{luo2017thinet} &
1.20 & 72.10 & 88.30 \\
& EagleEye~\citep{li2020eagleeye} &
1.00 & 74.20 & 91.77 \\
& MetaPruning~\citep{liu2019metapruning} &
1.00 & 73.40 & -     \\
& AutoSlim~\citep{yu2019autoslim} &
1.00 & 74.00 & -     \\
& HRank~\citep{lin2020hrank} &
1.55 & 71.98 & 91.01 \\
& \textbf{Ours} ($c_{\textnormal{flops}} = 0.25$) &
1.03 & 74.96 & 92.51 \\
\bottomrule
\end{tabular}
\end{table}





\section{Conclusion and discussion}

In this work, we introduce network pruning spaces for exploring general pruning principles.
Inspired by~\citep{radosavovic2019network,radosavovic2020designing}, we focus on the structure aspect of subnetwork architectures by comparing populations of subnetworks, instead of producing the best single pruning recipe.
Although our empirical studies do not lead to a common pattern from the perspective of architecture, we observe that there exists an optimal mean computation budget in a pruning regime.
Moreover, our observations suggest that the limitations of performance in different pruning regimes might be predictable by a function of the optimal mean computation budget.
Our work empirically provides insight into the existence of such a functional form that approximates the accuracy drops and characterizes filter pruning.



\bibliography{iclr2022_conference}

\begin{thebibliography}{24}
\providecommand{\natexlab}[1]{#1}
\providecommand{\url}[1]{\texttt{#1}}
\expandafter\ifx\csname urlstyle\endcsname\relax
  \providecommand{\doi}[1]{doi: #1}\else
  \providecommand{\doi}{doi: \begingroup \urlstyle{rm}\Url}\fi

\bibitem[Frankle \& Carbin(2019)Frankle and Carbin]{frankle2018lottery}
Jonathan Frankle and Michael Carbin.
\newblock The lottery ticket hypothesis: Finding sparse, trainable neural
  networks.
\newblock In \emph{ICLR}, 2019.

\bibitem[He et~al.(2016)He, Zhang, Ren, and Sun]{he2016deep}
Kaiming He, Xiangyu Zhang, Shaoqing Ren, and Jian Sun.
\newblock Deep residual learning for image recognition.
\newblock In \emph{CVPR}, pp.\  770--778, 2016.

\bibitem[He et~al.(2019)He, Liu, Wang, Hu, and Yang]{he2019filter}
Yang He, Ping Liu, Ziwei Wang, Zhilan Hu, and Yi~Yang.
\newblock Filter pruning via geometric median for deep convolutional neural
  networks acceleration.
\newblock In \emph{CVPR}, pp.\  4340--4349, 2019.

\bibitem[He et~al.(2018)He, Lin, Liu, Wang, Li, and Han]{he2018amc}
Yihui He, Ji~Lin, Zhijian Liu, Hanrui Wang, Li-Jia Li, and Song Han.
\newblock Amc: Automl for model compression and acceleration on mobile devices.
\newblock In \emph{ECCV}, pp.\  784--800, 2018.

\bibitem[Howard et~al.(2017)Howard, Zhu, Chen, Kalenichenko, Wang, Weyand,
  Andreetto, and Adam]{howard2017mobilenets}
Andrew~G Howard, Menglong Zhu, Bo~Chen, Dmitry Kalenichenko, Weijun Wang,
  Tobias Weyand, Marco Andreetto, and Hartwig Adam.
\newblock Mobilenets: Efficient convolutional neural networks for mobile vision
  applications.
\newblock \emph{arXiv preprint arXiv:1704.04861}, 2017.

\bibitem[Krizhevsky(2009)]{krizhevsky2009learning}
Alex Krizhevsky.
\newblock Learning multiple layers of features from tiny images.
\newblock \emph{Tech Report}, 2009.
\newblock URL \url{https://www.cs.toronto.edu/~kriz/cifar.html}.

\bibitem[Le \& Hua(2021)Le and Hua]{le2021network}
Duong~Hoang Le and Binh-Son Hua.
\newblock Network pruning that matters: A case study on retraining variants.
\newblock In \emph{ICLR}, 2021.

\bibitem[Li et~al.(2020)Li, Wu, Su, and Wang]{li2020eagleeye}
Bailin Li, Bowen Wu, Jiang Su, and Guangrun Wang.
\newblock Eagleeye: Fast sub-net evaluation for efficient neural network
  pruning.
\newblock In \emph{ECCV}, pp.\  639--654. Springer, 2020.

\bibitem[Li et~al.(2017)Li, Kadav, Durdanovic, Samet, and Graf]{li2017pruning}
Hao Li, Asim Kadav, Igor Durdanovic, Hanan Samet, and Hans~Peter Graf.
\newblock Pruning filters for efficient convnets.
\newblock In \emph{ICLR}, 2017.

\bibitem[Lin et~al.(2020)Lin, Ji, Wang, Zhang, Zhang, Tian, and
  Shao]{lin2020hrank}
Mingbao Lin, Rongrong Ji, Yan Wang, Yichen Zhang, Baochang Zhang, Yonghong
  Tian, and Ling Shao.
\newblock Hrank: Filter pruning using high-rank feature map.
\newblock In \emph{CVPR}, pp.\  1529--1538, 2020.

\bibitem[Liu et~al.(2019{\natexlab{a}})Liu, Mu, Zhang, Guo, Yang, Cheng, and
  Sun]{liu2019metapruning}
Zechun Liu, Haoyuan Mu, Xiangyu Zhang, Zichao Guo, Xin Yang, Kwang-Ting Cheng,
  and Jian Sun.
\newblock Metapruning: Meta learning for automatic neural network channel
  pruning.
\newblock In \emph{ICCV}, pp.\  3296--3305, 2019{\natexlab{a}}.

\bibitem[Liu et~al.(2019{\natexlab{b}})Liu, Sun, Zhou, Huang, and
  Darrell]{liu2019rethinking}
Zhuang Liu, Mingjie Sun, Tinghui Zhou, Gao Huang, and Trevor Darrell.
\newblock Rethinking the value of network pruning.
\newblock In \emph{ICLR}, 2019{\natexlab{b}}.

\bibitem[Loshchilov \& Hutter(2017)Loshchilov and Hutter]{loshchilov2017sgdr}
Ilya Loshchilov and Frank Hutter.
\newblock Sgdr: Stochastic gradient descent with warm restarts.
\newblock In \emph{ICLR}, 2017.

\bibitem[Luo et~al.(2017)Luo, Wu, and Lin]{luo2017thinet}
Jian-Hao Luo, Jianxin Wu, and Weiyao Lin.
\newblock Thinet: A filter level pruning method for deep neural network
  compression.
\newblock In \emph{ICCV}, pp.\  5058--5066, 2017.

\bibitem[Meng et~al.(2020)Meng, Cheng, Li, Luo, Guo, Lu, and
  Sun]{meng2020pruning}
Fanxu Meng, Hao Cheng, Ke~Li, Huixiang Luo, Xiaowei Guo, Guangming Lu, and Xing
  Sun.
\newblock Pruning filter in filter.
\newblock \emph{arXiv preprint arXiv:2009.14410}, 2020.

\bibitem[Molchanov et~al.(2019)Molchanov, Mallya, Tyree, Frosio, and
  Kautz]{molchanov2019importance}
Pavlo Molchanov, Arun Mallya, Stephen Tyree, Iuri Frosio, and Jan Kautz.
\newblock Importance estimation for neural network pruning.
\newblock In \emph{CVPR}, pp.\  11264--11272, 2019.

\bibitem[Radosavovic et~al.(2019)Radosavovic, Johnson, Xie, Lo, and
  Doll{\'a}r]{radosavovic2019network}
Ilija Radosavovic, Justin Johnson, Saining Xie, Wan-Yen Lo, and Piotr
  Doll{\'a}r.
\newblock On network design spaces for visual recognition.
\newblock In \emph{ICCV}, pp.\  1882--1890, 2019.

\bibitem[Radosavovic et~al.(2020)Radosavovic, Kosaraju, Girshick, He, and
  Doll{\'a}r]{radosavovic2020designing}
Ilija Radosavovic, Raj~Prateek Kosaraju, Ross Girshick, Kaiming He, and Piotr
  Doll{\'a}r.
\newblock Designing network design spaces.
\newblock In \emph{CVPR}, pp.\  10428--10436, 2020.

\bibitem[Renda et~al.(2020)Renda, Frankle, and Carbin]{renda2020comparing}
Alex Renda, Jonathan Frankle, and Michael Carbin.
\newblock Comparing rewinding and fine-tuning in neural network pruning.
\newblock In \emph{ICLR}, 2020.

\bibitem[Rosenfeld et~al.(2021)Rosenfeld, Frankle, Carbin, and
  Shavit]{rosenfeld2021predictability}
Jonathan~S Rosenfeld, Jonathan Frankle, Michael Carbin, and Nir Shavit.
\newblock On the predictability of pruning across scales.
\newblock In \emph{ICML}, pp.\  9075--9083. PMLR, 2021.

\bibitem[Russakovsky et~al.(2015)Russakovsky, Deng, Su, Krause, Satheesh, Ma,
  Huang, Karpathy, Khosla, Bernstein, et~al.]{russakovsky2015imagenet}
Olga Russakovsky, Jia Deng, Hao Su, Jonathan Krause, Sanjeev Satheesh, Sean Ma,
  Zhiheng Huang, Andrej Karpathy, Aditya Khosla, Michael Bernstein, et~al.
\newblock Imagenet large scale visual recognition challenge.
\newblock \emph{IJCV}, 115\penalty0 (3):\penalty0 211--252, 2015.

\bibitem[Tan \& Le(2019)Tan and Le]{tan2019efficientnet}
Mingxing Tan and Quoc Le.
\newblock Efficientnet: Rethinking model scaling for convolutional neural
  networks.
\newblock In \emph{ICML}, pp.\  6105--6114. PMLR, 2019.

\bibitem[Yu \& Huang(2019)Yu and Huang]{yu2019autoslim}
Jiahui Yu and Thomas Huang.
\newblock Autoslim: Towards one-shot architecture search for channel numbers.
\newblock \emph{arXiv preprint arXiv:1903.11728}, 2019.

\bibitem[Zoph \& Le(2017)Zoph and Le]{zoph2017neural}
Barret Zoph and Quoc~V Le.
\newblock Neural architecture search with reinforcement learning.
\newblock In \emph{ICLR}, 2017.

\end{thebibliography}
\bibliographystyle{iclr2022_conference}

\end{document}